\documentclass{article}

\usepackage{times}
\usepackage{graphicx} 
\usepackage{subfigure} 

\usepackage{natbib}

\usepackage{algorithm}
\usepackage{algorithmic}

\usepackage{hyperref}
\usepackage{natbib}

\usepackage{amsmath,amssymb}
\usepackage{multirow}
\usepackage{math_definitions}

\newcommand{\removed}[1]{}

\newcommand\err{\textit{err}}
\newcommand\loss{\ell}
\newcommand\popP{\textit{P}}

\usepackage[accepted]{icml_arxiv}

\icmltitlerunning{Reducing Runtime by Recycling Samples}

\begin{document} 

\twocolumn[
\icmltitle{Reducing Runtime by Recycling Samples}

\icmlauthor{Jialei Wang}{jialei@uchicago.edu}
\icmladdress{Department of Computer Science, University of Chicago, Chicago, IL 60637, USA}
\icmlauthor{Hai Wang}{haiwang@ttic.edu}
\icmladdress{Toyota Technological Institute at Chicago, Chicago, IL 60637, USA}
\icmlauthor{Nathan Srebro}{nati@ttic.edu}
\icmladdress{Toyota Technological Institute at Chicago, Chicago, IL 60637, USA}

\icmlkeywords{boring formatting information, machine learning, ICML}

\vskip 0.3in
]

\begin{abstract}
  Contrary to the situation with stochastic gradient descent, we
  argue that when using stochastic methods with variance reduction,
  such as SDCA, SAG or SVRG, as well as their variants, it could be beneficial to reuse
  previously used samples instead of fresh samples, even when fresh
  samples are available.  We demonstrate this empirically for SDCA, SAG and SVRG,
  studying the optimal sample size one should use, and also uncover
  behavior that suggests running SDCA for an integer number of epochs
  could be wasteful.
\end{abstract}

\section{Introduction}

When using a stochastic optimization approach, is it always beneficial
to use all available training data, if we have enough time to do so?
Is it always best to use a fresh example at each iteration, thus
maximizing the number of samples used?  Or is it sometimes better to
revisit an old example, even if fresh examples are available?

In this paper, we revisit the notion of ``more data less work'' for
stochastic optimization \citep{mdlw}, in light of recently proposed
variance-reducing stochastic optimization techniques such as SDCA
\citep{original-sdca-paper,shai-tong-analyzis}, SAG \citep{sag} and SVRG
\citep{svrg}.  We consider smooth SVM-type training, i.e.,~regularized
loss minimization for a smooth convex loss, in the data laden regime.
That is, we consider a setting where we have infinite data and are
limited only by time budget, and the goal is to get the best
generalization (test) performance possible within the time budget
(using as many examples as we would like).  We then ask what is
the optimal training set size to use?  If we can afford making $T$
stochastic iterations, is it always best to use $m=T$ independent
training examples, or might it be beneficial to use only $m<T$
training examples, revisiting some of the examples multiple times
(visiting each example $T/m$ times on average)? Can using less
training data actually improve performance (and conversely, using more
data hurt performance)?

We discuss how with Stochastic Gradient Descent (SGD), there is indeed
no benefit to using less data than is possible, but with
variance-reducing methods such as SDCA and SAG, it might indeed be
possible to gain by using a smaller training set, revisiting examples
multiple times.  We first present qualitative arguments focusing on
the error decomposition showing this could be possible (Section
\ref{sec:analysis}), also revisiting the ``more data less work'' SGD
upper bound analysis.  We also conduct careful experiments with SDCA,
SAG and SVRG on several standard datasets and empirically demonstrate
that using a reduced training set can indeed significantly improve
performance (Section \ref{sec:empirical}). In analyzing these
experiments, we also uncover a previously undiscovered phenomena
concerning the behavior of SDCA which suggests running SDCA for an
integer number of epochs could be bad, and which greatly affects the
``optimal sample size'' question (Section \ref{sec:closer}).

Following the presentation of SDCA, SVRG and SAG, a long list of variants and other
methods with similar convergence guarantees have also been presented,
including EMGD
\cite{zhang2013linear}, S2GD \cite{konevcny2013semi}, Iprox-SDCA
\citep{zhao2014}, Prox-SVRG \citep{xiao2014proximal}, SAGA
\citep{defazio2014saga}, Quartz \citep{qu2014}, AccSDCA
\cite{DBLP:conf/icml/Shalev-Shwartz014}, AccProxSVRG
\cite{nitanda2014stochastic}, Finito \cite{defazio2014finito},
SDCA-ADMM \cite{suzuki2014stochastic}, MISO
\cite{mairal2015incremental}, APCG \cite{lin2015accelerated}, APPA
\cite{frostig2015regularizing}, SPDC \cite{zhang2015spdc}, AdaSDCA
\cite{csiba2015adasdca}, Catalyst \cite{lin2015universal}, RPDG
\cite{lan2015optimal}, NU-ACDM \cite{zhu2015even}, Affine SDCA and
SVRG \citep{vainsencher2015local}, Batching SVRG
\citep{babanezhad2015stop}, and $\varepsilon\Ncal$-SAGA
\cite{hofmann2015neighborhood}, emphasizing the importance of these
methods.  We experiment with SAG, SVRG and especially SDCA as
representative examples of such methods---the ideas we outline apply
also to the other methods in this family.

\section{Preliminaries: SVM-Type Objectives and Stochastic Optimization}
\label{sec:preliminaries}

Consider SVM-type training, where we learn a linear predictor by
regularized empirical risk minimization with a convex loss (hinge loss
for SVMs, or perhaps some other loss such as logistic or smoothed
hinge). That is, learning a predictor $\wb$ by minimizing the
empirical objective:
\begin{equation}
  \label{eq:empobj}
  \min_\wb P_m(\wb) = \frac{1}{m} \sum_{i=1}^m \loss(
  \inner\wb{\xb_i},y_i) + \frac{\lambda}{2} \norm{\wb}^2,
\end{equation}
where $\loss(z)$ is a convex surrogate loss, $\{\xb_i,y_i\}$ are i.i.d
training samples from a source (population) distribution and our goal
is to get low generalization error
$\EE_{\xb,y}[\err(\inner\wb\xb,y)]$.  Stochastic optimization, in
which a single sample $\xb_i,y_i$ (or a small mini-batch of samples)
is used at each iteration, is now the dominant approach for problems
of the form \eqref{eq:empobj}.  The success of such methods has been
extensively demonstrated empirically
\cite{pegasos,original-sdca-paper,leon-bouton-webpage,sag,svrg}, and it
has also been argued that stochastic optimization, and stochastic
gradient descent (SGD) in particular, is in a sense optimal for the
problem, when what we are concerned with is the expect generalization
error \cite{DBLP:conf/nips/BottouB07,mdlw,ohad-remove-log-factors,DBLP:conf/aistats/DefossezB15}.

When using SGD to optimize \eqref{eq:empobj}, at each iteration we use
one random training sample $(\xb_i,y_i)$ and update
$\wb_{t+1}\leftarrow\wb_{t}-\eta g$ where $g=\nabla_\wb
\loss(\inner\wb{\xb_i},y_i)+\lambda \wb$ is a stochastic estimation of
$\nabla P_m(\wb)$ based on the single sample.  In fact, we can also view
$g$ as a stochastic gradient estimation of the regularized {\em population}
objective:
\begin{equation}
  \label{eq:popobj}
  \popP(\wb) = \EE_{\xb,y}[\loss(\inner\wb\xb,y)] + \frac{\lambda}{2}\norm\wb^2.
\end{equation}
That is, each step of SGD on the empirical objective
\eqref{eq:empobj}, can also be viewed as an SGD step on the population
objective \eqref{eq:popobj}.  If we sample from a training set of size
$m$ without replacements, the first $m$ iterations of SGD on the
empirical objective \eqref{eq:empobj}, i.e., one-pass-SGD, will exactly
be $m$ iterations of SGD on the population objective
\eqref{eq:popobj}.  But, sampling with replacement from a finite set
of $m$ samples creates dependencies between the samples used in
different iterations, when viewed as samples from the source
(population) distribution.  Such repeated use of samples harms the
optimization of the population objective \eqref{eq:popobj}.  Since the
population objective better captures the expected error, it seems we
would be better off using fresh samples, if we had them, rather than
reusing previously used sample points, in subsequent iterations of
SGD.  Let us understand this observation better.

\section{To Resample Or Not to Resample?}
\label{sec:resample}

Suppose we have an infinite amount of data available. E.g., we have a
way of obtaining samples on-demand very cheaply, or we have more data
than we could possibly use. Instead, our limiting resource is
running time. What is the best we can do with infinite data and a
time-budget of $T$ gradient calculation? One option is to run SGD on $T$
independent and fresh samples. We can think of this as SGD on the
population objective $\popP$, or as one-pass SGD (without replacement)
on an empirical objective $P_T$ (based on a training set of size
$T$). Could it possible be better to use only $m=c \cdot T<T$ samples, for
some $0<c<1$, and run SGD on $P_m$ for $T$ iterations?

\subsection{SGD Likes it Fresh}

One way to argue for the one-pass fresh-sample approach is that, in a
worst-case sense, one-pass SGD is optimal, in that it is guaranteed to
attain the best generalization error that can always be ensured (based
on the norm of the data and the predictor)\removed{\footnote{In fact, in such a
 one-pass approach, an explicit regularization term is not even
 needed.} \citep{icml-tutorial}}. Using SGD with less data can only
guarantee worse generalization error. Indeed, nothing we do with less
data can ensure better error. However, such an argument is based on
the worst-case behavior, which is rarely encountered in practice.
E.g., in practice we know that multi-pass SGD (i.e.,~running SGD for
more iterations using the same number of samples) typically do
reduce the generalization error. Could we argue for fresh samples
without reverting to worst-case analysis? Although doing so
analytically is tricky, as our understanding of better-than-worst-case
SGD behavior is very limited, we can get significant insight from
considering the error decomposition.

Let us consider the effect on the generalization error of running $T$
iterations of SGD on an empirical objective $P_m$ based on $m$
samples, versus $T$ iterations of SGD on an empirical objective
$P_{m'}$ based on $m'~(m' < m)$ samples. The running time in both cases is the
same. More importantly, the ``optimization error'', i.e.,~the
sub-optimality of the empirical objective will likely be
similar\footnote{\removed{This is at least the case for non-smooth objectives,
  like the SVM hinge loss considered by \cite{mdlw}, where the
  optimization error is dominated by the magnitude of the
  gradients---something that does not change as we change $m$. }With a
smaller data set the variance of the stochastic gradient estimate is
slightly reduced, but only by a factor of $1-1/m$, which might
theoretically very slightly reduce the empirical optimization error.
But, e.g., with over 1,000 samples the reduction is by less than a
tenth of a percent and we do not believe this low order effect has any
significance in practice.}. However, the estimation error, that
is the difference between optimizing the population objective
\eqref{eq:popobj} and the empirical objective is lower as we have more
samples. More precisely, we have that $\popP(\wb_m)-\inf_{\wb} \popP(\wb) \leq
O(1/(\lambda m))$ where $\wb_m=\arg\min P_m(\wb)$
\citep{fast-rates-for-regularized-objectives}. To summarize, if using
more samples, we have the same optimization error for the same
runtime, but better estimation error, and can therefor expect that our
predictions are better. Viewed differently, and as pointed out by
\cite{mdlw}, with a larger sample size we can get to the same
generalization error in less time.

This indeed seems to be the case for SGD. But is it the case also for
more sophisticated stochastic methods with better optimization
guarantees?

\subsection{Reduced Variance Stochastic Optimization}

Stochastic Gradient Descent is appropriate for any objective for which
we can obtain stochastic gradient estimates. E.g., we can use it
directly on the expected objective \eqref{eq:popobj}, even if we can't
actually calculate it, or its gradient, exactly. But in the past
several years, several stochastic optimization methods have been
introduced that are specifically designed for objectives which are
finite averages, as in \eqref{eq:empobj}. SDCA
\cite{original-sdca-paper,shai-tong-analyzis,DBLP:conf/icml/Shalev-Shwartz014} and SAG
\cite{sag,Schmidt13minimizingfinite} are both stochastic optimization methods with almost
identical cost-per iteration as SGD, but they
maintain information on each of the $m$ training points, in the form
of dual variables or cached gradients, that help them make reduced
variance steps in subsequent passes over the data \citep[see, e.g.,
discussion in][Section 4]{svrg}, thus improving convergence to the
optimum of \eqref{eq:empobj}. This lead to the introduction of SVRG
\citep{svrg,kakade2015svrg}, which also reduces the variance of stochastic steps by
occasionally recalculating the entire gradient (on all $m$ training
points), and achieves a similar runtime guarantee as SAG and SDCA.
For both SDCA and SAG, and also for SVRG in relevant regimes. The number of iterations required to achieve a sub optimality of
$\epsilon$ on \eqref{eq:empobj} when the loss $\loss(\cdot)$ is smooth
is
\begin{equation}
  \label{eq:sdcabound}
T=O\left(\rbr{\frac{1}{\lambda}+m}\log \frac{1}{\epsilon}\right),
\end{equation}
compared to $O\left(\frac{1}{\lambda \epsilon}\right)$ for SGD. That
is, these methods can reduce the optimization error faster than SGD,
but unlike SGD their runtime depends on the sample size $m$. Say
differently, with a smaller sample size, they can potentially obtain a
smaller optimization error in the same amount of time (same number of
iterations).

How does this affect the answer to our question?  What is the best we
can do with infinite data and a time-budget of $T$ iterations with
such methods? Could it be better to use only $m=c \cdot T<T$ samples, for
some $0<c<1$?

\subsection{Error Decomposition for Reduced Variance Methods}

Let us revisit the error decomposition discussion from before. If we
use $m'<m$ samples, the estimation error could indeed be larger.
However, unlike for SGD, using less samples provides more opportunity
for variance reduction, and as discussed above and can be seen from
\eqref{eq:sdcabound}, can {\em reduce} the optimization error (or said
differently, using less samples can allow us to obtain the same
optimization error faster). That is, if we use $m'<m$ samples, we
will have a larger estimation error, but a smaller optimization error.
It might therefor be beneficial to balance these two errors, and with
the right balance there is potentially for an overall decrease in the
generalization error, if the decrease in the optimization error
out-weights the increase in the estimation error. In Section
\ref{sec:empirical} we empirically investigate the optimal sample size
$m=cT$ that achieves the best balance and lowest test error, and show
that it is indeed frequently beneficial to reuse examples, and that
this can lead to significant reduce in test error using the same
number of iterations. But first, we revisit the SGD upper bound
analysis and understand what changes when we consider reduced
variance methods instead.

\begin{table*}[t]
\begin{center}
\begin{tabular}{|c|c|c|c|c|c|c|c|c|c|c|}\hline
\multirow{2}{*}{$T\backslash$  Dataset}& \multicolumn{2}{|c|}{covtype}& \multicolumn{2}{|c|}{ijcnn1} & \multicolumn{2}{|c|}{a9a}& \multicolumn{2}{|c|}{svmguide1} & \multicolumn{2}{|c|}{w8a}\\
\cline{2-11}
&{IID}&{PERM} &{IID}&{PERM} &{IID}&{PERM} &{IID}&{PERM} &{IID}&{PERM}\\
\hline
1000 &0.975  &0.925 &0.950 &0.900 &1.000 &0.950 &0.250 &0.950  &1.000 &0.975\\
2000 &0.525  &0.925 &0.950 &0.925 &0.875 &0.925 &0.150 &0.950  &1.000 &0.925\\
4000 &0.375  &0.950 &0.650 &0.950 &0.825 &0.925 &0.125 &0.975  &1.000 &0.975\\
8000 &0.225  &0.950 &0.400 &0.925 &0.750 &0.900 &N/A &N/A      &0.875 &0.925\\
16000&0.175  &0.950 &0.350 &0.975 &0.625 &0.950 &N/A &N/A      &0.625 &0.950\\
32000&0.125  &0.950 &0.300 &0.975 &0.250 &0.875 &N/A &N/A      &0.425 &0.975\\
\hline
\end{tabular}
\end{center}
\caption{The Optimal $c$ when using SDCA under a time budget, with IID sampling and random permutation.}
\label{tab:optimalc}
\end{table*}

\section{Upper Bound Analysis}
\label{sec:analysis}

\begin{figure}[t]
\begin{center}
\includegraphics[width=0.25\textwidth]{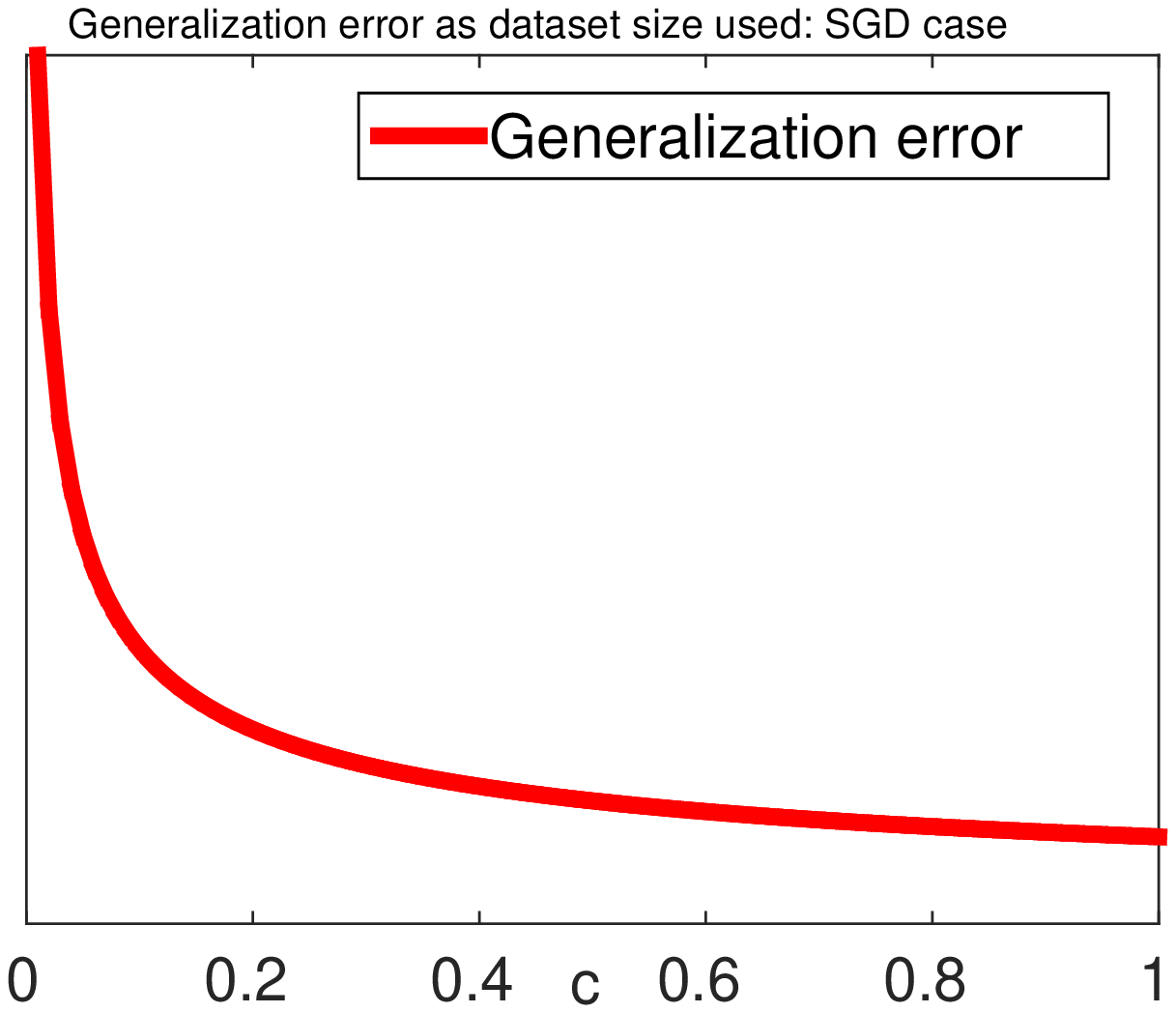}%
\includegraphics[width=0.25\textwidth]{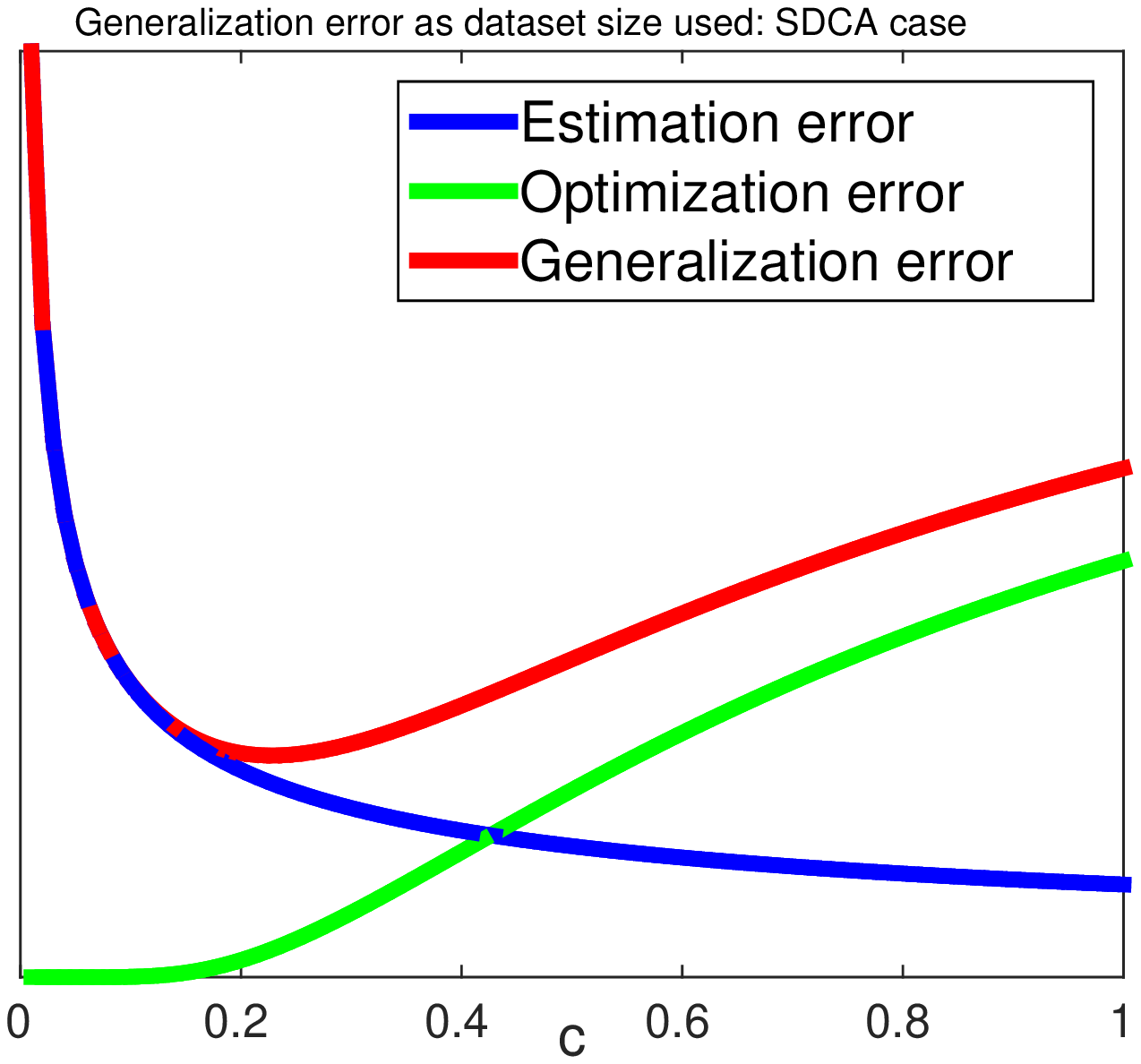}%
\end{center}
\makebox[0.25\textwidth]{\it SGD}\makebox[0.25\textwidth]{\it SDCA}
\caption{Illustration of generalization errors as $c$ varied.}
\label{fig:illustration}
\end{figure}

\begin{table}[t]
\begin{center}
\begin{tabular}{c|c|c}\hline
Dataset & $\#$ of instances & $\#$ of features\\\hline
svmguide1 &7,089 &4 \\\hline
a9a &48,842  &123 \\\hline
w8a &64,700  &300 \\\hline
ijcnn1 &141,691 &22 \\\hline
covtype &581,012 &54 \\
\hline
\end{tabular}
\end{center}
\caption{Statistics of datasets in this paper.}
\label{tab:datasetsats}
\end{table}

In this Section, we revisit the ``More Data Less Work'' SGD upper
bound analysis \cite{mdlw}. This analysis, which is based on
combining the estimation error and the SGD optimization error upper
bounds, was used to argue that for SGD increasing the training set
size can only reduce runtime and improve performance. We revisit the
analysis considering also the optimization error upper bound
\eqref{eq:sdcabound} for the reduced variance methods. We will see
that even for the reduced variance methods, relying on the norm-based
upper bounds alone does not justify an improvement with a reduced
sample size (i.e.,~a choice of $c<1$). However, as was mentioned
earlier, such an estimation error upper bound is typically too
pessimistic. We will see that heuristically assuming a lower
estimation error, does not justify a choice of $c<1$ for SGD, but {\em
  does} justify it for the reduced variance methods.

\newcommand{\wref}{\wb_0}
\newcommand{\wsgd}{\tilde{\wb}_{\small\textsc{sgd}}}
\newcommand{\wsdca}{\tilde{\wb}_{\small\textsc{rv}}}
\newcommand{\epssgd}{\epsilon_{\small\textsc{sgd}}}
\newcommand{\epssdca}{\epsilon_{\small\textsc{rv}}}

The analysis is based on the existence of a ``reference
predictor'' $\wref$ with norm $\norm{\wref}$ and expected risk
$L(\wref) = \EE[\loss(\inner\wref\xb,y)]$ \cite{mdlw}. We denote
$\wb_m$ the exact optimum of the empirical problem \eqref{eq:empobj}
and $\wsgd$ and $\wsdca$ the outputs of SGD (Pegasos) and of a reduced
variance stochastic method (e.g. SDCA) respectively after $T$
iterations using a training set of size $m=cT$. The goal is to bound
the generalization error of these predictors in terms of $L(\wref)$,
$\norm{\wref}$ and other explicit parameters.  We assume
$\norm{\xb}\leq 1$ and that the loss is $1$-Lipschitz and $1$-smooth.

The generalization errors can be bounded by the following error
decomposition (with high probability) \cite{mdlw}:
\begin{equation}
L(\tilde{\wb}) - L(\wref) \removed{\leq P(\tilde{\wb}) - P_m(\tilde{\wb}) \\
&+ P_m(\tilde{\wb}) - P_m(\wb_m) \\
&+ P_m(\wb_m) - P_m(\wref) \\
&+ P_m(\wref) - P(\wref) \\
&+ \frac{\lambda}{2} \|\wref\|^2 - \frac{\lambda}{2} \|\tilde{\wb}\|^2 \\}
\leq \epsilon(T)
+ \frac{\lambda}{2} \|\wref\|^2 + O\rbr{\frac{1}{\lambda cT}} \\
\label{eq:decom}
\end{equation}
where $\epsilon(T)\geq P_m(\tilde{\wb})-P_m(\wb_m)$ is a bound on the
suboptimality of \eqref{eq:empobj} (the ``optimization error''), and
$O(\frac{1}{\lambda cT})\geq P(\wb_m)-P(\wref)$ is the estimation
error bound\cite{fast-rates-for-regularized-objectives}. We will consider
what happens when we bound the optimization error $\epsilon(T)$ as:
\begin{align*}
&\epssgd(T) \leq O(1/(\lambda T))\\
\textrm{and as:}\quad &\epssdca(T) \leq \exp(-T/(1/\lambda+cT)).
\end{align*}
Consider the last two terms of \eqref{eq:decom} regardless of the
optimization algorithm used, even with the optimal choice $\lambda =
O\rbr{\sqrt{\frac{1}{cT\norm{\wref}^2}}} $, these two terms are at least
$O\rbr{\sqrt{\frac{\norm{\wref}^2}{cT}}}$, yielding an optimal choice of
$c=1$, and no improvement over one-pass SGD. This is true for both
SGD and the reduced variance methods, and is not surprising, since we
know that relying only on the norm of $\wref$, one-pass SGD already
yields the best possible guarantee---nothing will yield a better upper
bound using $T$ gradient estimates.

But the above analysis is based on a wort-case bound on the estimation
error of an $\ell_2$-regularized objective, which also suggests and
optimal setting of $\lambda \propto 1/\sqrt{m}$ and that multiple
passes of SGD (when the training set size is fixed) does not improve
the generalization error over a single pass of SGD (i.e.,~that taking
$T>m$ iterations is not any better than making $T=m$ iterations of
SGD, with a fixed $m$). In practice, we know that the estimation
error is often much lower, the optimal $\lambda$ is closer to $1/m$,
and that taking multiple passes of SGD certainly {\em does} improve
performance \cite{pegasos}.

\begin{figure*}[t]
\begin{center}
\includegraphics[width=0.25 \textwidth]{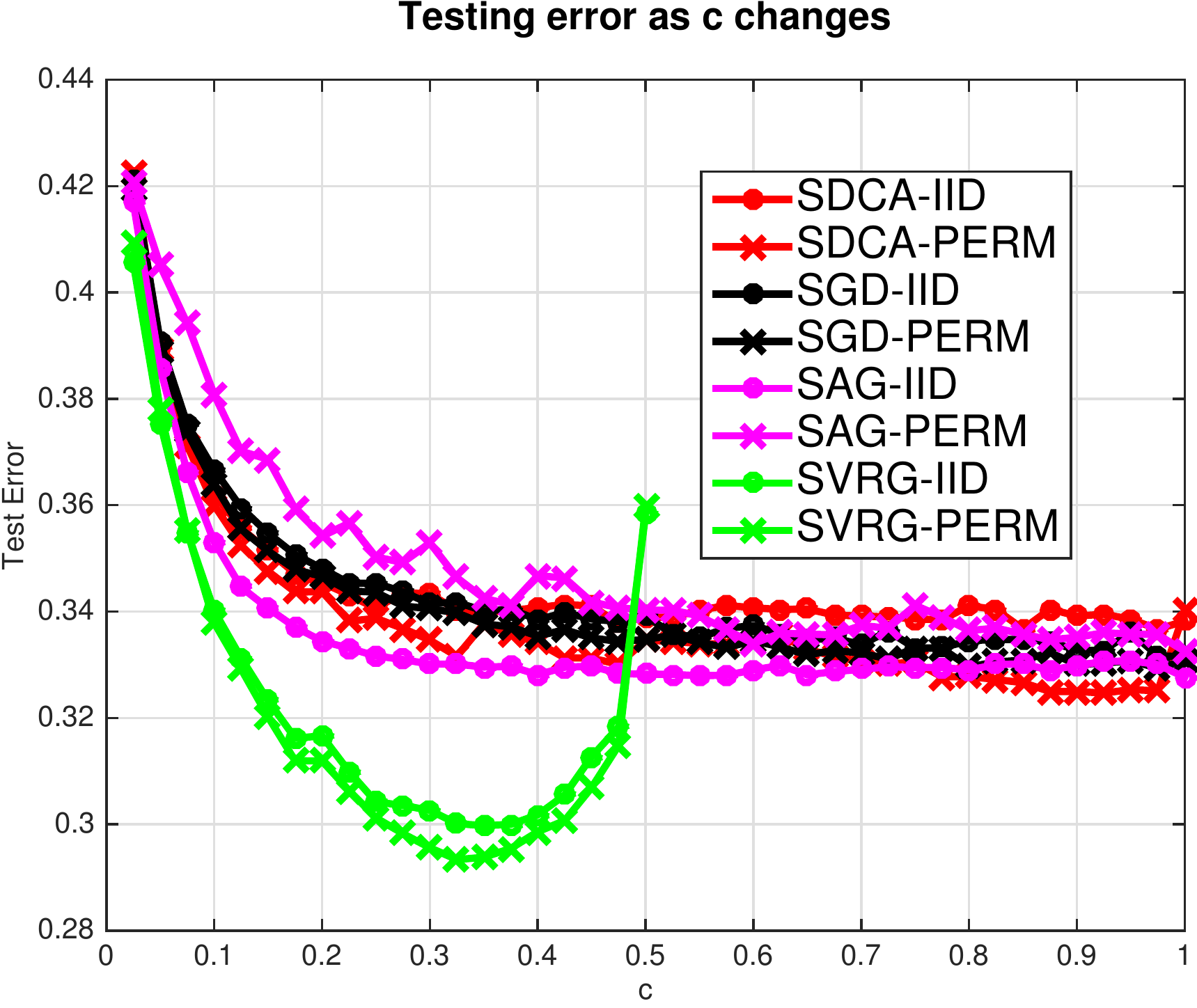}%
\includegraphics[width=0.25 \textwidth]{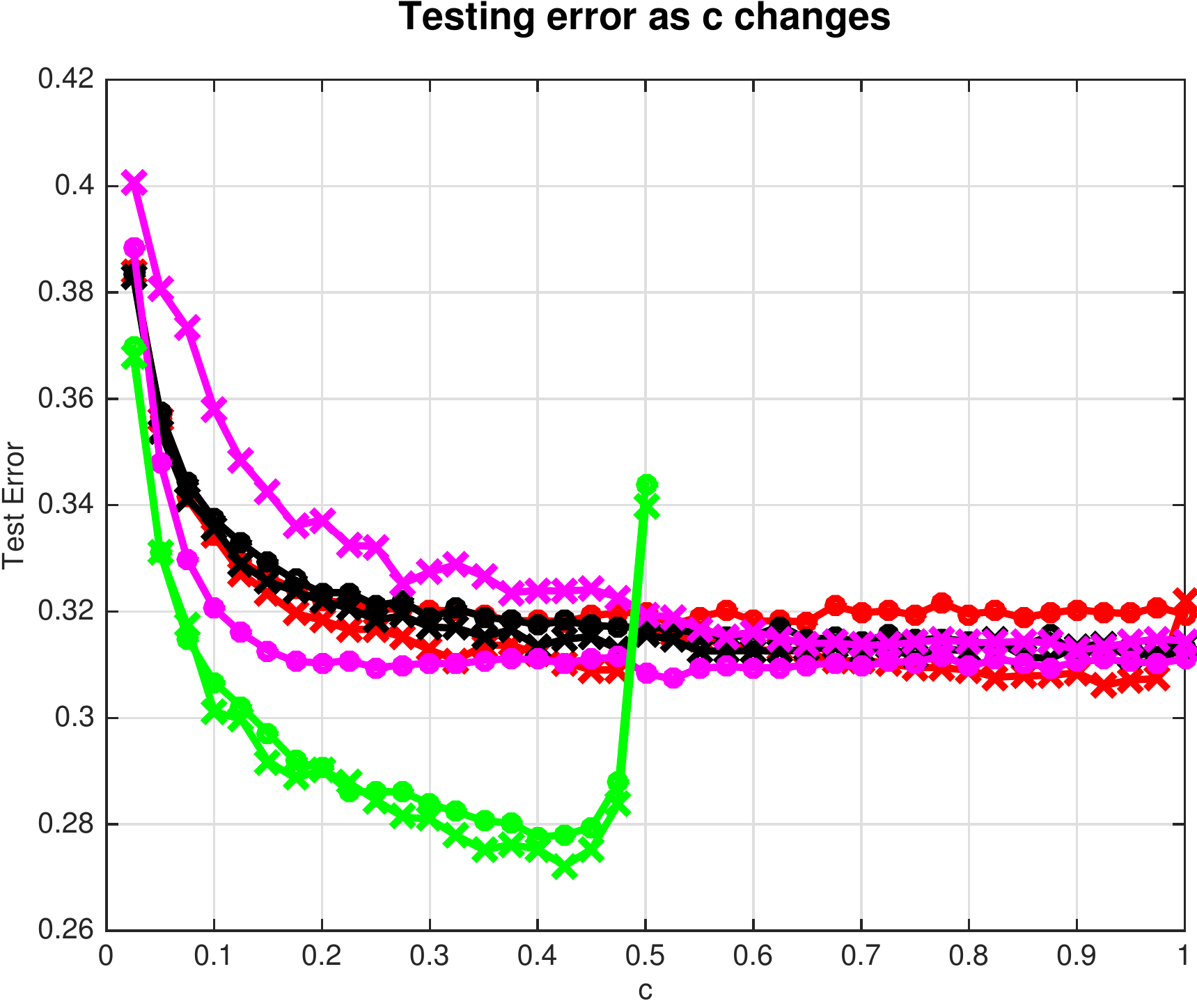}%
\includegraphics[width=0.25 \textwidth]{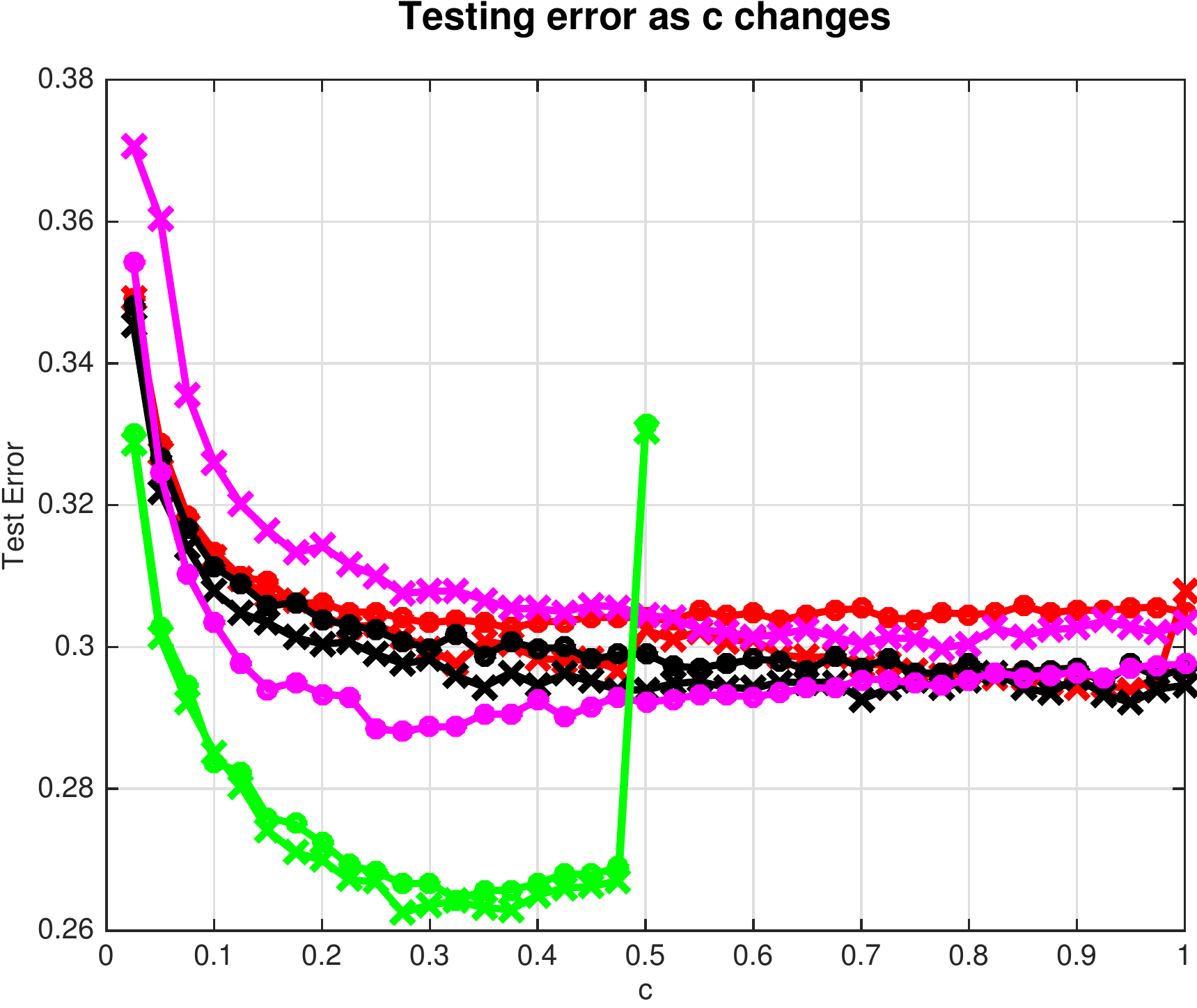}%
\includegraphics[width=0.25 \textwidth]{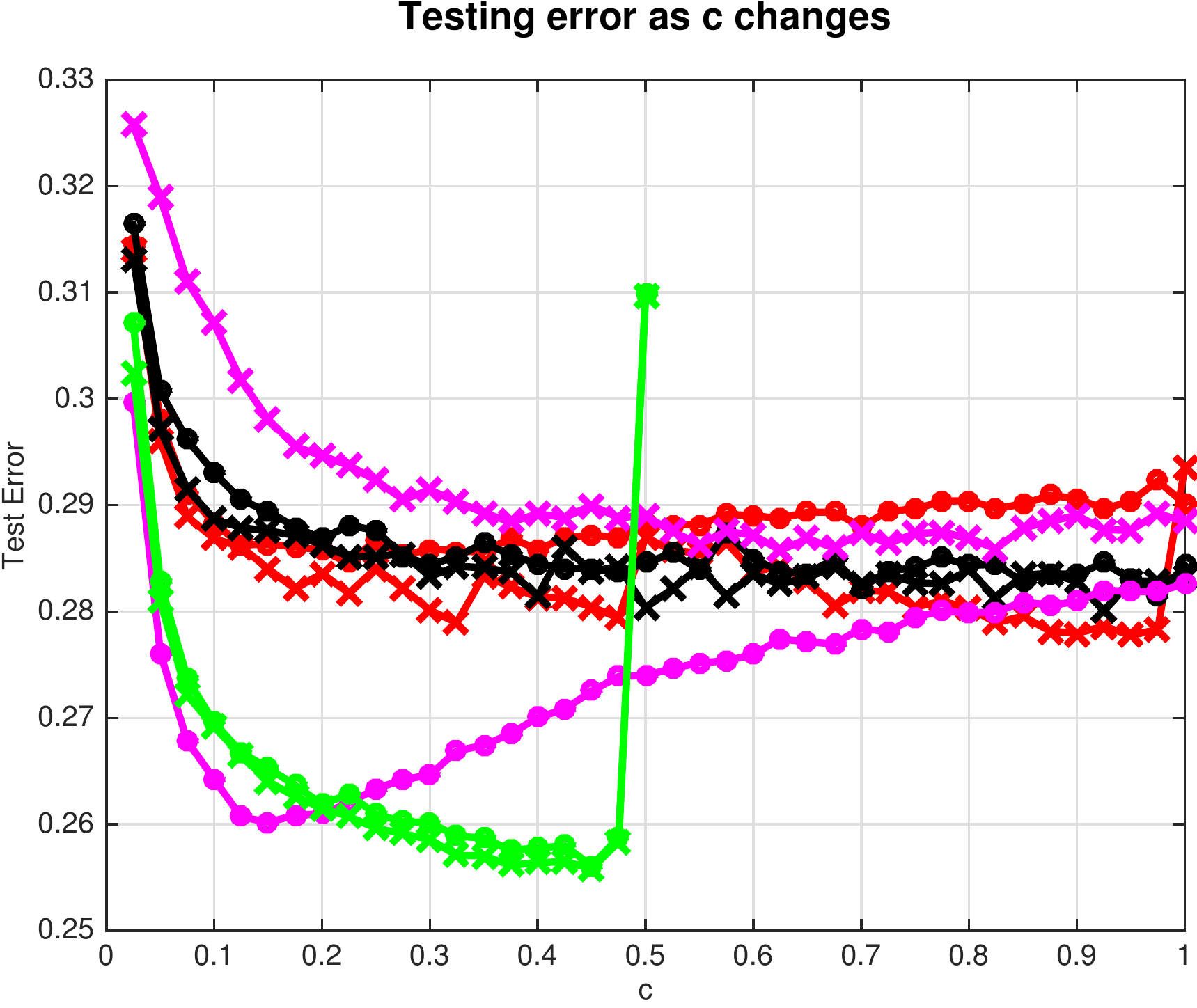}%
\end{center}
\makebox[\textwidth]{{\it covtype, from left to right: T = 1000,2000,4000,8000}}
\begin{center}
\includegraphics[width=0.25 \textwidth]{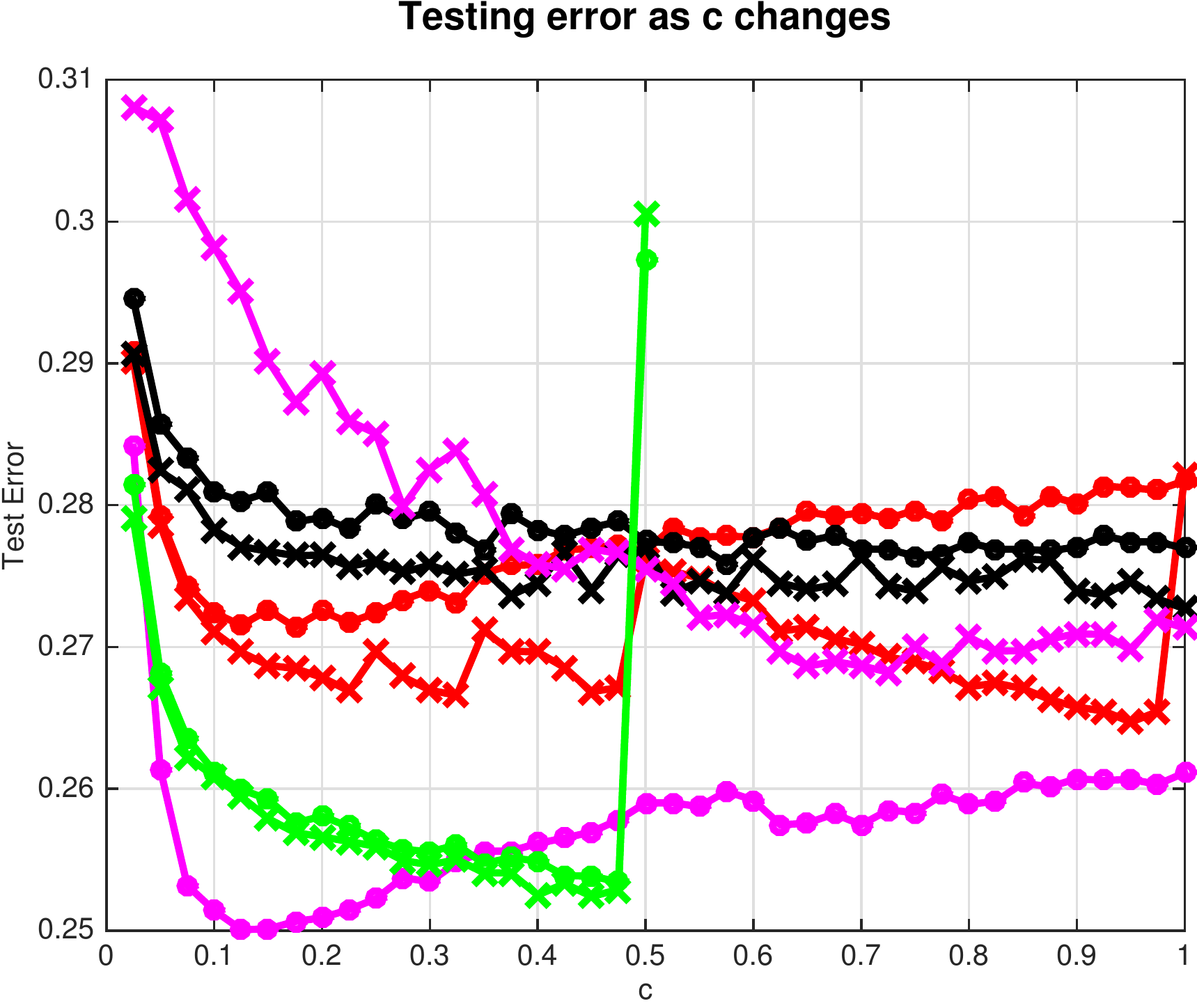}%
\includegraphics[width=0.25 \textwidth]{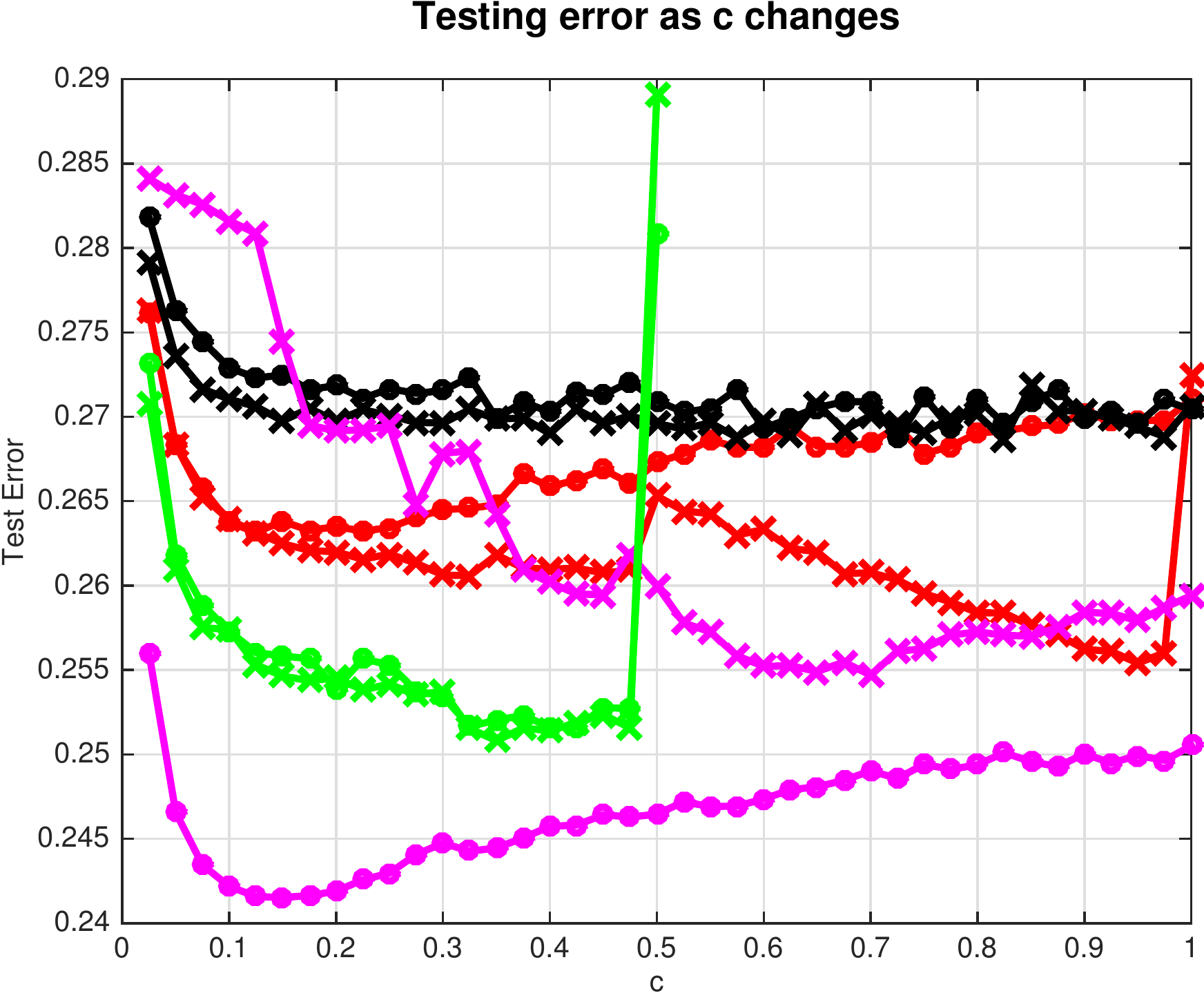}%
\includegraphics[width=0.25 \textwidth]{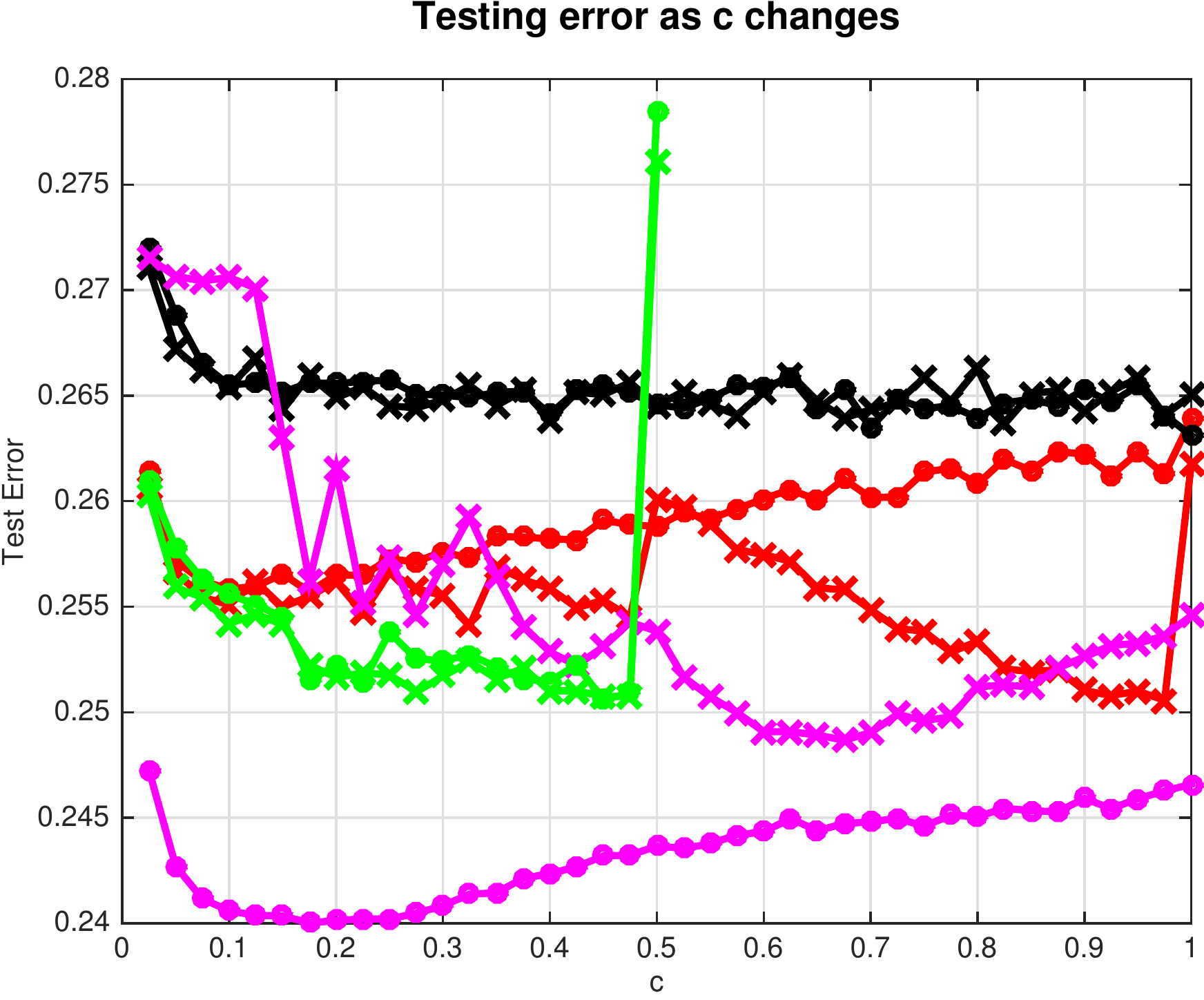}%
\includegraphics[width=0.25 \textwidth]{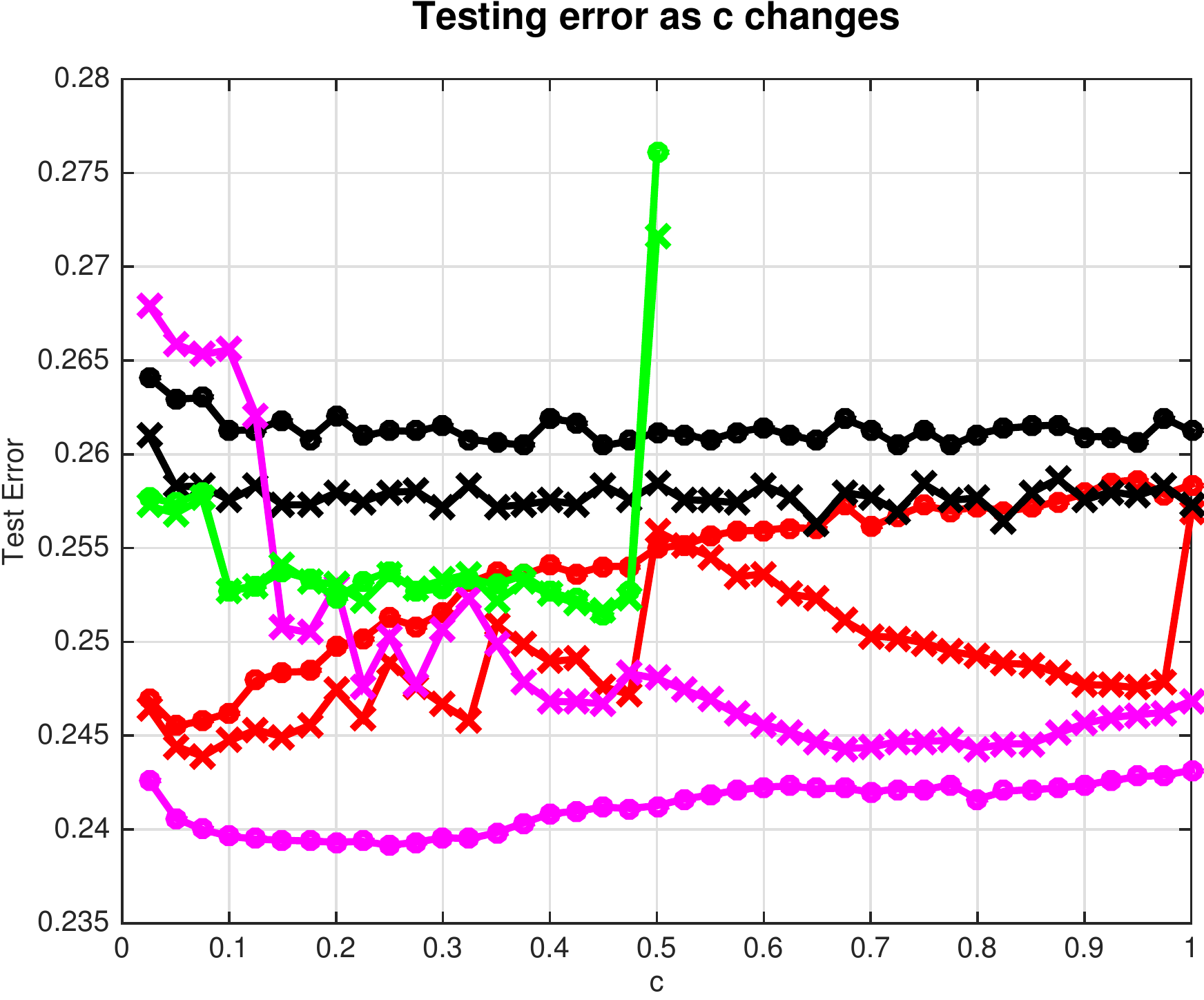}%
\end{center}
\makebox[\textwidth]{{\it covtype, from left to right: T = 16000,32000,64000,128000}}
\begin{center}
\includegraphics[width=0.25 \textwidth]{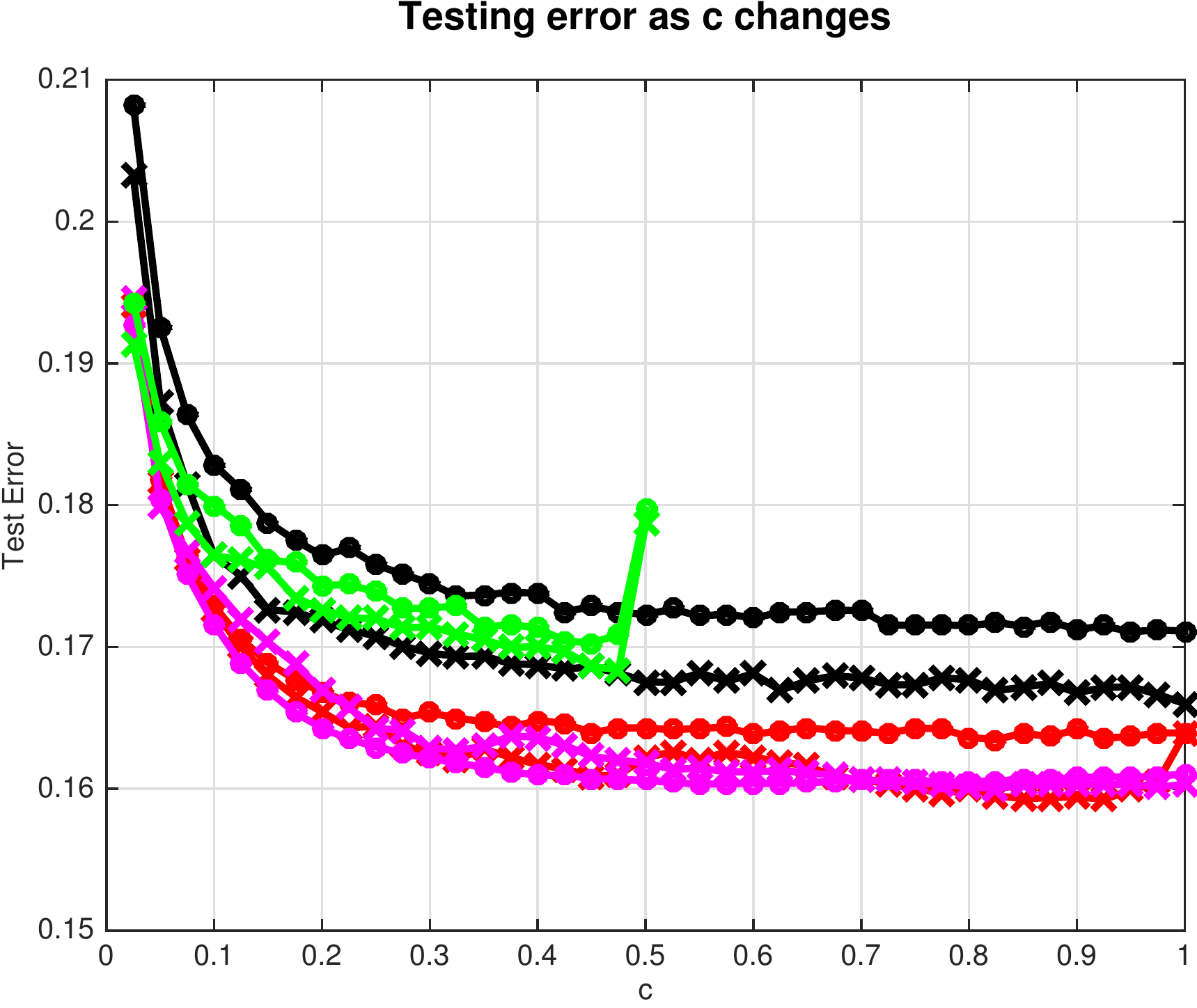}%
\includegraphics[width=0.25 \textwidth]{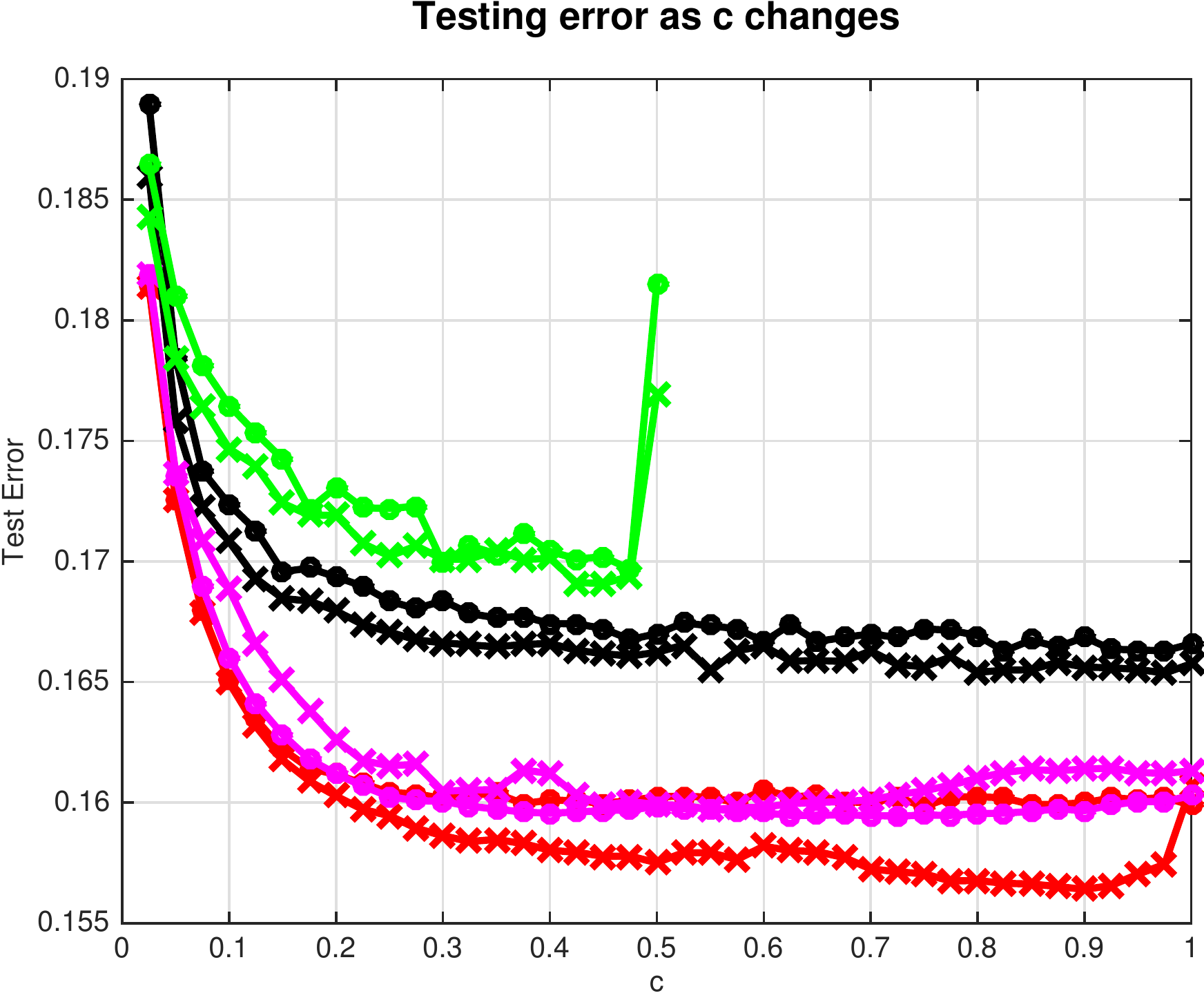}%
\includegraphics[width=0.25 \textwidth]{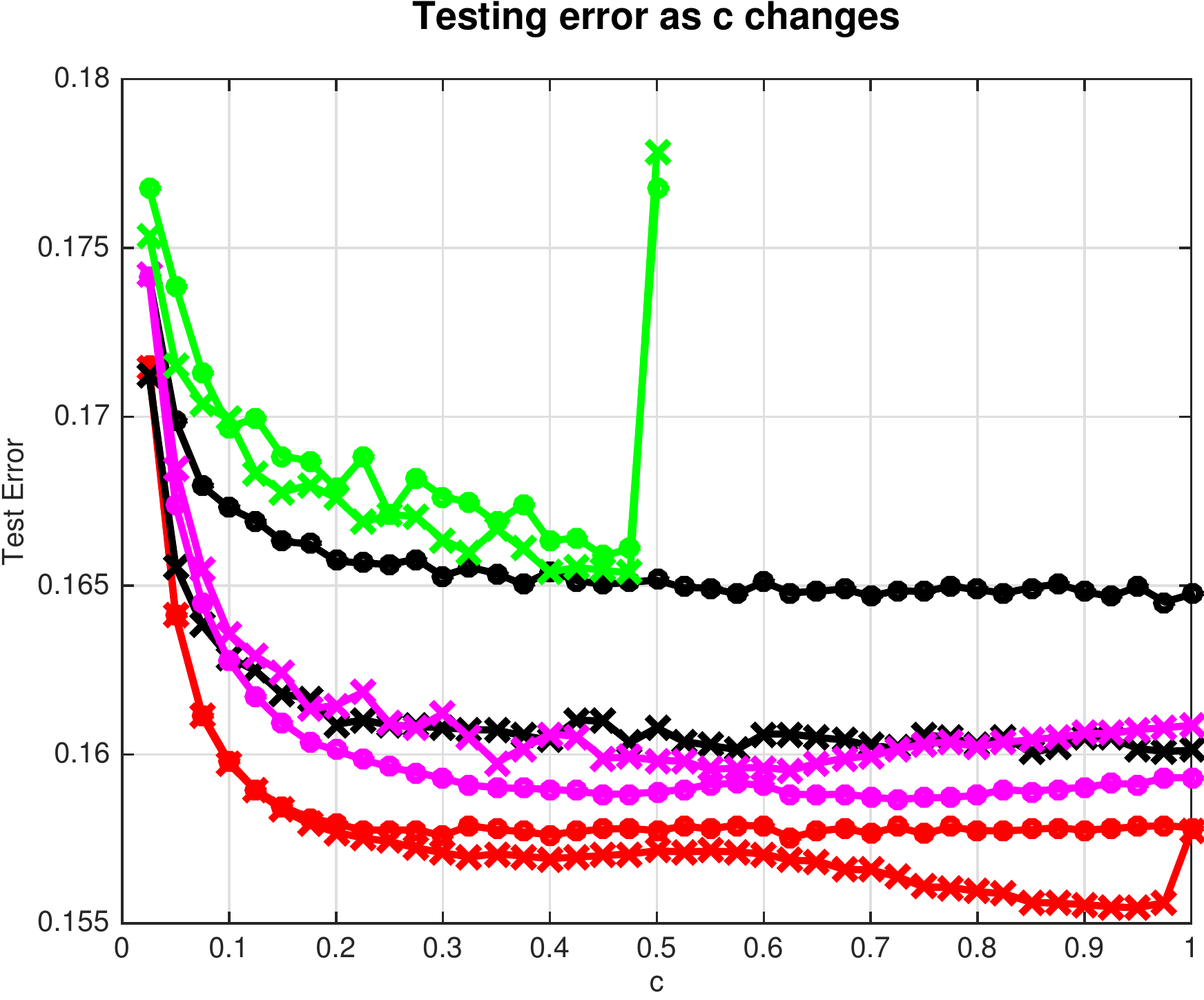}%
\includegraphics[width=0.25 \textwidth]{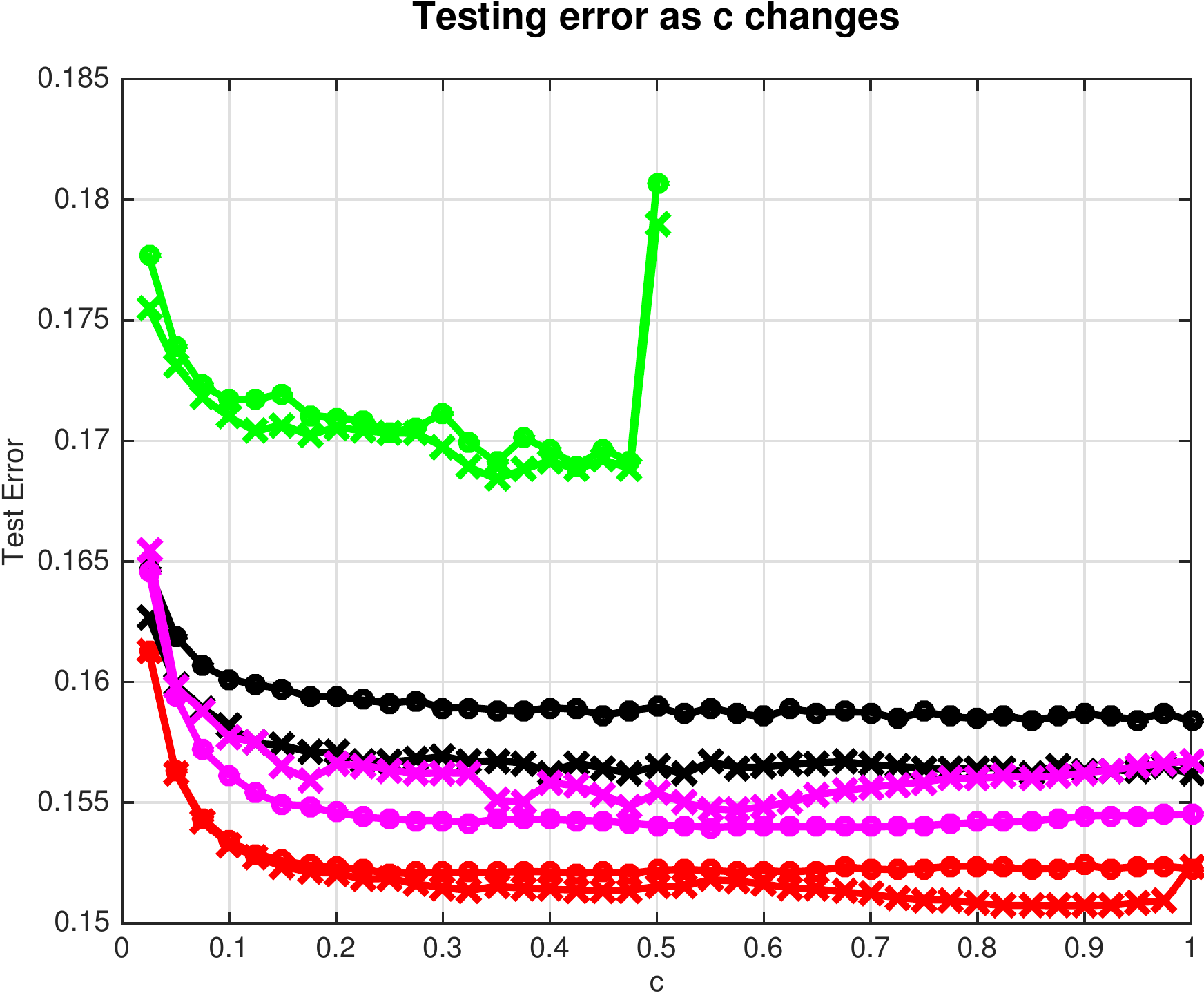}%
\end{center}
\makebox[ \textwidth]{{\it a9a, from left to right: T = 4000,8000,16000,32000}}
\caption{\small Illustration of generalization errors as $c$ varied}
\label{fig:testerror1}
\end{figure*}

\begin{figure*}[t]
\begin{center}
\includegraphics[width=0.25 \textwidth]{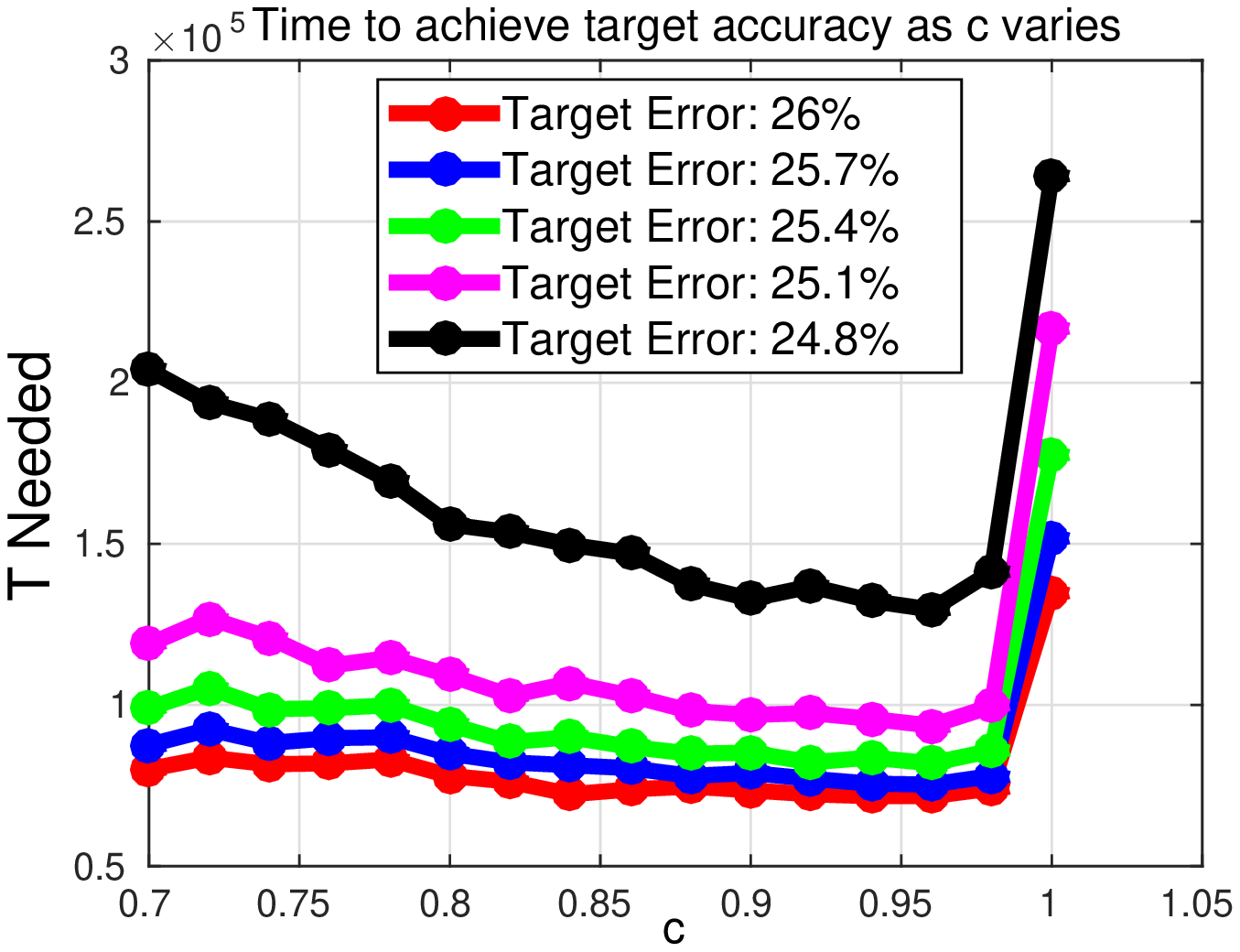}%
\includegraphics[width=0.25 \textwidth]{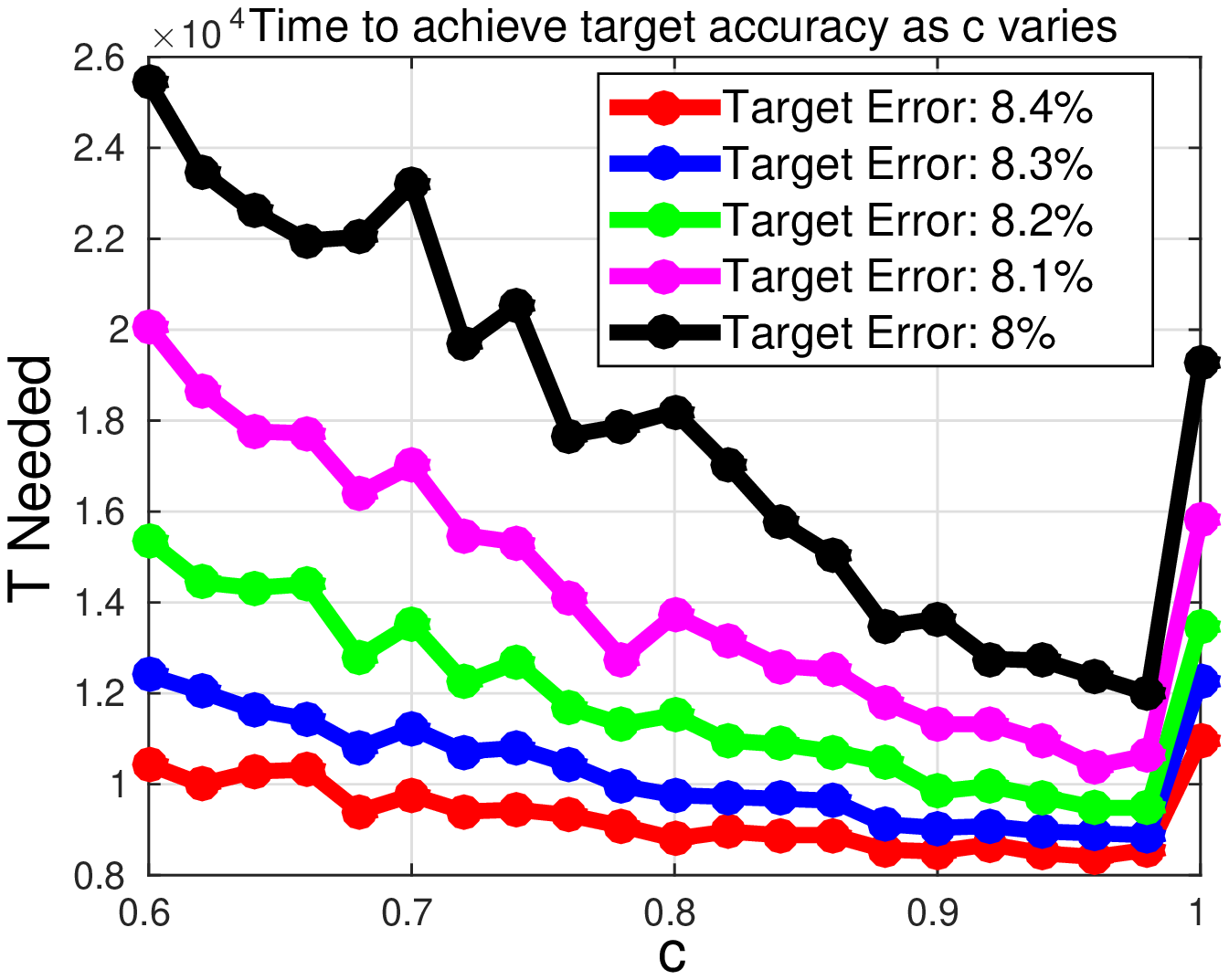}%
\includegraphics[width=0.25 \textwidth]{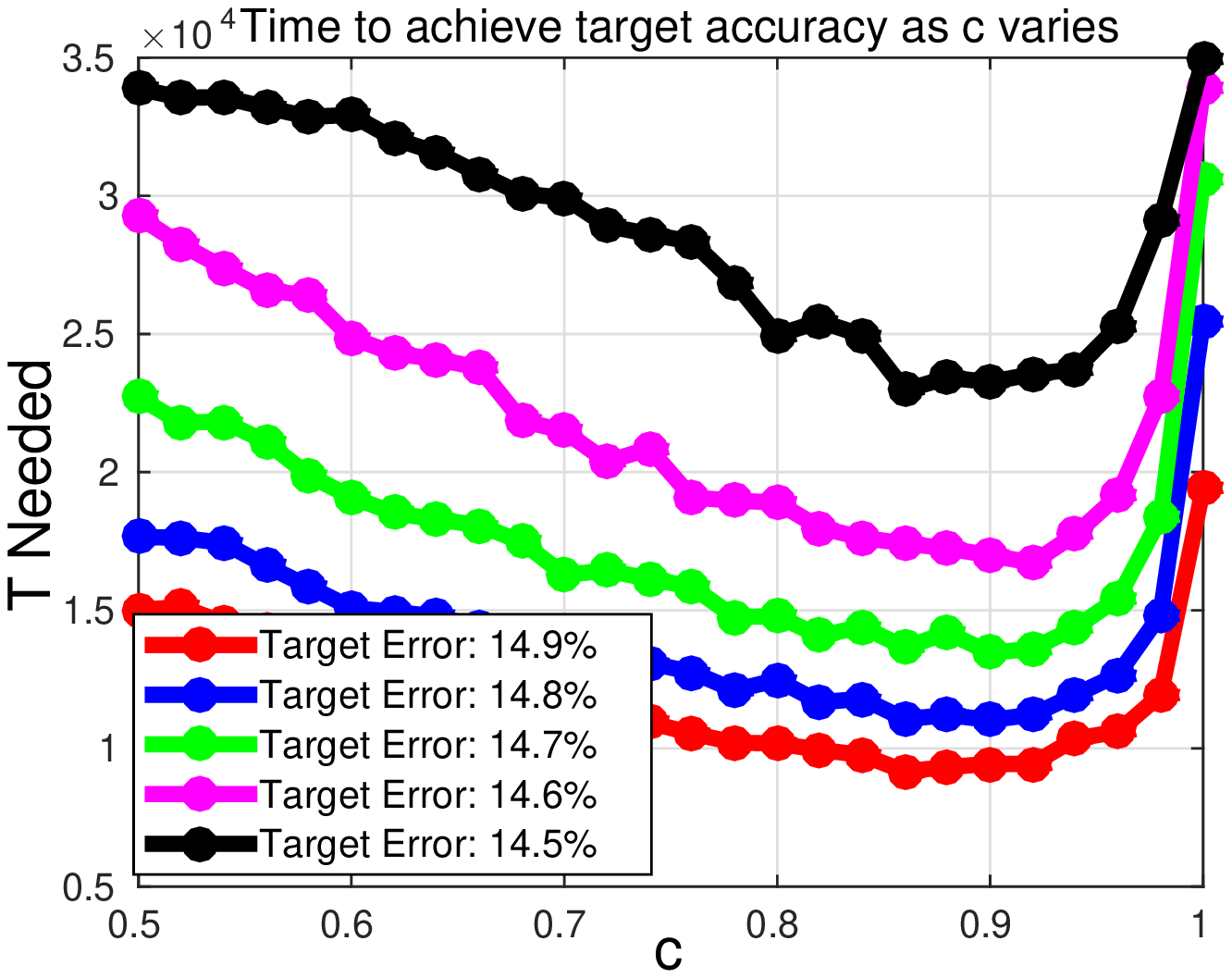}%
\includegraphics[width=0.25 \textwidth]{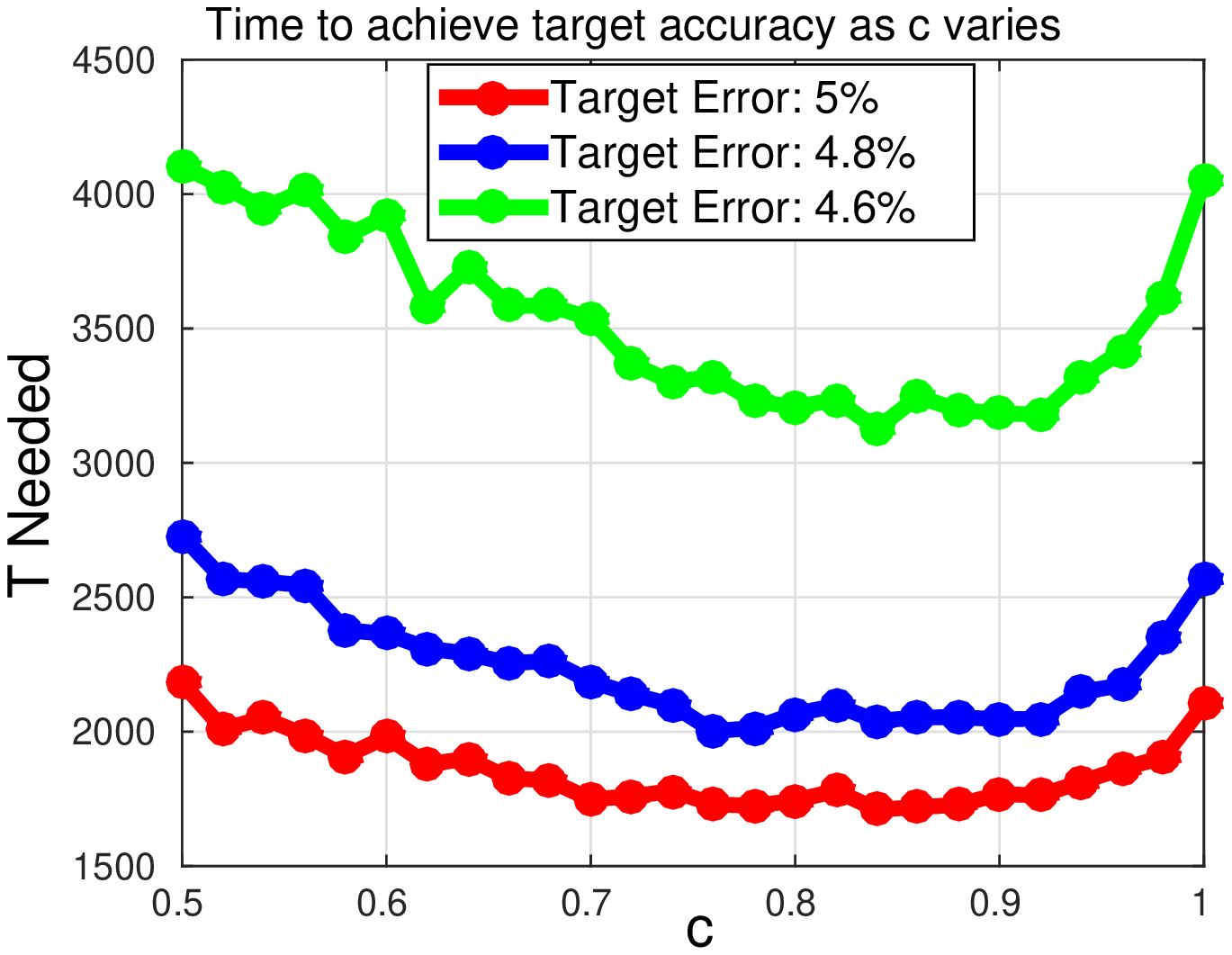}%
\end{center}
\makebox[0.245\textwidth]{\it covtype}
\makebox[0.245\textwidth]{\it ijcnn1}
\makebox[0.245\textwidth]{\it a9a}
\makebox[0.245\textwidth]{\it svmguide1}
\caption{Illustration of the practical significance by choosing the optimal $c$ when SDCA.}
\label{fig:illustration_exp}
\end{figure*}

\begin{figure*}[htpb]
	\begin{center}
		\includegraphics[width=0.25 \textwidth]{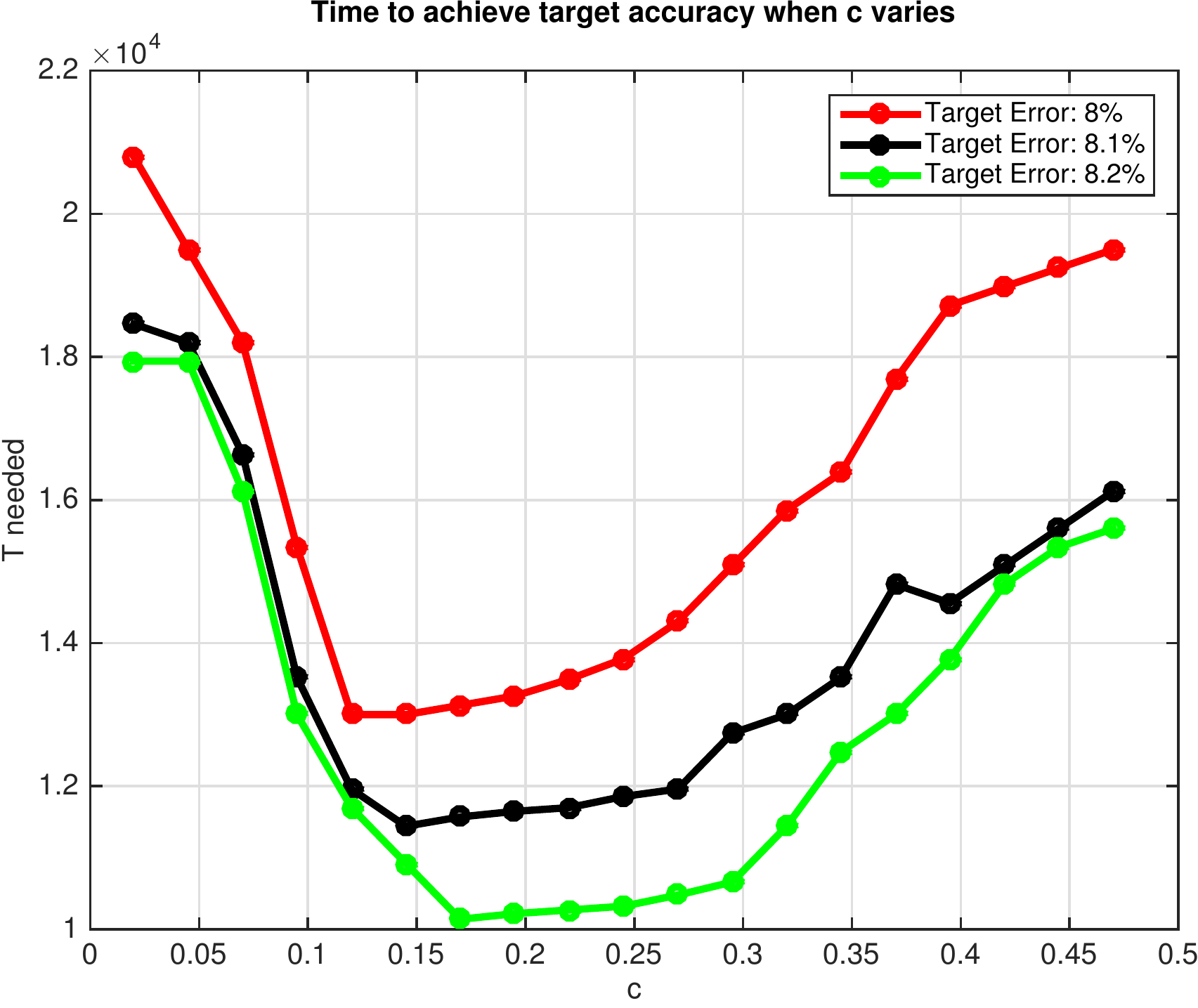}%
		\includegraphics[width=0.25 \textwidth]{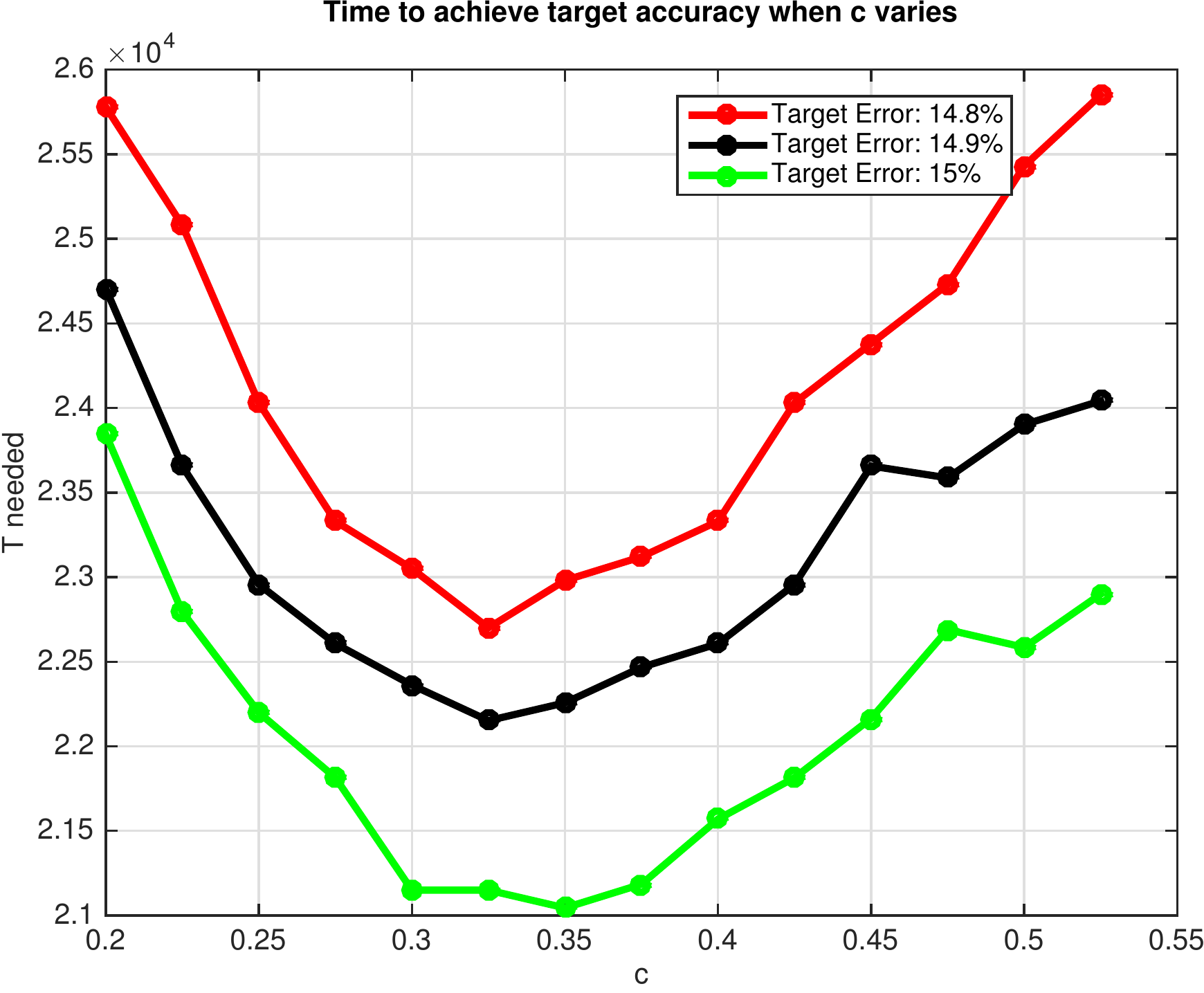}%
		\includegraphics[width=0.25 \textwidth]{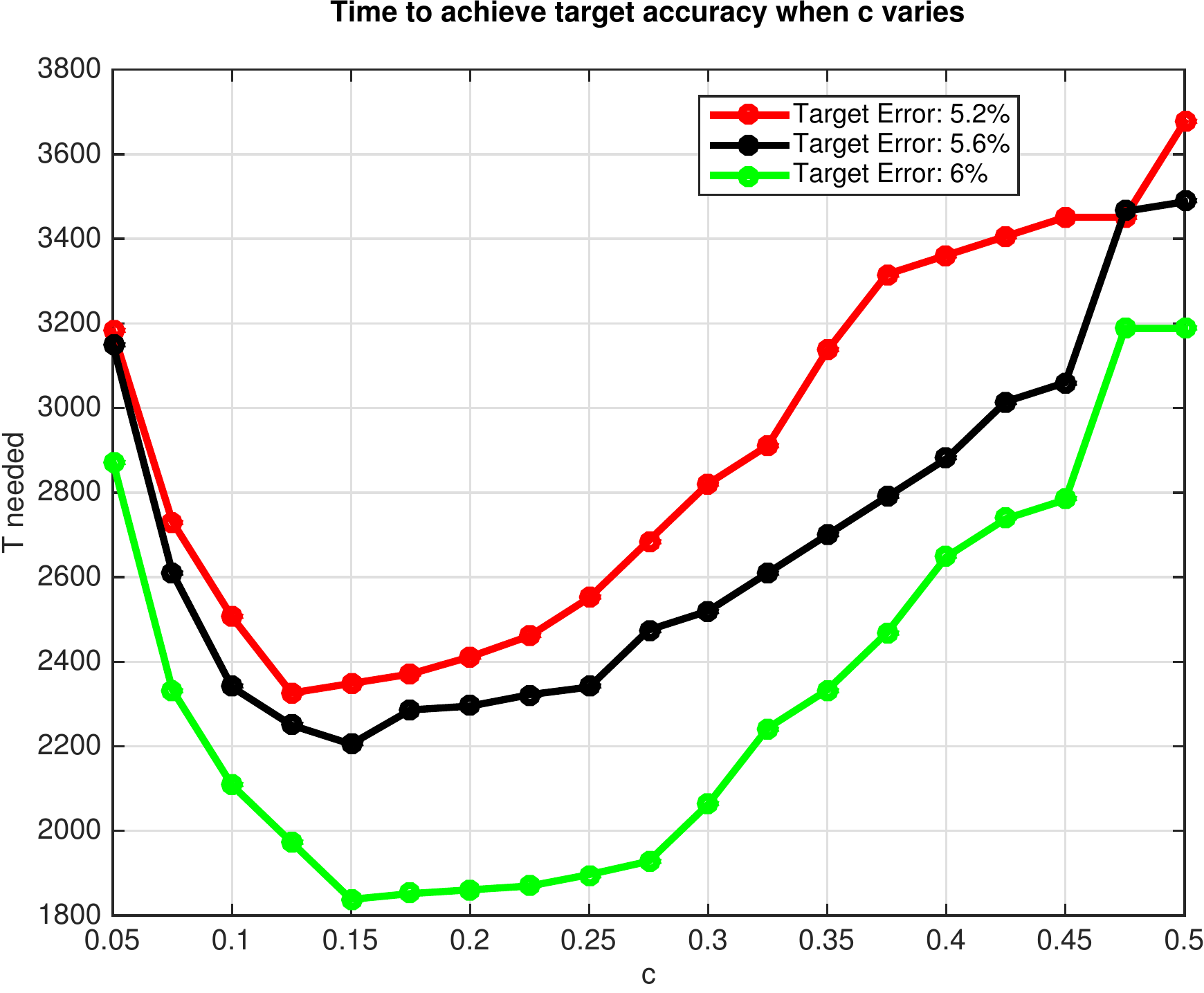}%
		\includegraphics[width=0.25 \textwidth]{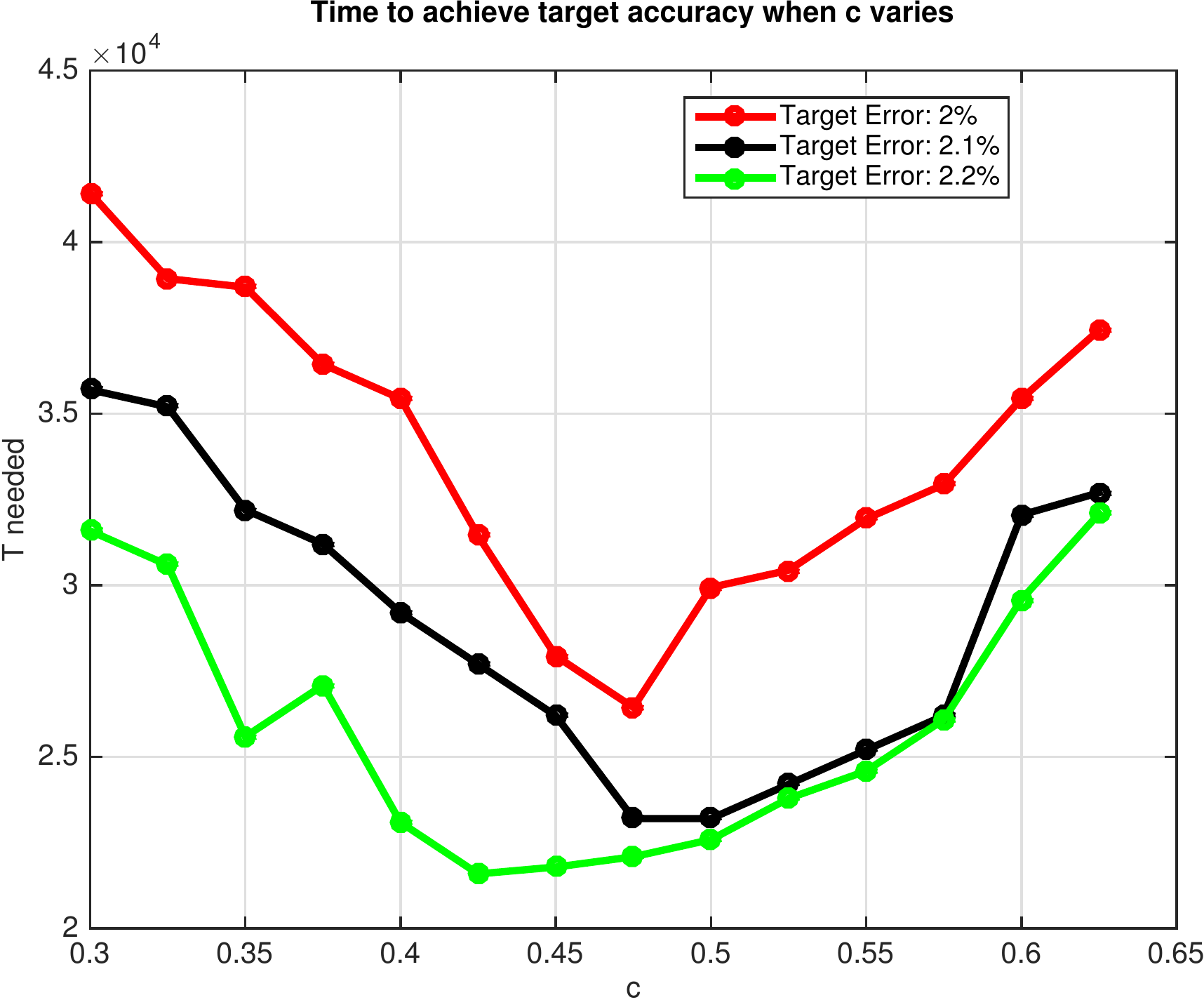}%
	\end{center}
	\makebox[0.245\textwidth]{\it ijcnn1}
	\makebox[0.245\textwidth]{\it a9a}
	\makebox[0.245\textwidth]{\it svmguide1}
	\makebox[0.245\textwidth]{\it w8a}
	\caption{Illustration of the practical significance by choosing the optimal $c$  when SAG.}
	\label{fig:illustration_sag}
\end{figure*}

Let us consider what happens when the estimation error is small. To
be concrete, let us consider a low-dimension problem where $d\ll
\norm{\wref}^2$, though the situation would be similar if for whatever
other reason the estimation error would be lower than its norm-based
upper bound\footnote{This could happen, for example, if low estimation
  error actually happens due to some other low complexity in the
  system, other than a bound on $\norm{\wref}$ and
  $\norm{\xb}$---either the dimensionality of the data, or perhaps the
  intrinsic effective dimensionality, or some combination of norm and
  dimensionality, or even some other norm of the data.  Note that such
  control would have much less of an affect on the optimization, which
  is more tightly tied to the Euclidean norm.}. In $d$ dimensions we
have\footnote{This is the uniform convergence guarantee of {\em
    bounded} functions with pseudo-dimension $d$ \cite{Pollard}. Although
  the hinge loss is not strictly speaking bounded, what we need here
  is only that it is bounded at $\wref$ and $\wb$, which is not
  unreasonable.} $P_m(\wb) - P(\wb) \leq
O\left(\sqrt{\frac{d}{m}}\right)$ yielding:
\begin{equation}
  \label{eq:dbound}
  L(\tilde{\wb}) - L(\wref) \leq  \epsilon(T) + \frac{\lambda}{2} \|\wref\|^2 + O\left(\sqrt{\frac{d}{cT}}\right).
\end{equation}

With SGD, the first two terms still yield
$\Omega(\sqrt{\norm{\wref}/T})$ even with the best $\lambda$, and the
best bound is attained for $c=1$ (although, as observed empirically, a
large range of values of $c$ do not affect performance significantly,
as the first two terms dominate the third, $c$-dependent term).

However, plugging in $\epssdca(T)$, we can use a much smaller
$\lambda= O\left(\sqrt{\frac{d}{cT\norm{\wref}^4}}\right)$ to get:
\begin{align}
  \label{eq:almostfinal}
  L(\tilde{\wb}) - L(\wref) \leq& \exp\left(-T/\left(\sqrt{\frac{cT \norm{\wref}^4}{d}} + cT\right)\right) 
  \nonumber \\
  &+ O\left(\sqrt{\frac{d}{cT}}\right).
\end{align}
As long as $T \geq \Omega\left( \frac{\|\wref\|^4}{d} \right)$, the above is
optimized with $c= \Theta\left( \frac{1}{\sqrt{\log T}} \right)$ and
yields:
\begin{equation}
  \label{eq:finalbound}
  L(\tilde{\wb}) - L(\wref) \leq O\left(\sqrt{\frac{d \log T}{T}}\right).
\end{equation}

This heuristic upper bound analysis suggests that unlike for SGD, when
the estimation error is smaller than its norm-based upper bound, and
we are allowing a large number of iterations $T$, then using a reduced
training set of size $m=cT<T$, with $c<1$, might be beneficial. Figure
\ref{fig:illustration} shows a cartoon of the error decomposition for
SGD and SDCA based on this heuristic analysis. 

What we have done here is revisiting the upper bound SGD analysis and
understand how it might be different for reduced variance methods such
as SDCA and SAG. However, this is still an upper bound analysis based
on heuristic assumptions and estimation error upper bound---a precise
analysis seems beyond reach using current methodology, much in the
same way that we cannot quite analyze why multiple passes of SGD (for
a fixed training set size) are beneficial.

\section{Empirical Investigation}
\label{sec:empirical}

\begin{table*}[t]
	\begin{center}
		
		\begin{tabular}{|c|c|c|c|c|c|c|c|c|c|c|c|}\hline	
			\multirow{2}{*}{$T\backslash$  Dataset}& \multicolumn{3}{|c|}{covtype}& \multicolumn{3}{|c|}{ijcnn1} & \multicolumn{3}{|c|}{w8a}  \\ 
			\cline{2-10}
			
			&{c}&{error}&{error\ (c=1)} &{c}&{error}&{error\ (c=1)} &{c}&{error} &{error\ (c=1)} \\ 
			\hline
			
			1000 & 1.000 & 0.328  &0.328 &0.125& 0.096 & 0.097 &0.575 &0.027 &0.028 \\ 
			2000 & 0.525& 0.307  &0.311 &0.100& 0.092 &0.094 &0.55 & 0.026 &0.026 \\ 
			4000 &0.275&0.288  &0.298 &0.100& 0.089 &0.094 &0.350& 0.025 &0.025 \\ 
			8000 &0.150&0.260  &0.283 &0.075& 0.088 &0.091 &0.275& 0.022 &0.024 \\ 
			16000&0.125&0.250  &0.261 & 0.075& 0.084 & 0.089 &0.300& 0.021 &0.023 \\ 
			32000& 0.150 & 0.242  & 0.251 & 0.025& 0.082 &0.083 &0.225 & 0.018 & 0.020 \\ 
			\hline
			
		\end{tabular}
	\end{center}
	\caption{\small The Optimal $c$ and their test error when using SAG under a time budget with IID sampling.}
	\label{tab:optimalc_sag}
\end{table*}

\begin{table*}[t]
	\begin{center}
		\begin{tabular}{|c|c|c|c|c|c|c|c|c|c|c|c|}\hline	
			\multirow{2}{*}{$T\backslash$  Dataset}& \multicolumn{3}{|c|}{covtype}& \multicolumn{3}{|c|}{ijcnn1} & \multicolumn{3}{|c|}{a9a}  \\ 
			\cline{2-10}

			&{c}&{error}&{error\ (c=0.5)} &{c}&{error}&{error\ (c=0.5)} &{c}&{error} &{error\ (c=0.5)} \\ 
			\hline

			1000   &0.350 & 0.300 & 0.358 &0.350 & 0.082 &0.098 &0.475 & 0.181 &0.193 \\
			2000   &0.400 & 0.278 & 0.344 &0.400 & 0.072 &0.091 &0.475 & 0.178 &0.188 \\
			4000   &0.325 & 0.264 & 0.331 &0.475 & 0.070 &0.087 &0.450 & 0.170 &0.180 \\
			8000   &0.450 & 0.256 & 0.310 &0.425 & 0.068 &0.083 &0.475 & 0.170 &0.182 \\
			16000  &0.475 & 0.253 & 0.297 &0.350 & 0.066 &0.084 &0.450 & 0.166 &0.177 \\
			32000  &0.425 & 0.252 & 0.281 &0.375 & 0.066 &0.083 &0.425 & 0.169 &0.175 \\
			
			\hline
		\end{tabular}
	\end{center}
	\caption{\small The Optimal $c$ when using SVRG under a time budget with IID sampling.}
	\label{tab:optimalc_svrg_iid}
\end{table*}

\begin{table*}[htpb]
	\begin{center}
		\begin{tabular}{|c|c|c|c|c|c|c|c|c|c|c|c|}\hline	
			\multirow{2}{*}{$T\backslash$  Dataset}& \multicolumn{3}{|c|}{covtype}& \multicolumn{3}{|c|}{ijcnn1} & \multicolumn{3}{|c|}{a9a}  \\ 
			\cline{2-10}

			&{c}&{error}&{error\ (c=0.5)} &{c}&{error}&{error\ (c=0.5)} &{c}&{error} &{error\ (c=0.5)} \\ 
			\hline

			1000 &  0.325 & 0.293  &0.360 &0.350& 0.081 & 0.094 & 0.475& 0.178 &0.190\\ 
			2000 &  0.425 &0.272 &  0.340 &0.425& 0.072 & 0.091 & 0.450& 0.176 &0.186\\ 
			4000 &  0.275 & 0.263  &0.330 &0.450& 0.071 & 0.087 & 0.475& 0.168 &0.179\\ 
			8000 &  0.450& 0.256  & 0.310 &0.400& 0.067 & 0.085 & 0.450& 0.169 &0.177\\ 
			16000&  0.450& 0.252  & 0.301 &0.475& 0.066 & 0.082 & 0.475& 0.165 &0.178\\ 
			32000&  0.350& 0.251 & 0.289  &0.350& 0.066 & 0.082 & 0.350& 0.168 &0.172\\ 
			\hline
		\end{tabular}
	\end{center}
	\caption{The Optimal $c$ when using SVRG under a time budget with permutation.}
	\label{tab:optimalc_svrg_perm}
\end{table*}

To investigate the benefit of using a reduced training set
empirically, we conducted experiments with SDCA, SAG and SVRG (and
also SGD/Peagsos) on the five datasets described in Table
\ref{tab:datasetsats}, downloaded from the LIBSVM website
\cite{libsvm}. We first fixed the time budget $T$, and randomly
sampled $cT$ instances from the data pool. Then we ran SGD, SDCA, SAG
and SVRG with $T$ iterations on the $cT$ sample, and tested the
classification performance in an unseen test dataset which consists
$30\%$ of the total instances. A value of $c=1$ corresponds to using
all fresh samples, while with $c<1$ we reuse some samples.  We tried
$c = \{0.025,0.05,0.075,...0.975,1\}$, and for every setting of $c$
and $T$, we follow the same protocol that optimizing $\lambda$ to
achieve the best prediction performance on test dataset (following
\citet{mdlw}). For the all these algorithms, we tried both i.i.d
sampling (with replacement), as well as using a (fresh) random
permutation over the training set in each epoch, thus avoiding
repeated samples inside an epoch. Although most theoretical guarantees
are for i.i.d.~sampling, Such random-permutation sampling is unknown
to typically converge faster than i.i.d sampling and is often used in
practice (see \citealt{recht2012beneath,gurbuzbalaban2015random} for
recent attempts at analyzing random permutation sampling).  All
datasets are prepared for binary classification problem and we used
the smoothed hinge loss. To overcome randomness, we repeat our
experiments $500$ times and report the average classification error.
\footnote{In both SAG and SVRG algorithm, a constant stepsize is used
  in different iterations.  To obtain the best performance, we tune
  the stepsize for each dataset and $T$ combination.  In SVRG, one
  pass SGD is used to initialize, and we set $m=2n$.}

The results with SDCA, SAG and SVRG are shown in Figure
\ref{fig:testerror1} (see also additional plots in appendix), where we plot the test error as a function of
the parameter $c$ (training set size as ratio of number of
iterations), while fix the time budget (number of iterations) $T$, and in Table
\ref{tab:optimalc}, \ref{tab:optimalc_sag}, \ref{tab:optimalc_svrg_iid}, \ref{tab:optimalc_svrg_perm} where we summarize the optimal $c$. On all
datasets, the optimal $c$ for large enough
$T$ is less than $1$. The advantage of using $c<1$ and resampling
data is more significant on covtype and svmguide1, which are both low
dimensional, matching the theory.

Another way of looking at the same results is asking ``what is the
runtime required to achieve a certain target accuracy?''. For various
target accuracies and each value of $c$, we plot in Figure \ref{fig:illustration_exp}, \ref{fig:illustration_sag}
the minimal $T$ such that using $cT$ samples and $T$ iterations
achieves the desired accuracy. Viewed this way, we see how using less
data can indeed reduce runtime.

In SDCA, both with i.i.d and random permutation sampling we often benefit from
$c<1$. Not surprising, sampling ``without replacement'' (random
permutation sampling) is generally better. But the behavior for
random permutation sampling is particularly peculiar, with the optimal
$c$ always very close to 1, and with multi-modal behavior with modes
in inverse integers, $c=1,1/2,1/3,1/4,\ldots$. To understand this
better, we looked more carefully at the behavior of SDCA iterations.

\section{A Closer Look at SDCA-Perm}
\label{sec:closer}

\begin{figure*}[t]
\begin{center}
\includegraphics[width=0.25 \textwidth]{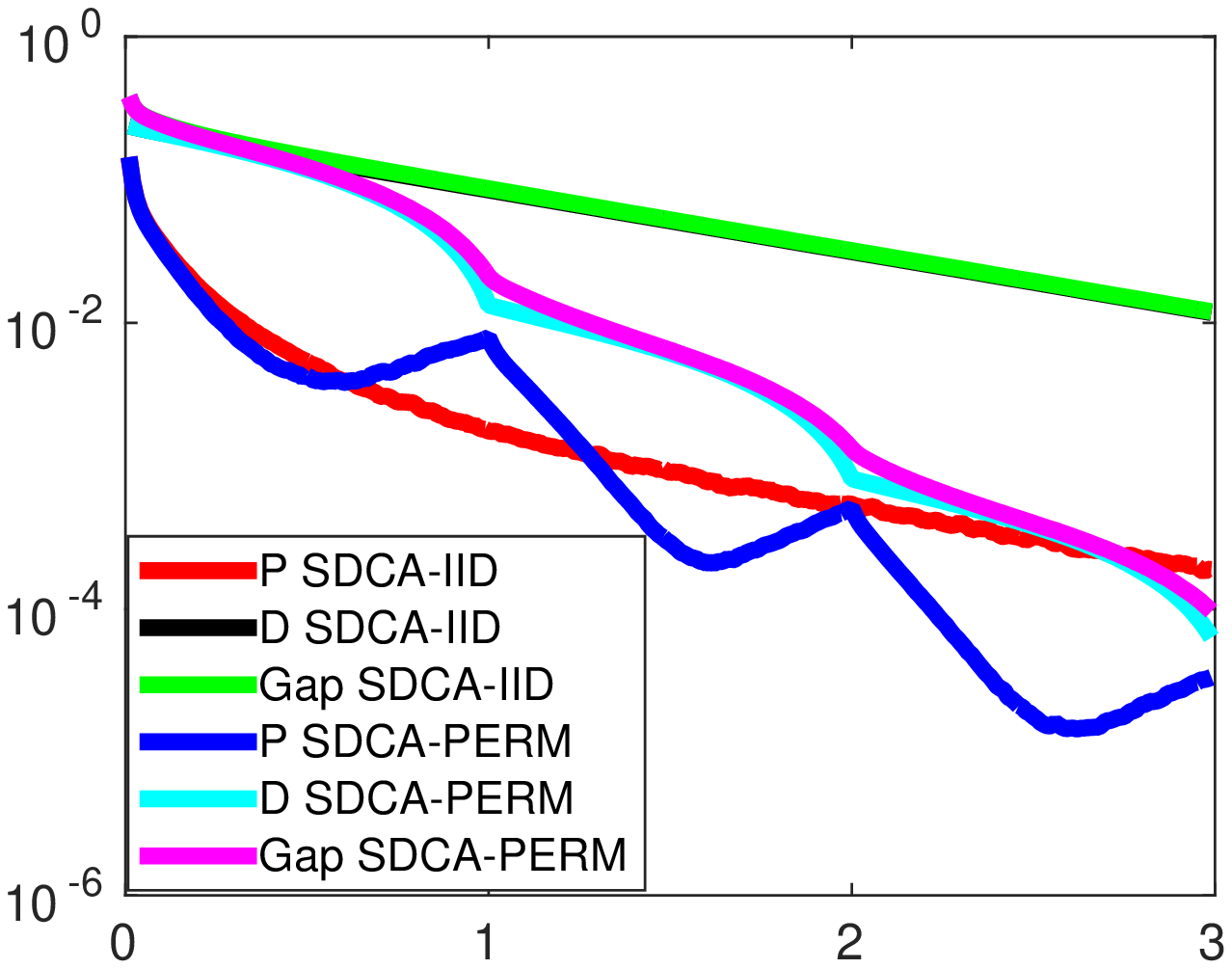}%
\includegraphics[width=0.25 \textwidth]{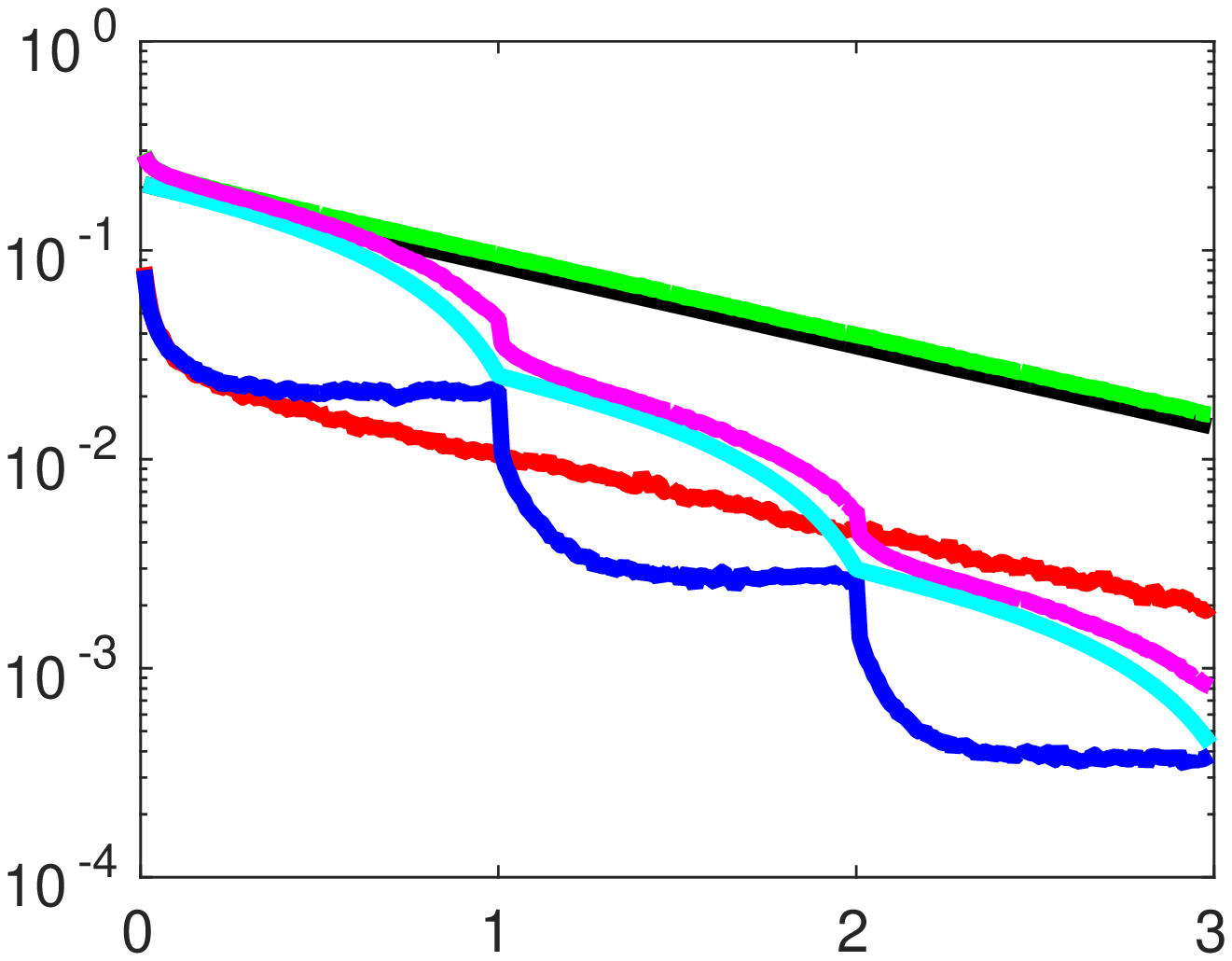}%
\includegraphics[width=0.25 \textwidth]{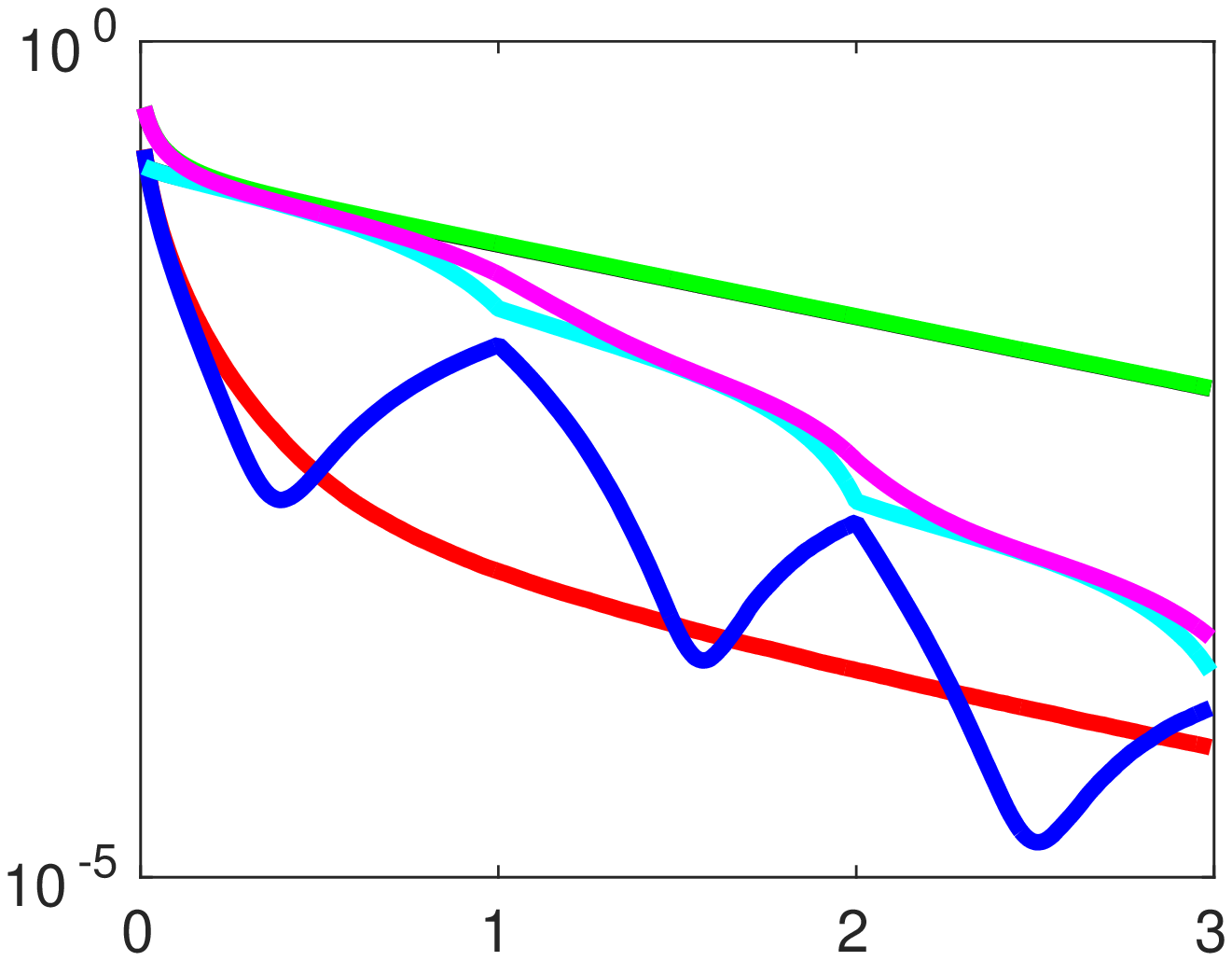}%
\includegraphics[width=0.25 \textwidth]{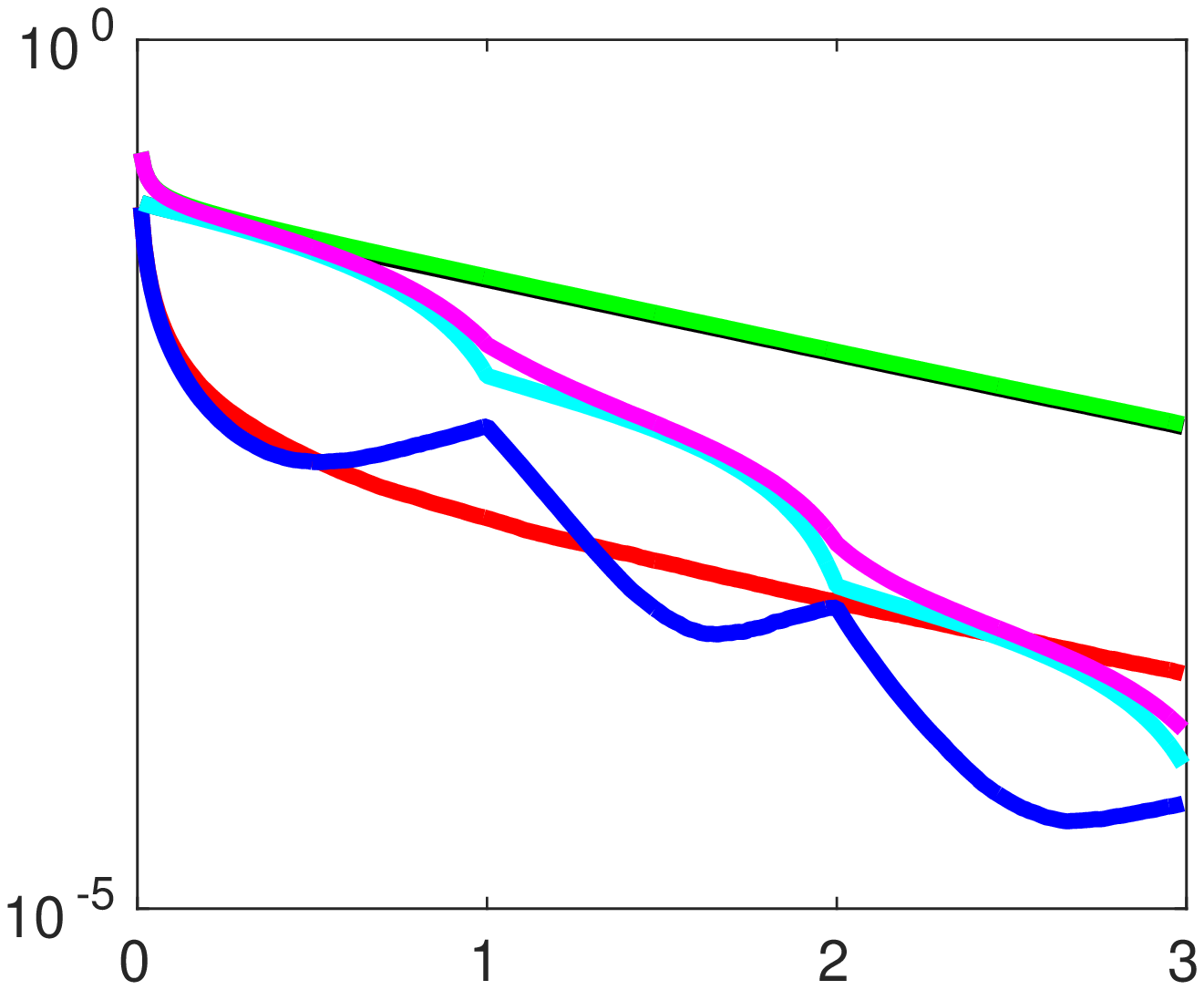}%
\end{center}
\makebox[0.245 \textwidth]{{\it a9a, $\lambda$:0.1}}
\makebox[0.245 \textwidth]{{\it a9a, $\lambda$:0.01}}
\makebox[0.245 \textwidth]{{\it w8a, $\lambda$:0.1}}
\makebox[0.245 \textwidth]{{\it w8a, $\lambda$:0.01}}
\begin{center}
\includegraphics[width=0.25 \textwidth]{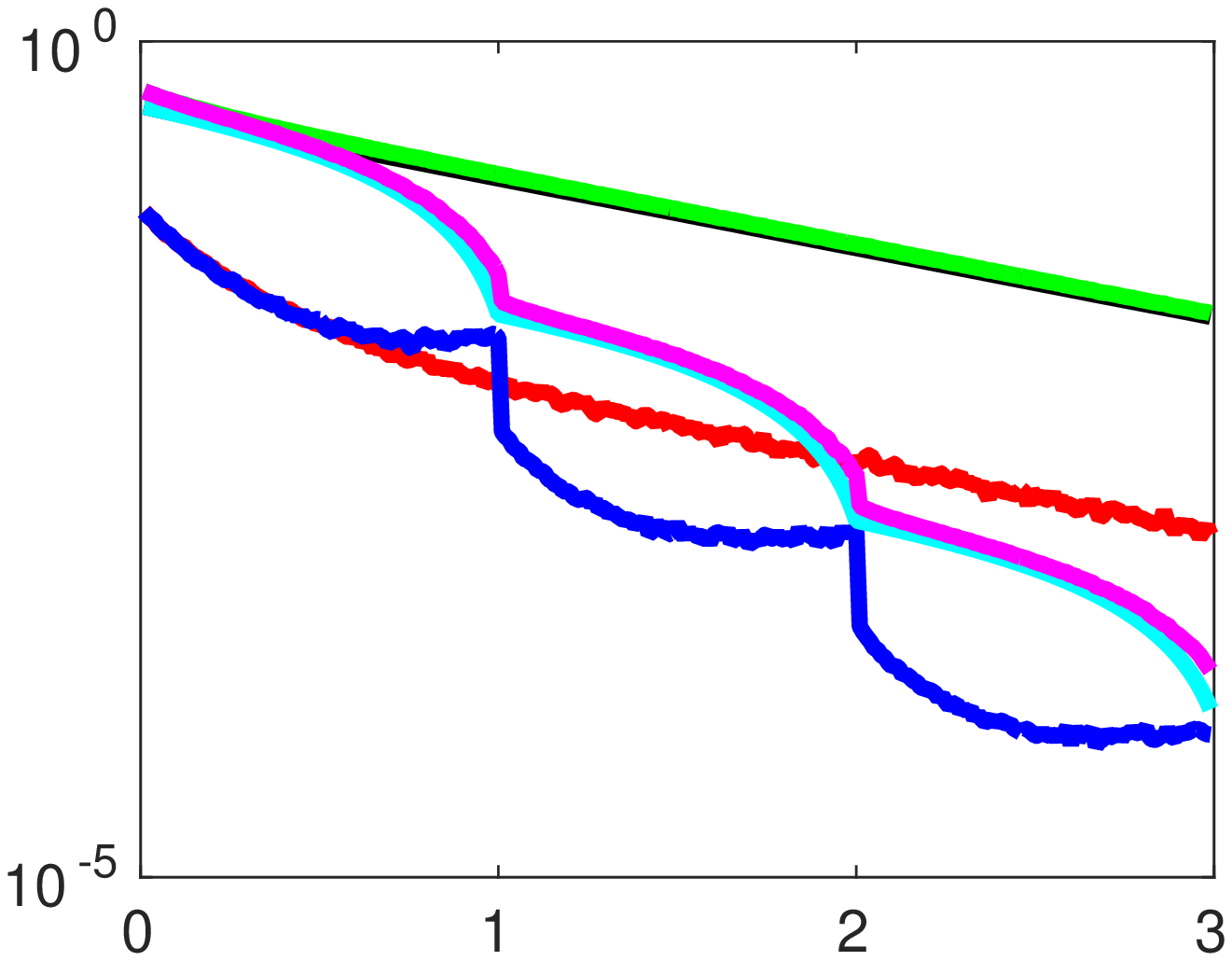}%
\includegraphics[width=0.25 \textwidth]{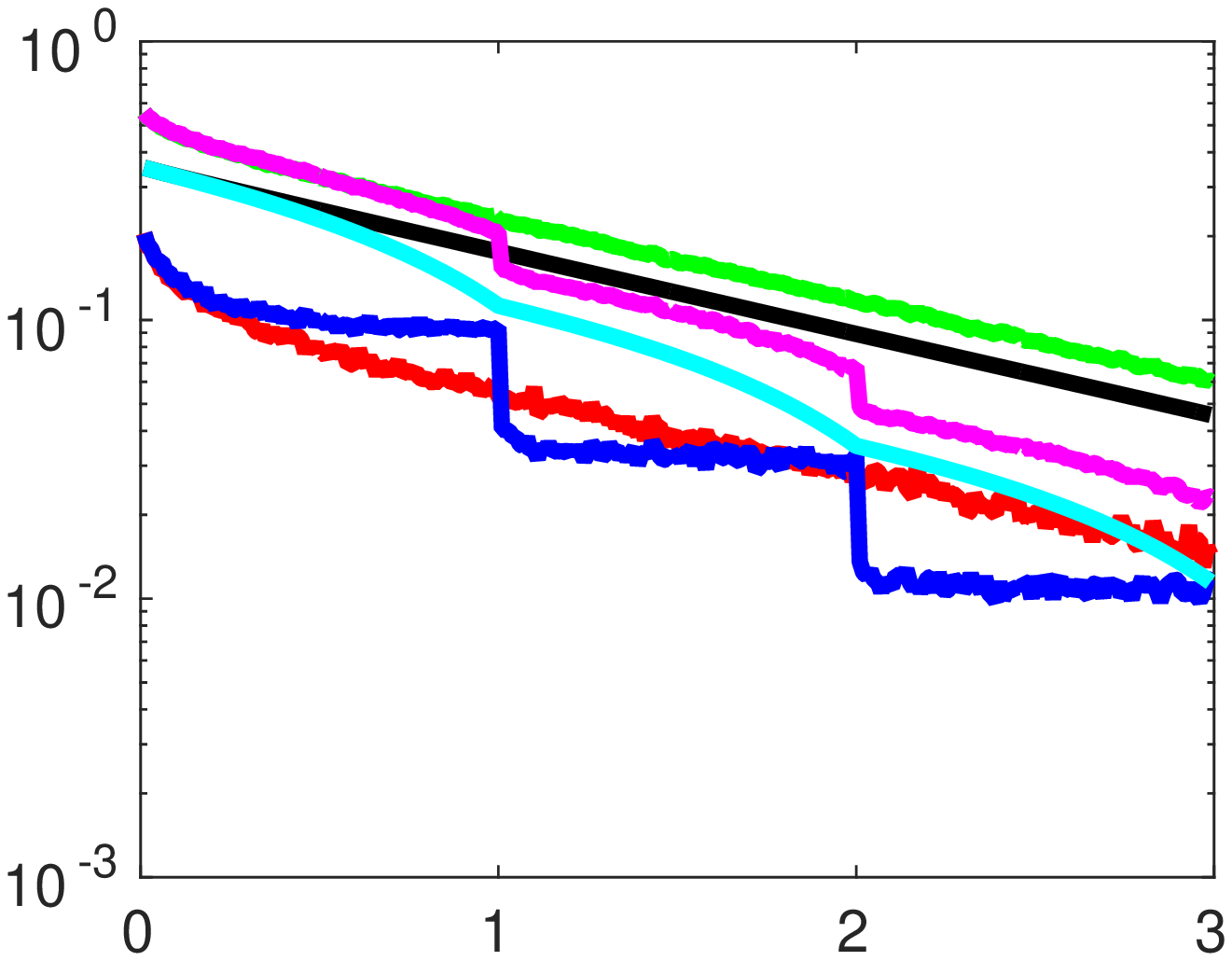}%
\includegraphics[width=0.25 \textwidth]{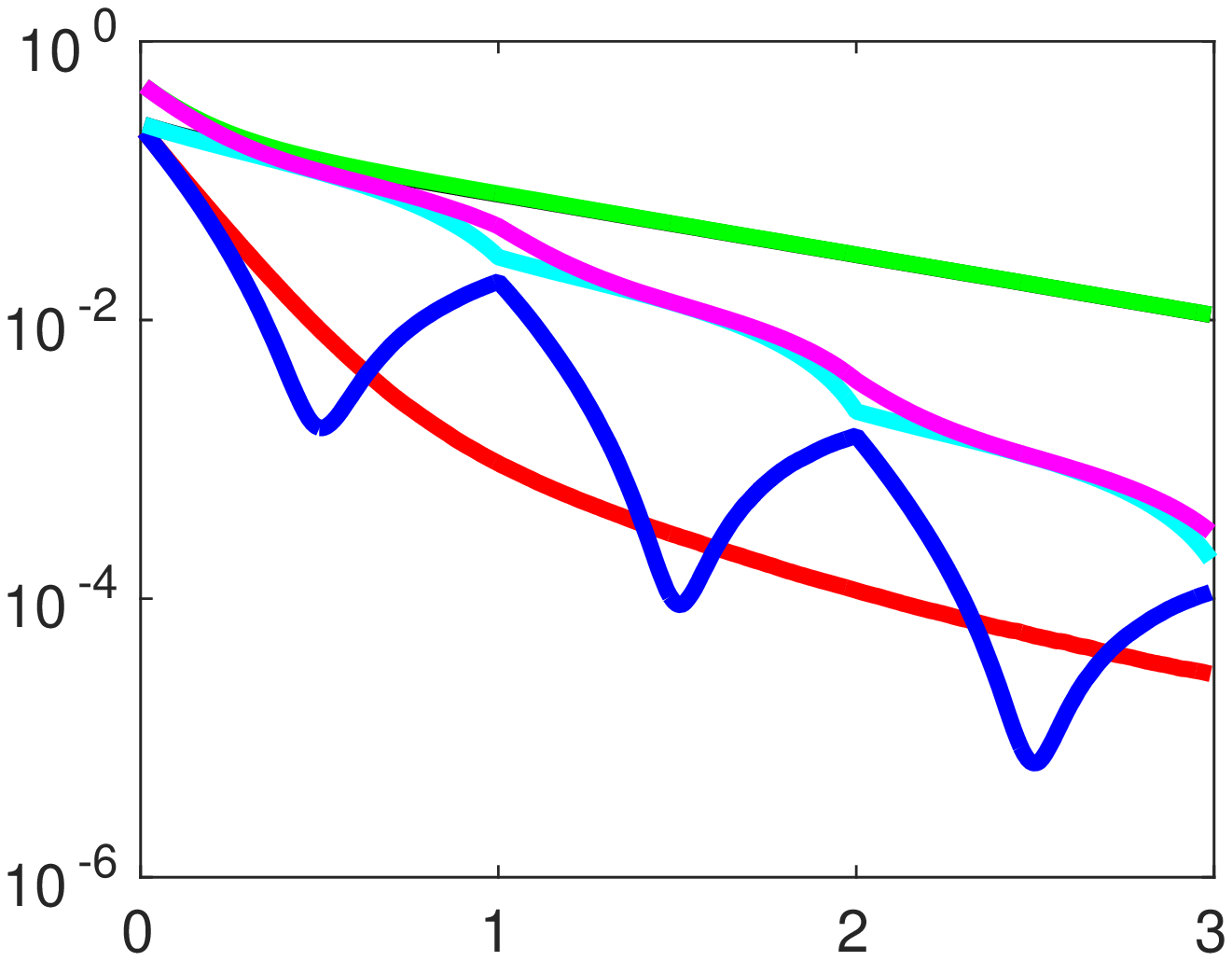}%
\includegraphics[width=0.25 \textwidth]{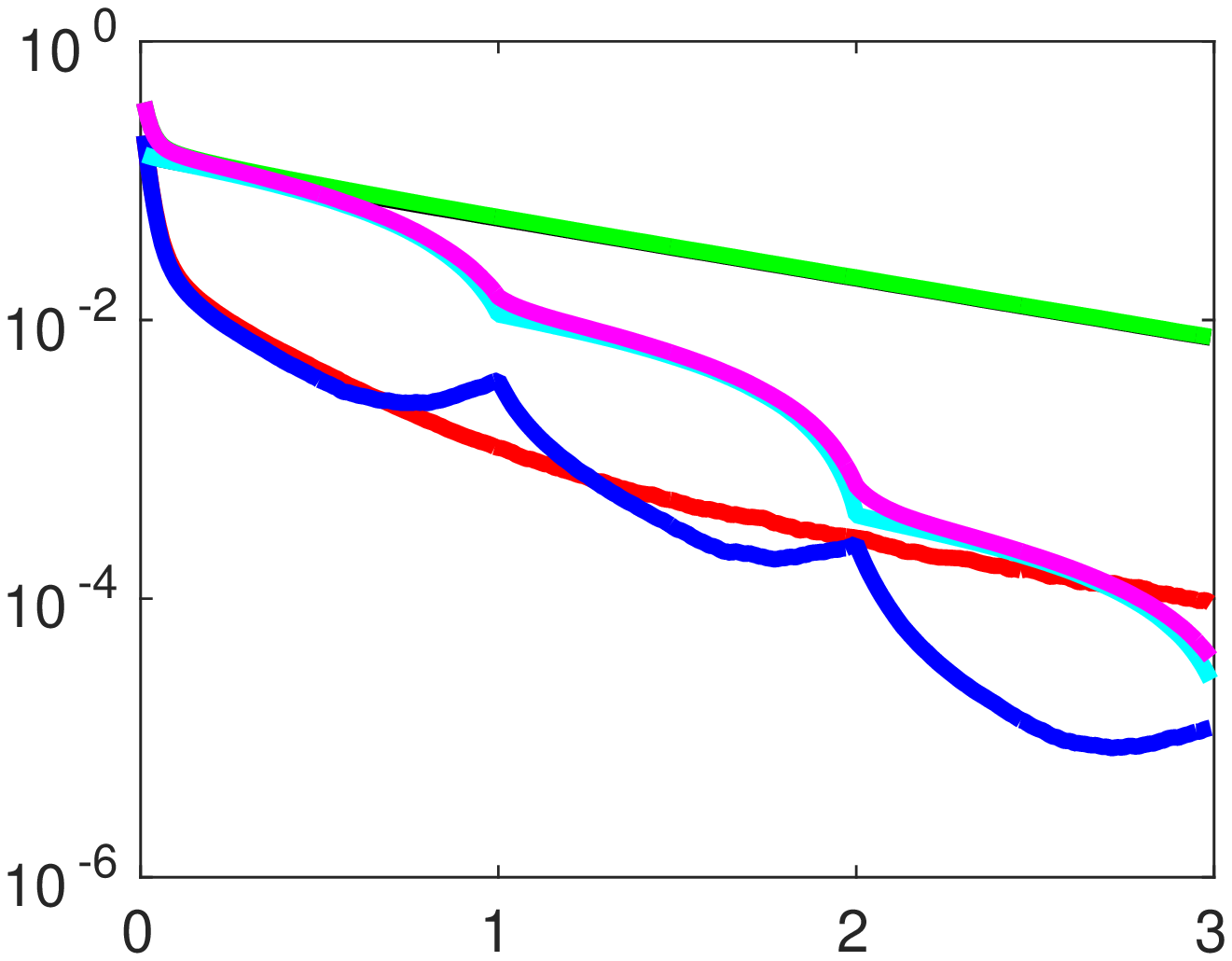}%
\end{center}
\makebox[0.245 \textwidth]{{\it covtype, $\lambda$:0.1}}
\makebox[0.245 \textwidth]{{\it covtype, $\lambda$:0.01}}
\makebox[0.245 \textwidth]{{\it ijcnn1, $\lambda$:0.1}}
\makebox[0.245 \textwidth]{{\it ijcnn1, $\lambda$:0.01}}
\caption{The convergence behavior of SDCA-Perm}
\label{fig:convergence}
\end{figure*}

\begin{figure*}[t]
\begin{center}
\includegraphics[width=0.25 \textwidth]{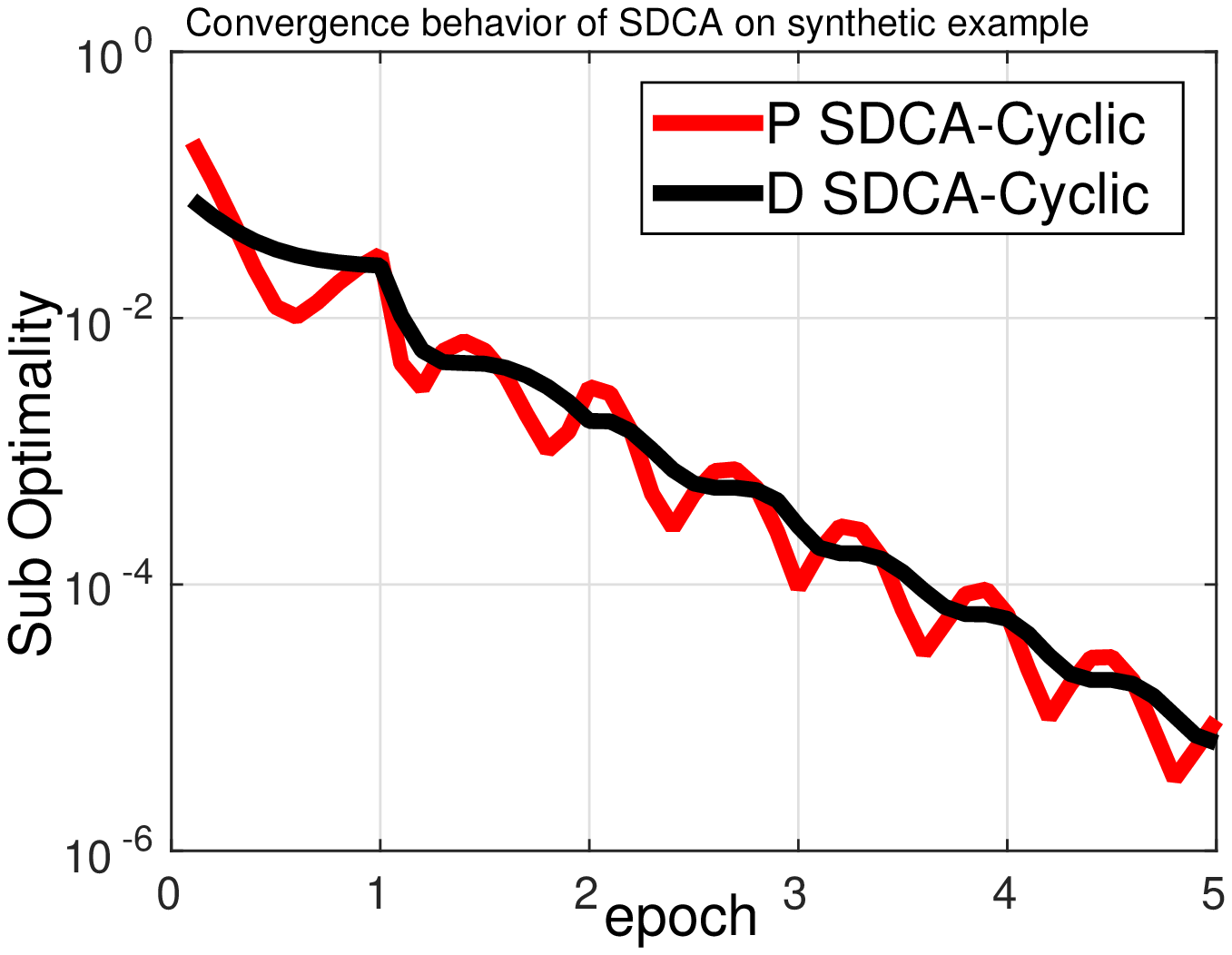}%
\includegraphics[width=0.25 \textwidth]{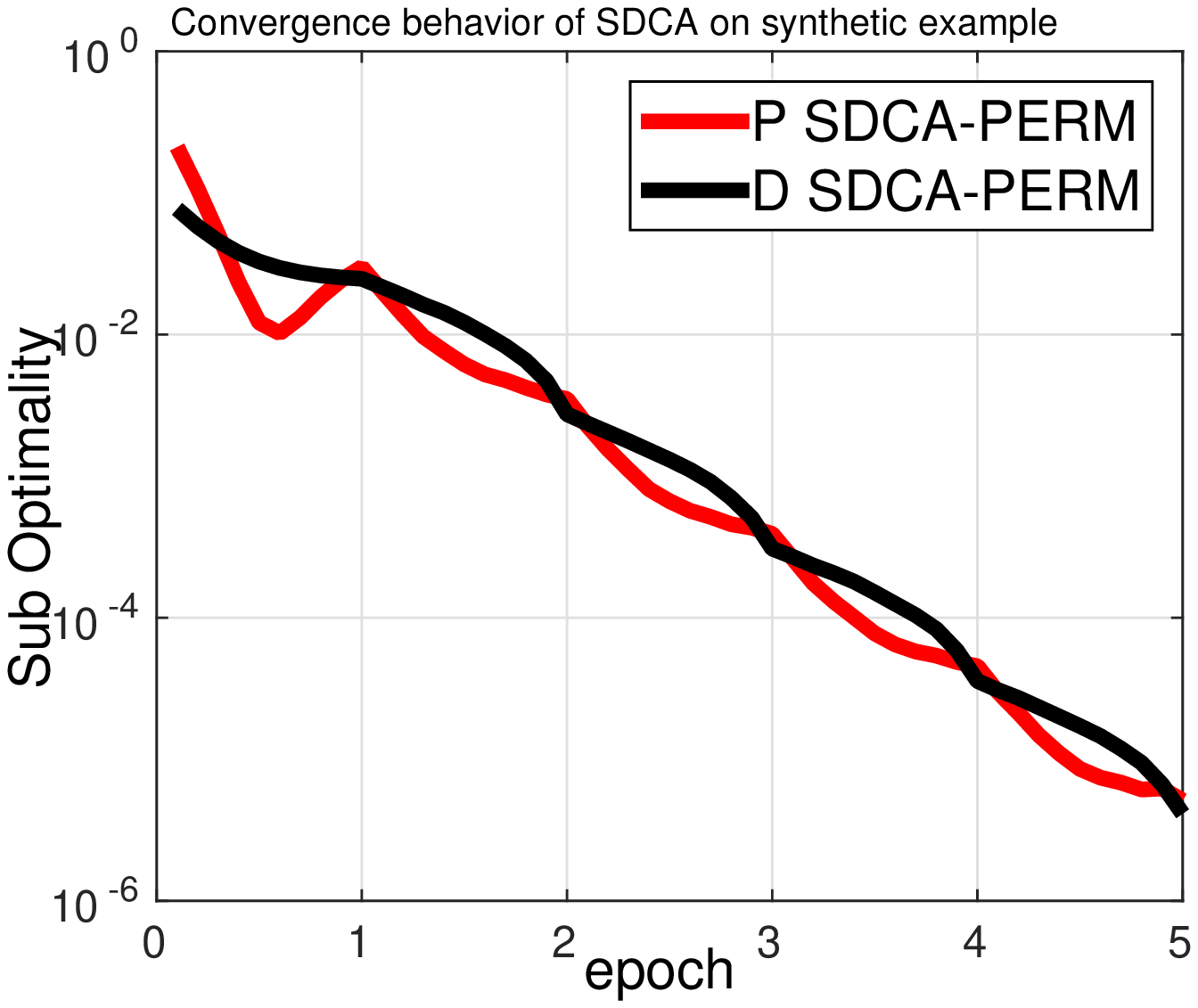}%
\includegraphics[width=0.25 \textwidth]{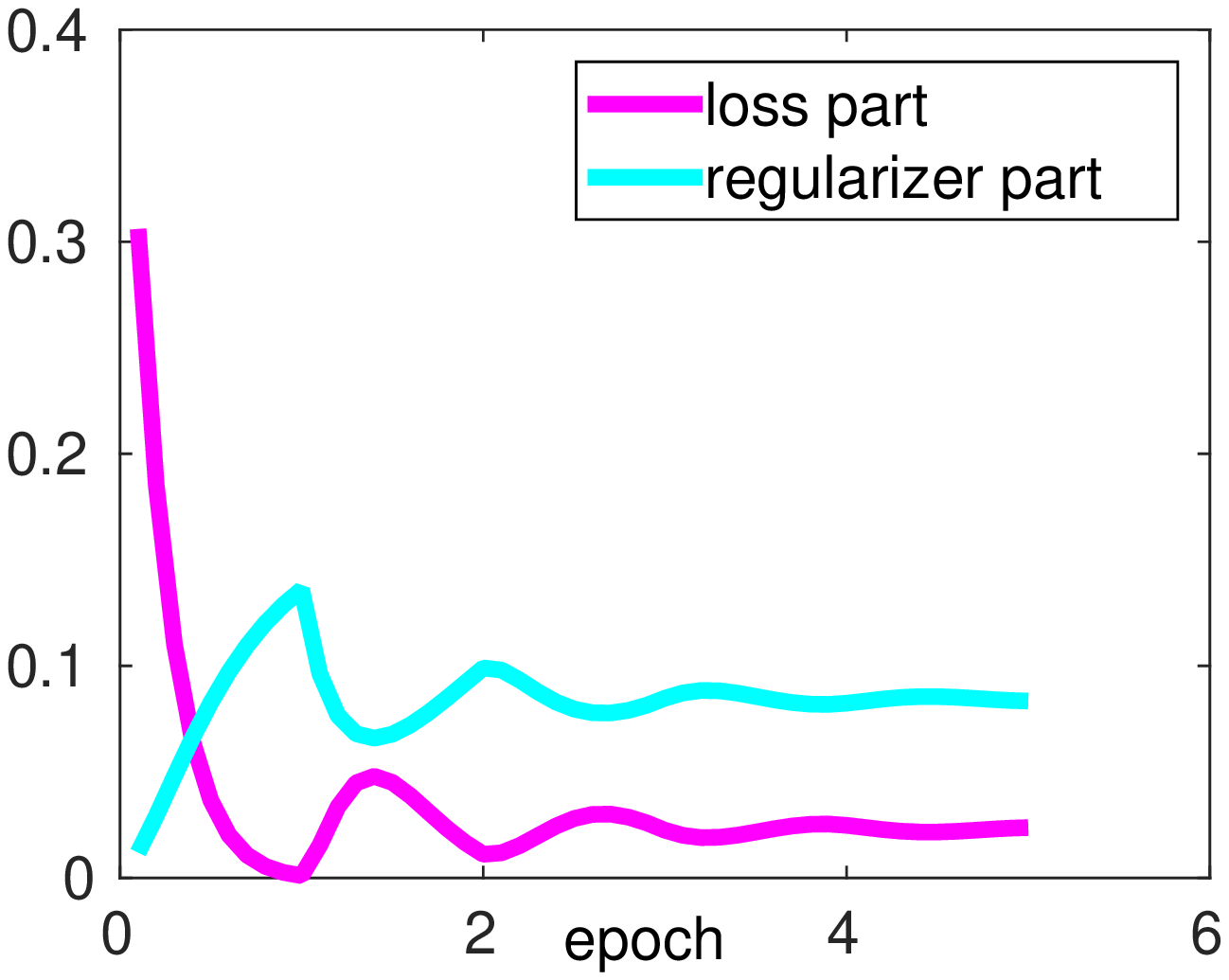}%
\includegraphics[width=0.25 \textwidth]{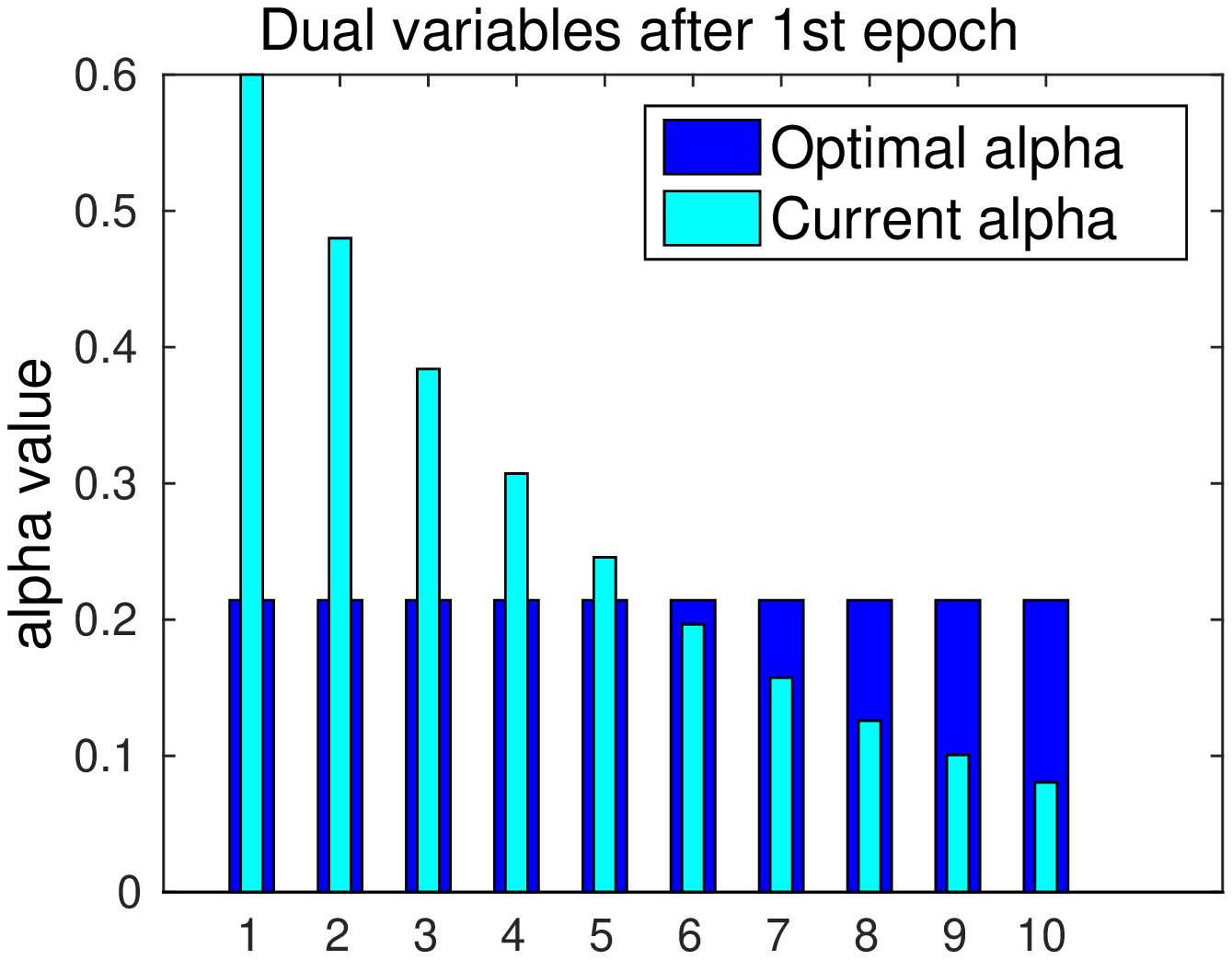}%
\end{center}
\makebox[0.25 \textwidth]{(a)}\makebox[0.25 \textwidth]{(b)}\makebox[0.25 \textwidth]{(c)}\makebox[0.25 \textwidth]{(d)}
\begin{center}
\includegraphics[width=0.25 \textwidth]{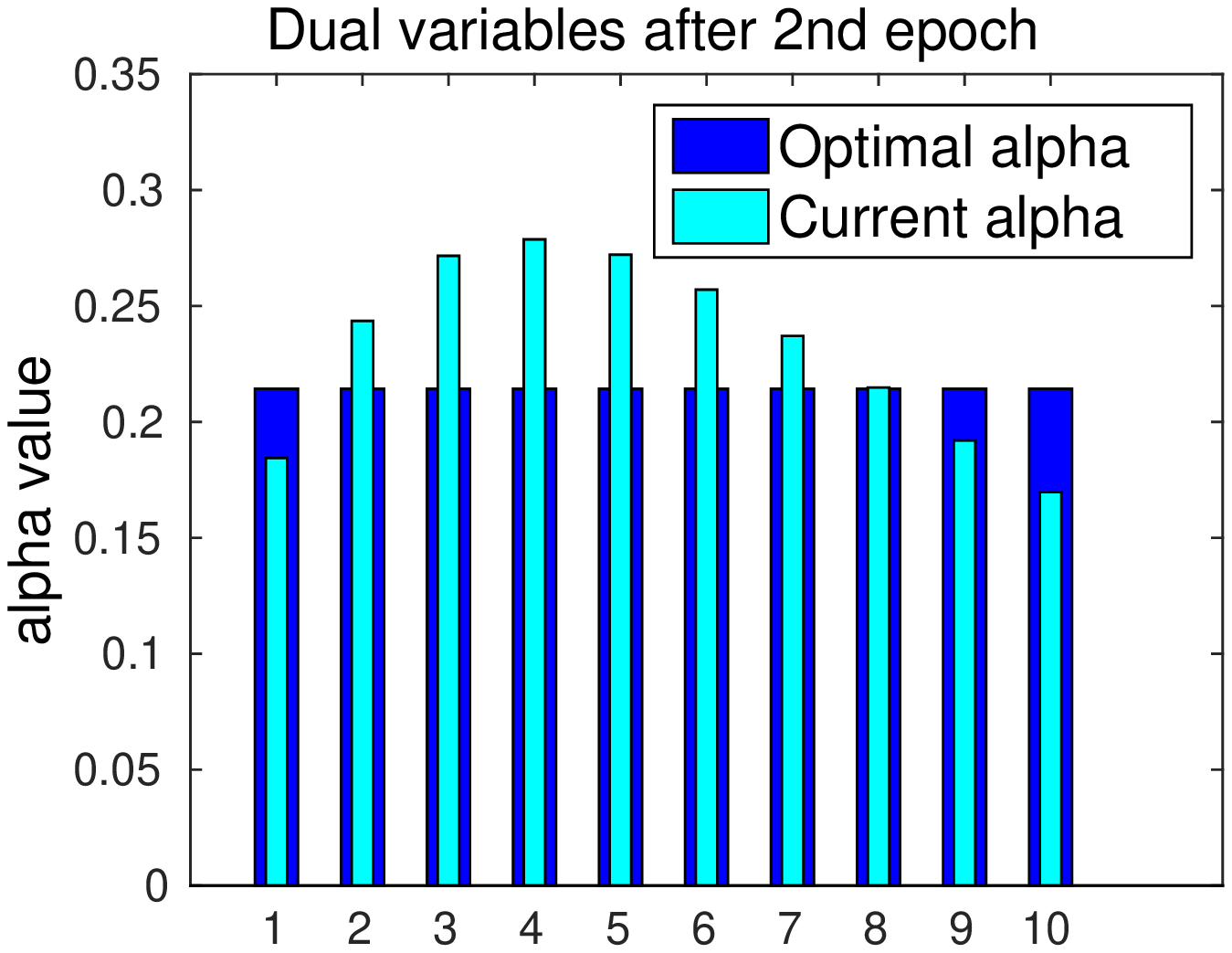}%
\includegraphics[width=0.25 \textwidth]{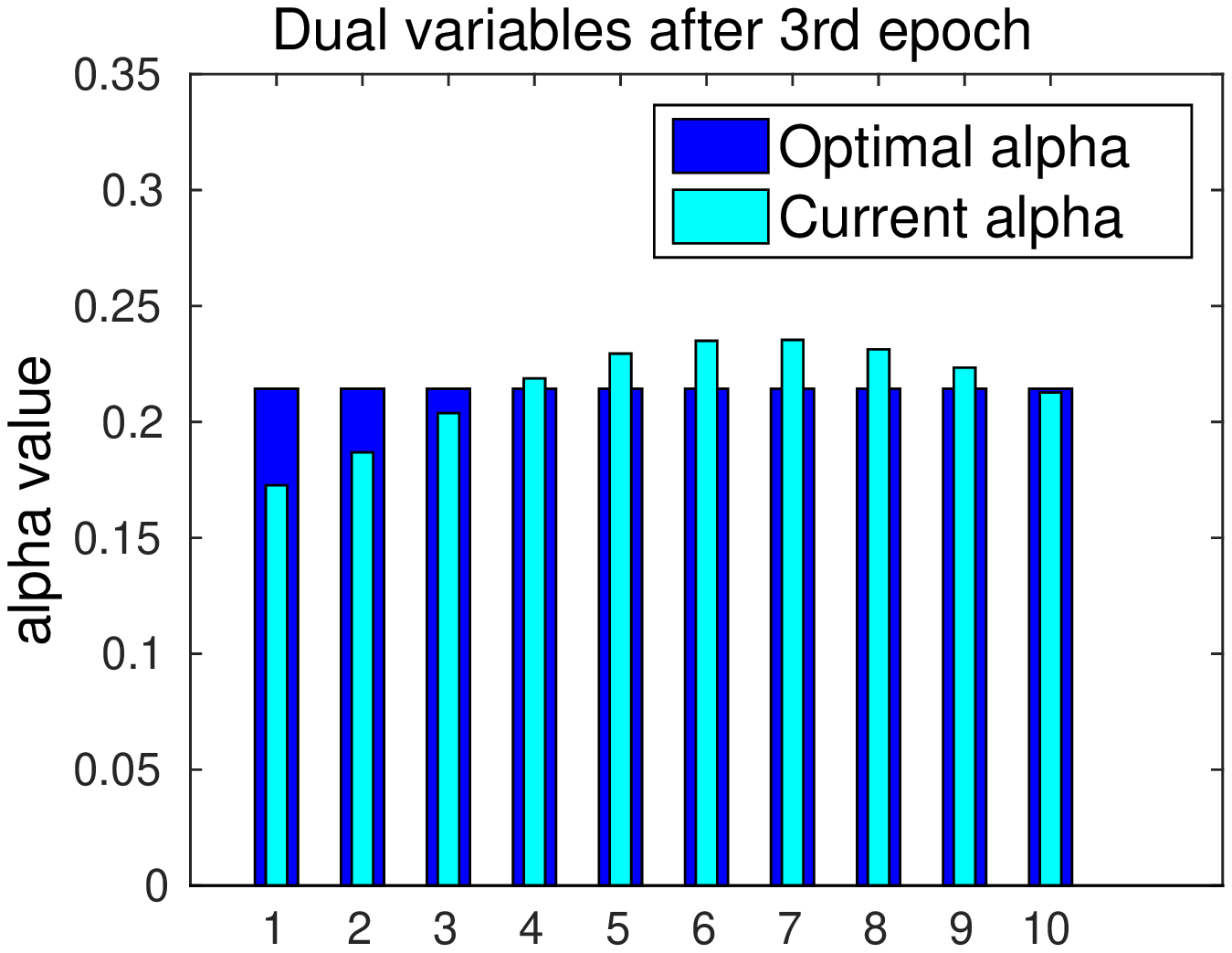}%
\includegraphics[width=0.25 \textwidth]{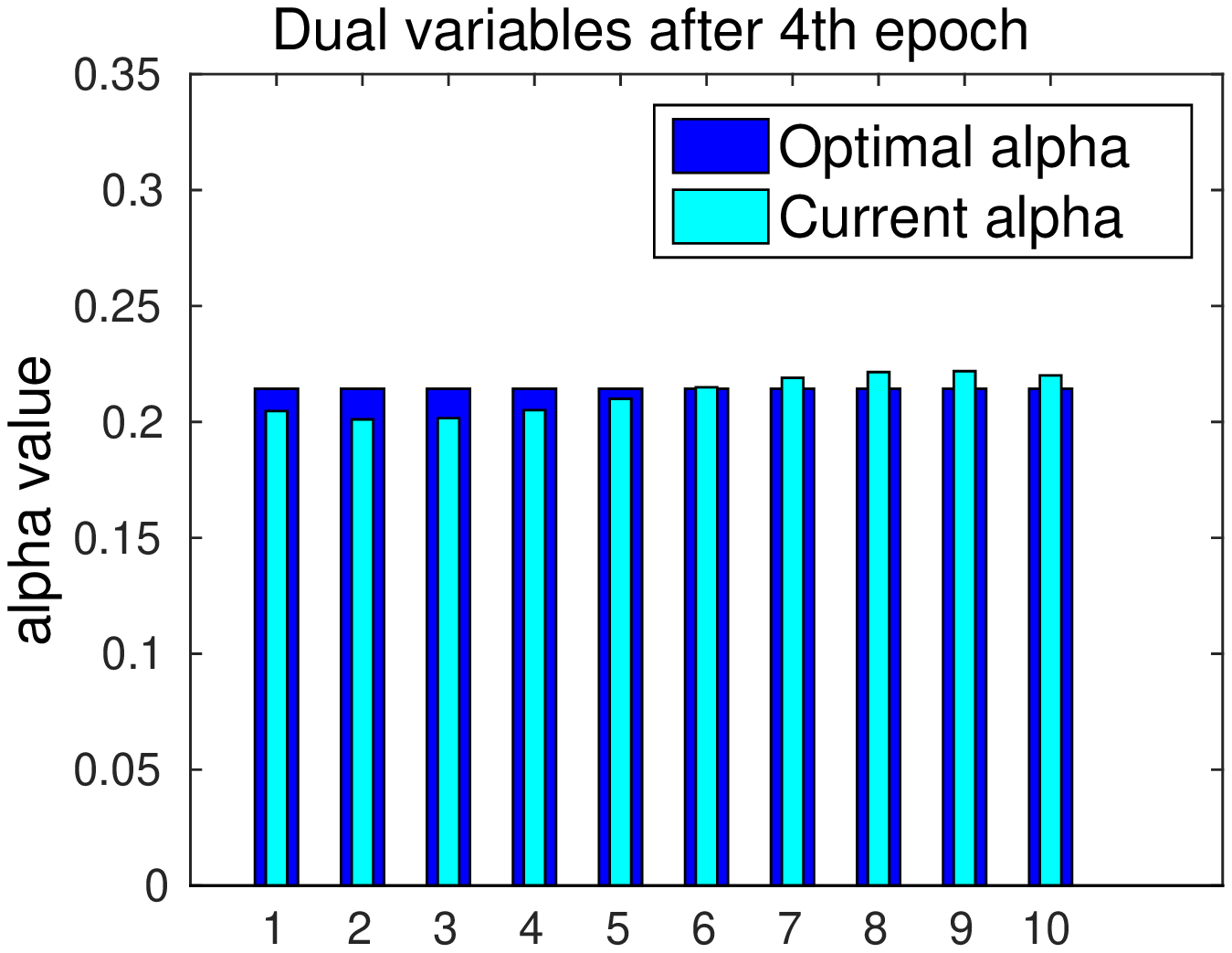}%
\includegraphics[width=0.25 \textwidth]{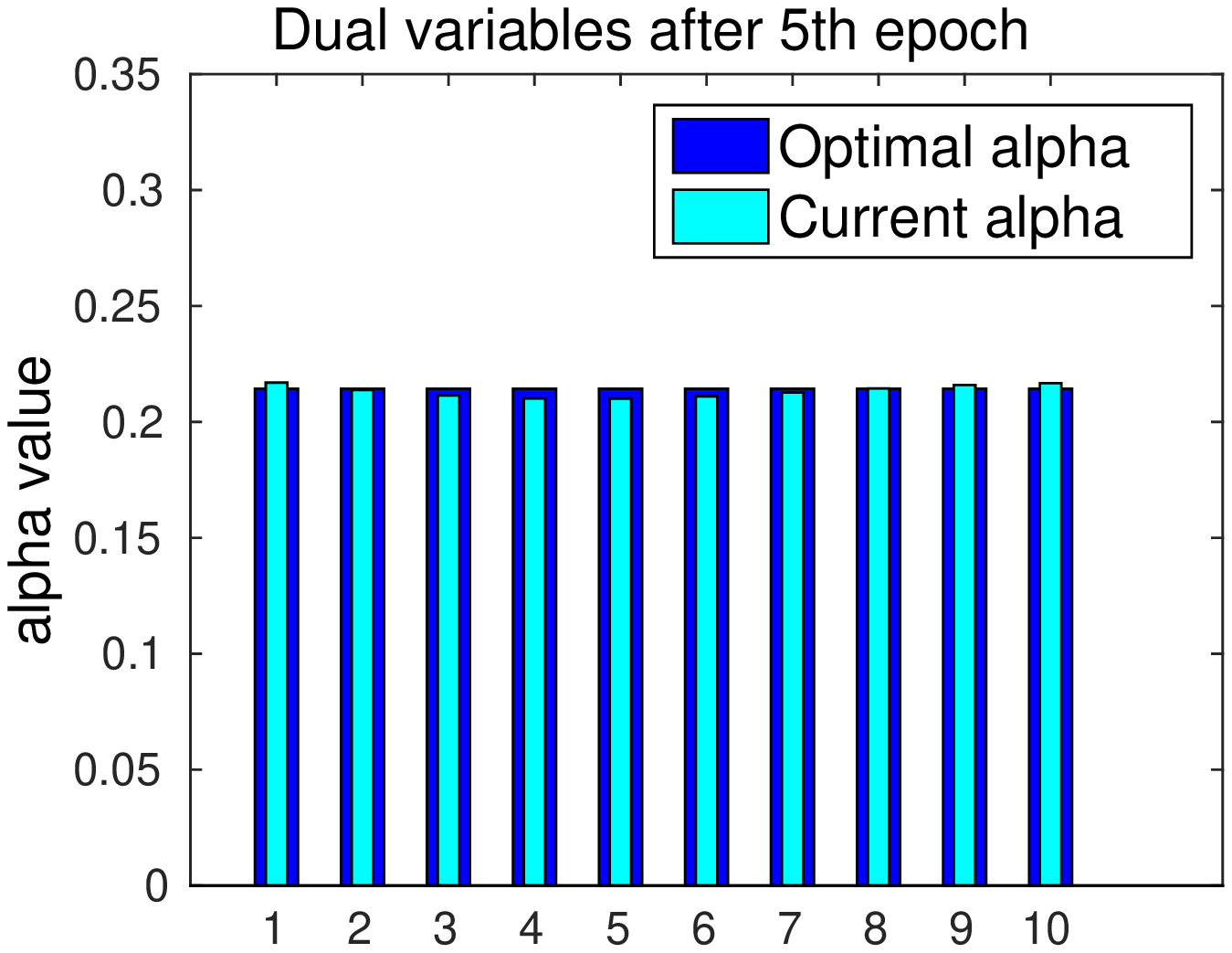}%
\end{center}
\makebox[0.25 \textwidth]{(e)}\makebox[0.25 \textwidth]{(f)}\makebox[0.25 \textwidth]{(g)}\makebox[0.25 \textwidth]{(h)}
\caption{A synthetic example to demonstrate to behavior of SDCA}
\label{fig:demonstration}
\end{figure*}

In this section, we explore why for SDCA with random permutation, the
optimal $c$ is usually just below 1 (around $0.9<c<1$). We show a
previously unexplained behavior of SDCA-Perm (i.e.~using an
independent random permutation at each epoch) that could be useful for
understanding the test error as $c$ changes.  All theoretical analysis
of SDCA we are aware of are of i.i.d.~sampling (with replacement), and
although known to work well in practice, not much is understood
theoretically on SDCA-Perm.  Here we show its behavior is more
complex than what might be expected.

Many empirical studies of SDCA plot the sub-optimality only after
integer numbers of epochs. Furthermore, often only the dual, or
duality gap, is investigated. Here we study the detailed behavior of
the primal suboptimality, especially at the epoch transition period.
We experimented with the same datasets as used in previous section,
randomly choose a $m = 4000$ subset (we observe the same experimental
phenomenon for all dataset size, here we report on subsets of size $m
= 4000$ for simplicity). We test with $\lambda$: $0.1$ and $0.01$ (the
optimal regularization lies between these two values). We ran SDCA-IID
and SDCA-Perm $500$ times in Figure \ref{fig:convergence} plot the
average behavior across the runs: the primal sub-optimality,
dual-optimality and the duality gap of the iterates. We observe that:
\begin{itemize}
\item The behavior of SDCA-IID is as expected monotonic and mostly
  linear. Also, as is well know, SDCA-Perm usually converges faster
  than the SDCA-IID after the first epoch.
\item SDCA-Perm displays a periodic behavior at each epoch with
  opposite behaviors for the primal and dual suboptimaitiesl: the primal
  decreases quickly at the beginning of the epoch, but is then flat
  and sometimes even increases toward the end of the epoch;  The dual
  suboptimality usually decreases slowly at the beginning, but then
  drops toward the end of the epoch.
\end{itemize}

This striking phenomena is consistent across data sets. The periodic
behavior explains why for SDCA-Perm the optimal $c$ is usually between
$0.9$ and $1$: since the primal improves mostly at the beginning of an
epoch, we will prefer to run SDCA-Perm just more than integer number
of epochs to obtain low optimization error. Returning to Figure
\ref{fig:testerror1}, we can further see that the locally best $c$, for SDCA-Perm
are indeed just lower than integer fractions (just before
$1/2,1/3,1/4$ etc), again corresponding to running SDCA-Perm for a bit
more than an integer number of epochs.

To understand the source of this phenomena, consider the following
construction: A data set with $10$ data points in $\RR^{11}$, where each
data point $\xb_i$ has two non-zero entries: a value of $1$ in
coordinate $i$, and a random sign at the last coordinate.  The
corresponding label $y_i$ is set to the last coordinate of $\xb_i$.
Let us understand the behavior of SDCA on this dataset. In Figure
\ref{fig:demonstration}(a-b) we plot the behavior of SDCA-Perm on such
synthetic data, as well the behavior of SDCA-Cyclic. SDCA-Cyclic is a
deterministic (and thus easier to study) variant where we cycle
through the training examples in order instead of using a different
random permutation at each iterations. We can observe the phenomena
for both variants, and will focus on SDCA-Cyclic for simplicity.  In
Figure \ref{fig:demonstration}(c) we plot the loss and norm parts of
the primal objective separately, and observe that the increase in the
primal objective at the end of each epoch is due to an increase in the
norm without any reduction in the loss. To understand why this
happens, we plot the values of the 10 dual variables at the end of
each epoch (recall that the variables are updated in order). The
first variables updates at each epoch are set to rather large values,
larger than their values at the optimum, since such a value is optimal
when the other dual variables are zero. However, once other variables
are increased, in order to reduce the norm, the initial variables set
must be decreased---this is not possible without revisiting them
again. Although real data sets are not as extreme case, it seems that such
a phenomena do happen also there.

\section{Conclusion}
\label{sec:conclusion}

We have shown that contrary to Stochastic Gradient Descent, when using
variance reducing stochastic optimization approaches, it might be
beneficial to use less samples in order to make
more than one pass over (some of) the training data. This behavior is
qualitatively different from the observation made about SGD where
using more samples can only reduce error and runtime. Furthermore, we
showed that the optimal training set size (i.e.,~optimal amount of
recycling) for SDCA with random permutation sampling (so-called
``sampling without replacement'') rests heavily on a previously
undiscovered phenomena that we uncover here.

Our observations provide empirical guidance for using SDCA, SAG and SVRG:

First, it suggests that even when data is plentiful, it might be
beneficial to use a limited training set size in order to reduce
runtime or improve accuracy after a fixed number of iterations.  For
SDCA-Perm , it seems that the optimal strategy is often to use a slightly smaller training set
than the maximal possible, and for SVRG the optimal strategy is to use
a $m$ slightly smaller than $T/2$. For SAG the optimal number of
examples is more variable. Our observations are mostly empirical,
backed only by qualitative reasoning---obtaining a firmer
understanding with more specific guidelines of the optimal number of
samples to use would be desirable.

Second, the behavior of the SDCA primal objective that we uncover
suggests that performing an integer number of epochs (passes over the
data), as is frequently done in practice and is the default for most
SDCA packages, can significantly hurt the performance of SDCA.  This
is true regardless of whether we are in a data-laden regime or in a
data-limited regime where we are performing multiple passes out of
necessity. Instead, our observations suggest it is often advantageous
to perform a few more iterations into the next epochs in order to
significantly improve the solution. Further understanding of the
non-monotone SDCA behavior is certainly desirable (and challenging),
and we hope that pointing out the phenomena can lead to further
research on understanding it, and then to devising improved methods
with more sensible behavior.

\clearpage

{
\bibliographystyle{icml2016}
\bibliography{references}
}

\clearpage
\onecolumn
\section*{Appendix: Additional Empirical Results}

\begin{figure*}[h]
\vspace{-0.15 cm}
\begin{center}
\includegraphics[width=0.25 \textwidth]{newfigures4/covtype_1000_testerror.eps}%
\includegraphics[width=0.25 \textwidth]{newfigures4/covtype_2000_testerror.eps}%
\includegraphics[width=0.25 \textwidth]{newfigures4/covtype_4000_testerror.eps}%
\includegraphics[width=0.25 \textwidth]{newfigures4/covtype_8000_testerror.eps}%
\end{center}
\makebox[\textwidth]{{\it covtype, from left to right: T = 1000,2000,4000,8000}}
\begin{center}
\includegraphics[width=0.25 \textwidth]{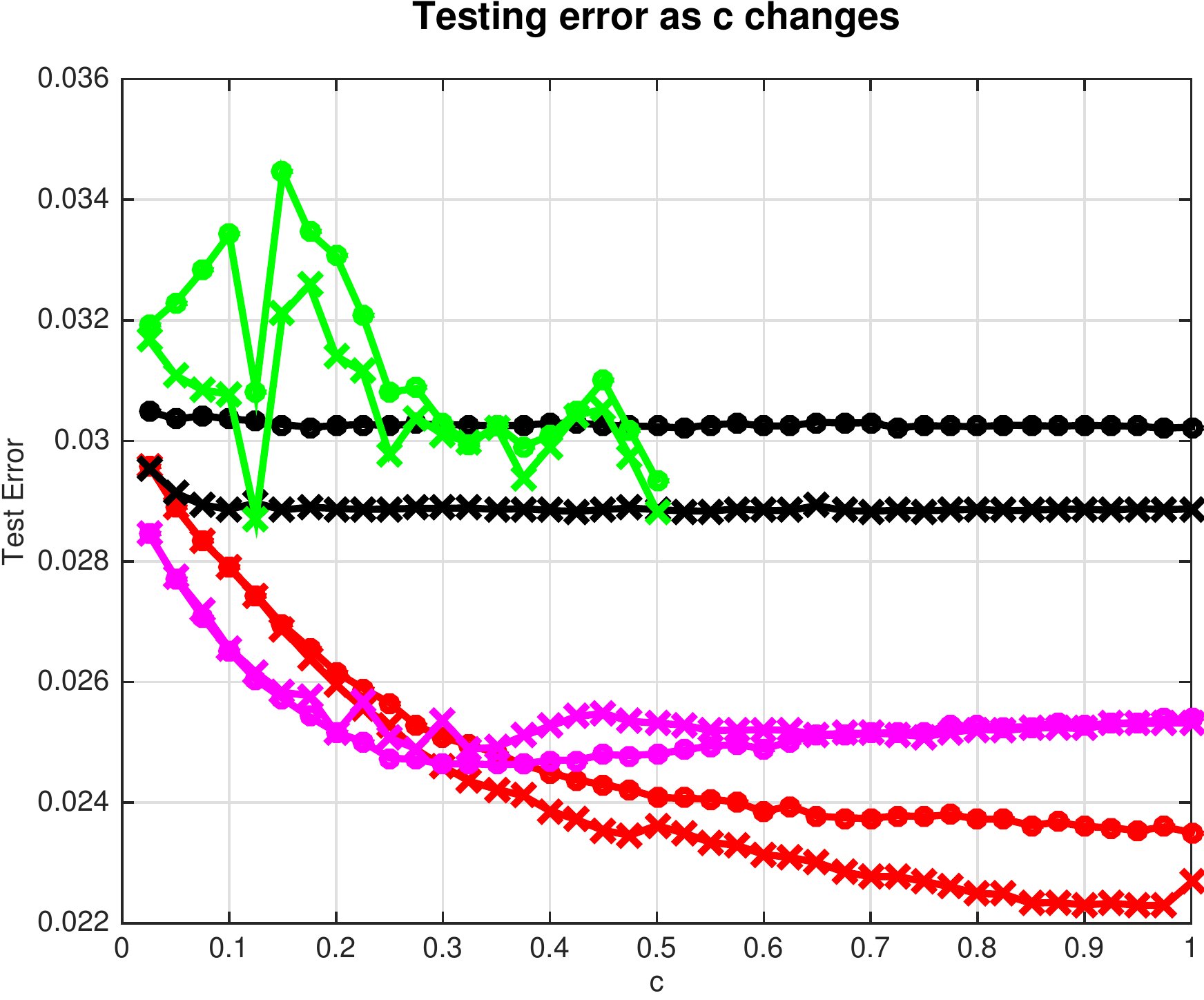}%
\includegraphics[width=0.25 \textwidth]{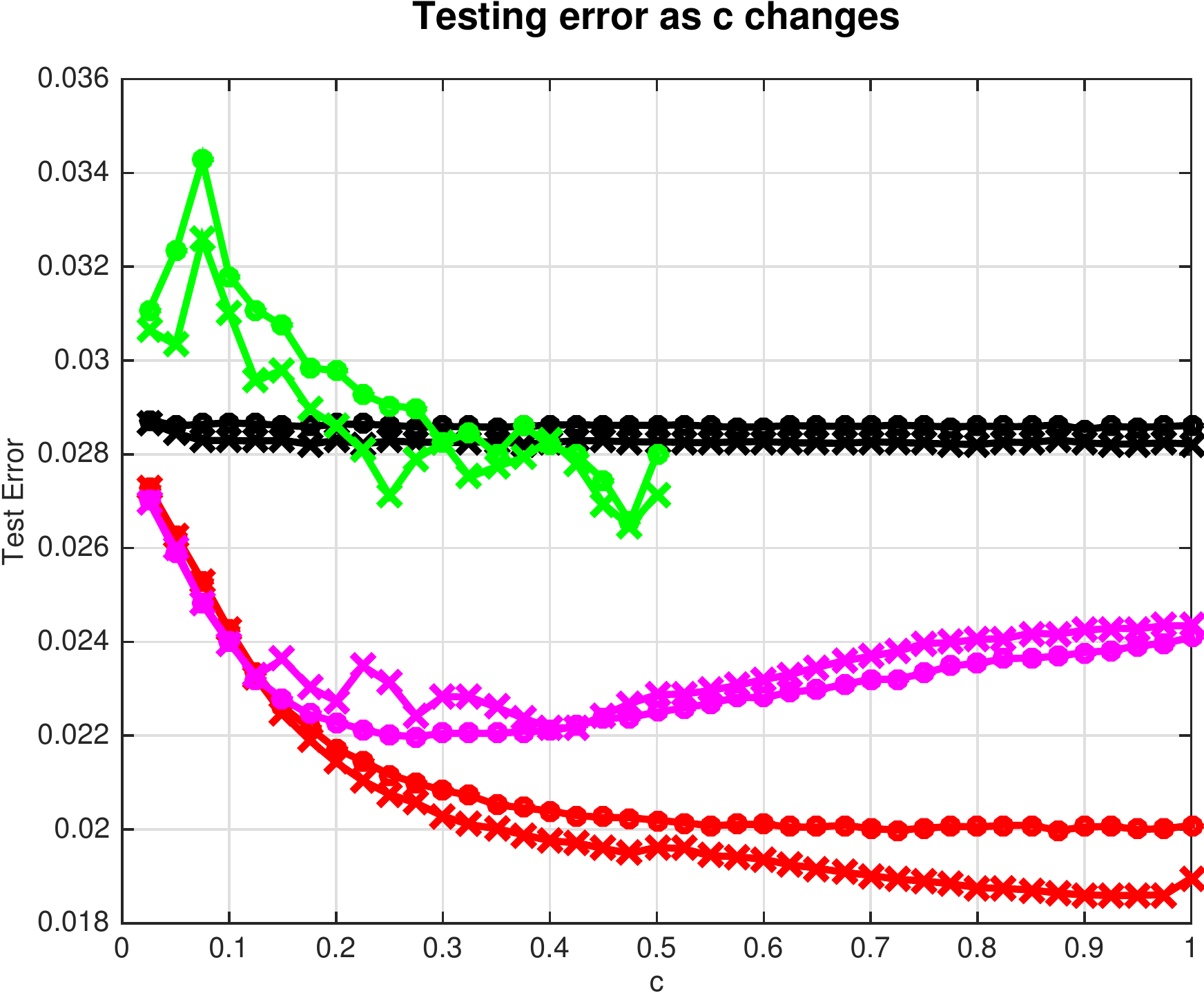}%
\includegraphics[width=0.25 \textwidth]{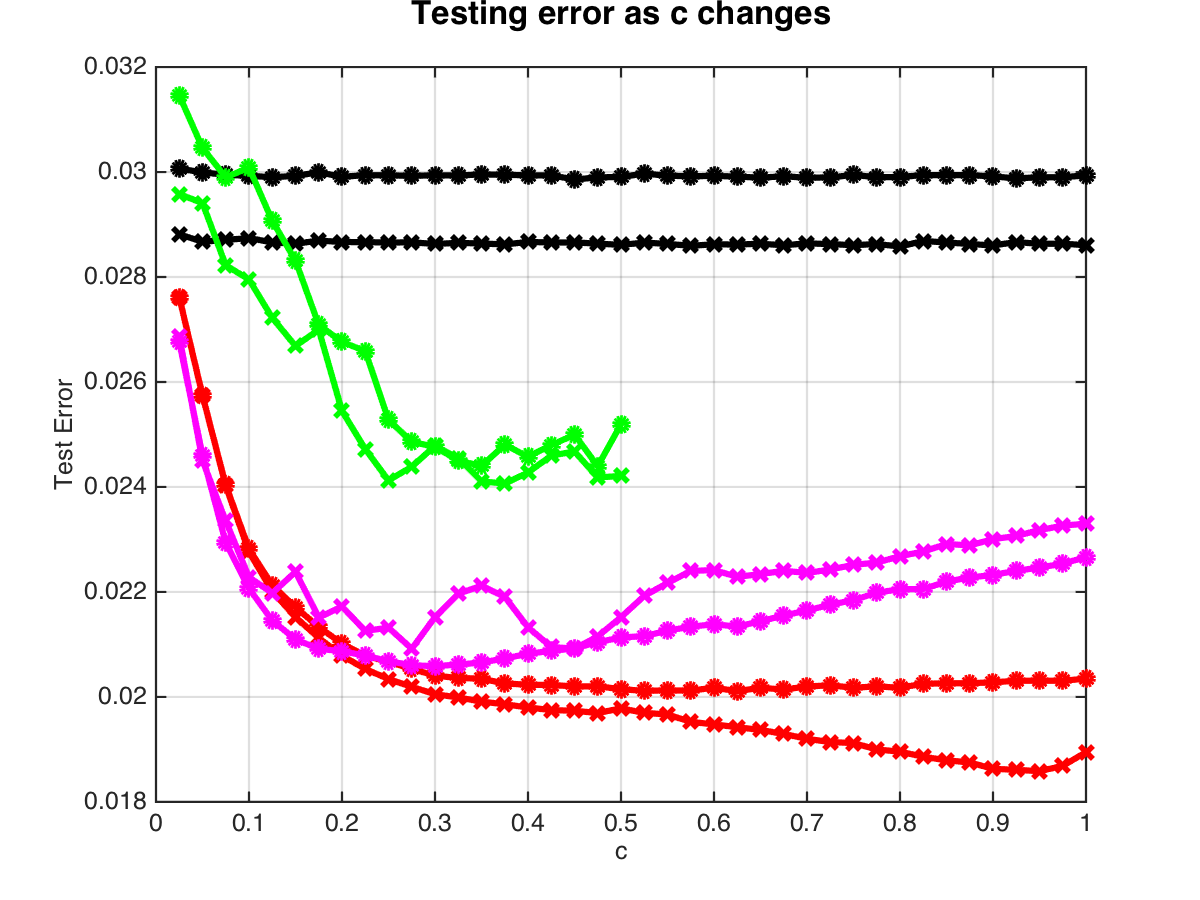}%
\includegraphics[width=0.25 \textwidth]{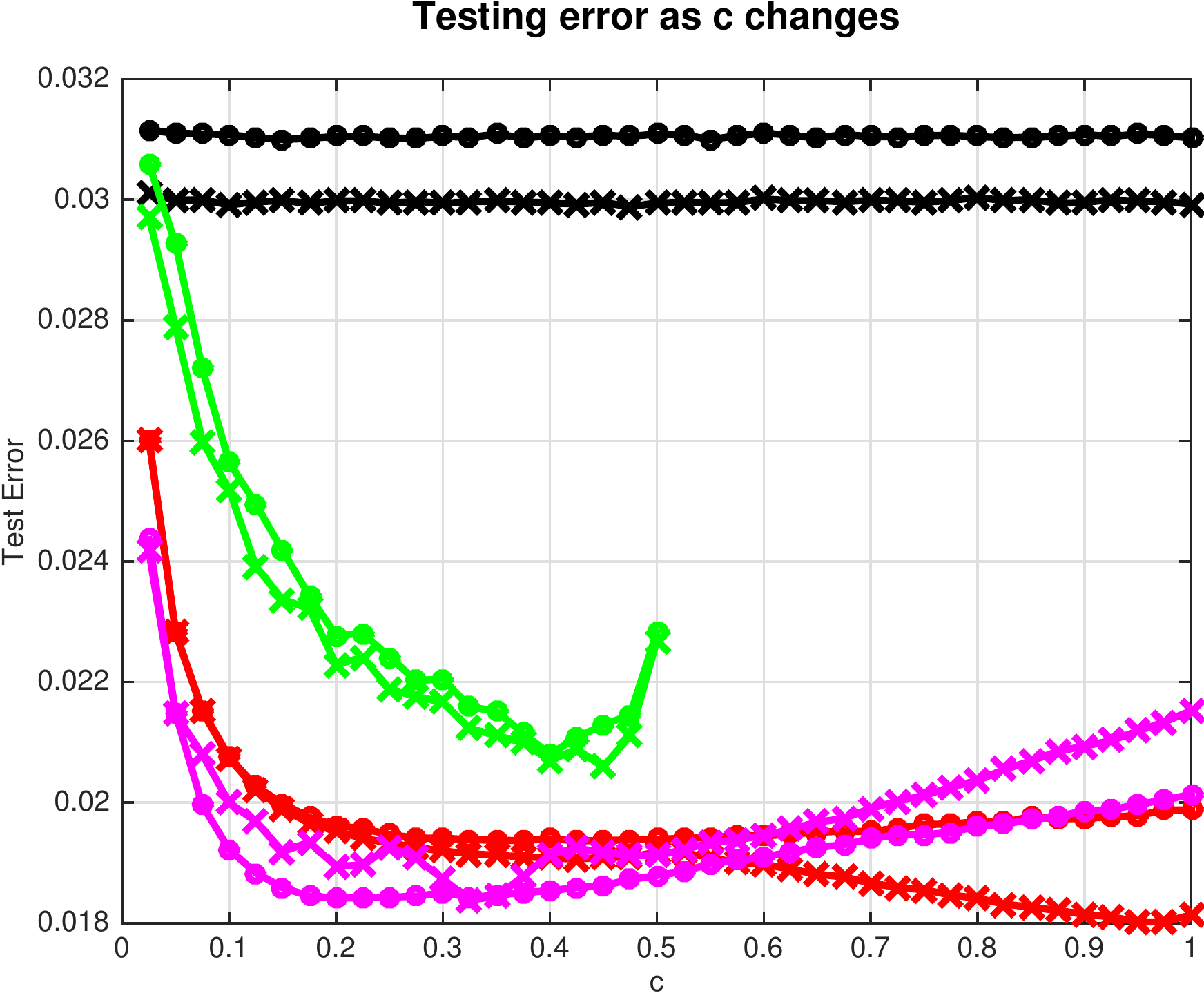}%
\end{center}
\makebox[ \textwidth]{{\it w8a, from left to right: T = 4000,8000,16000,32000}}
\begin{center}
\includegraphics[width=0.25 \textwidth]{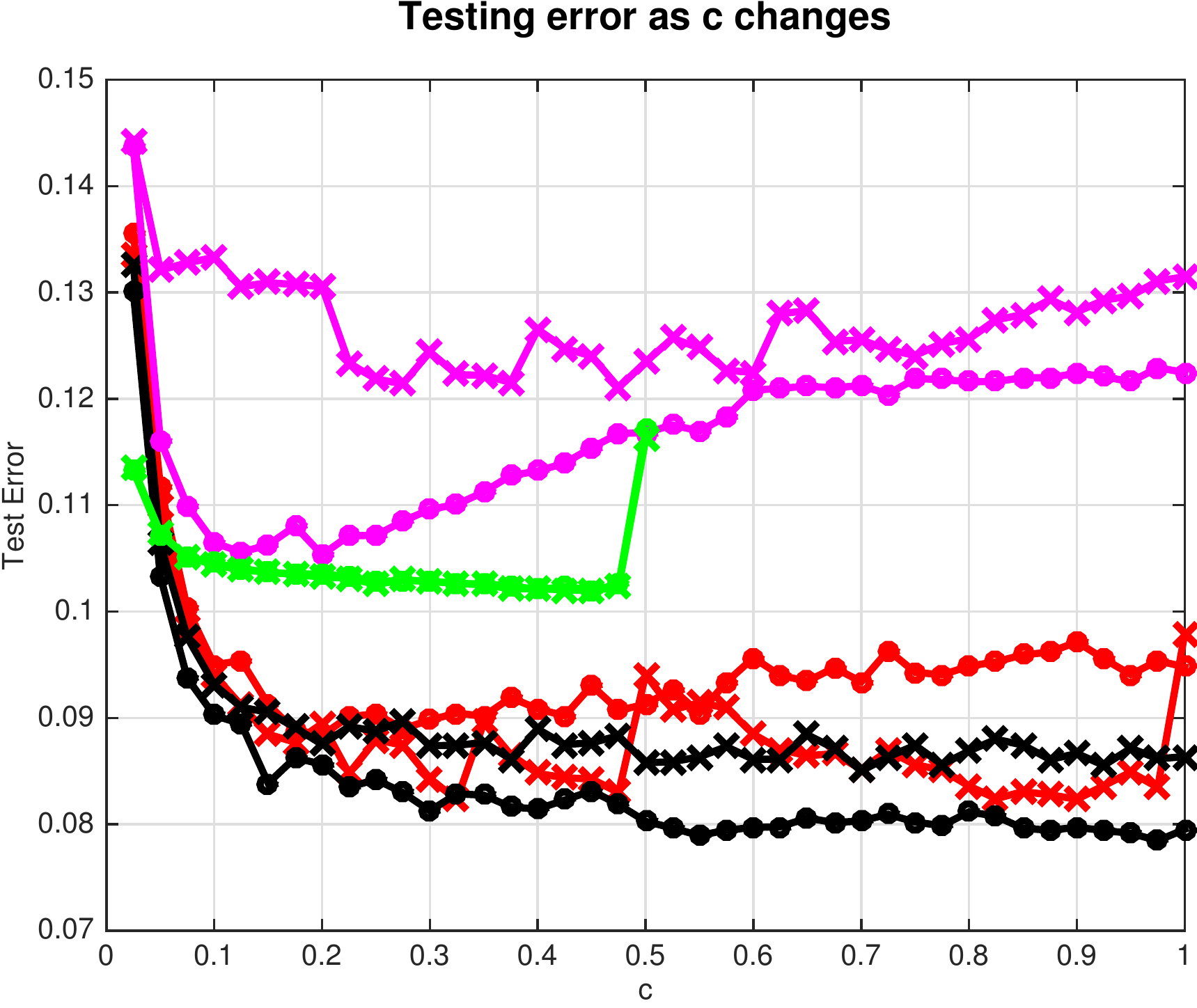}%
\includegraphics[width=0.25 \textwidth]{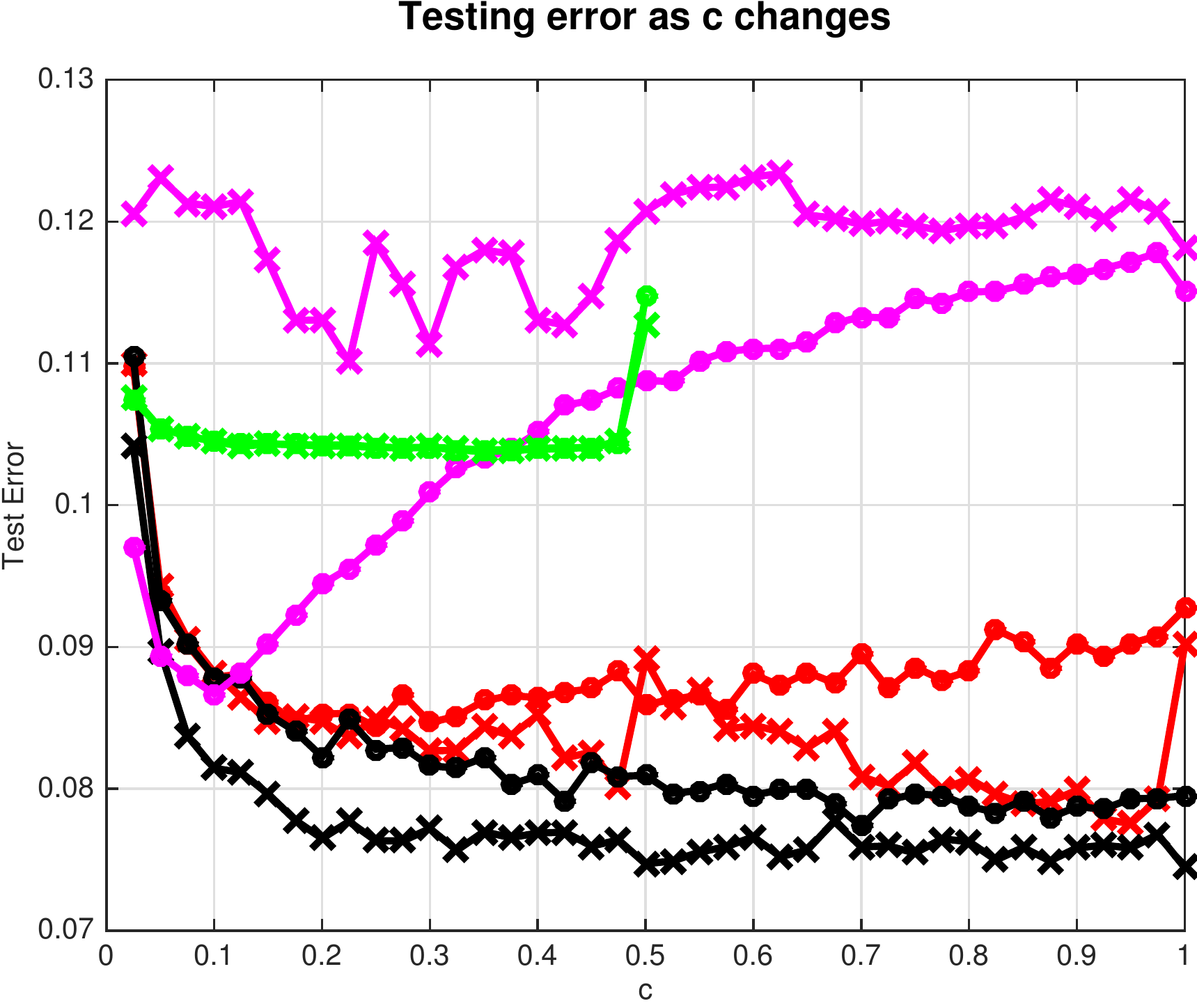}%
\includegraphics[width=0.25 \textwidth]{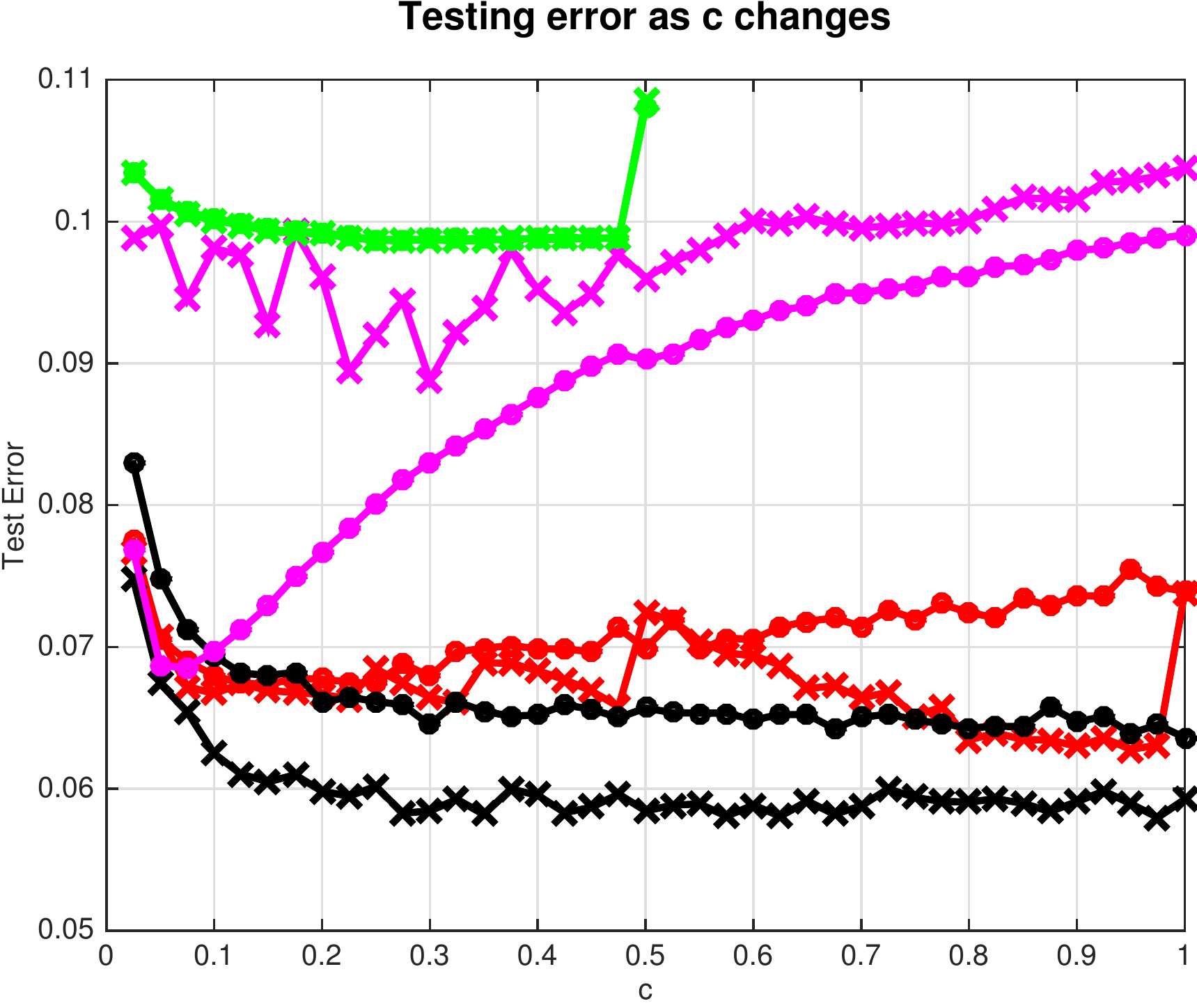}%
\includegraphics[width=0.25 \textwidth]{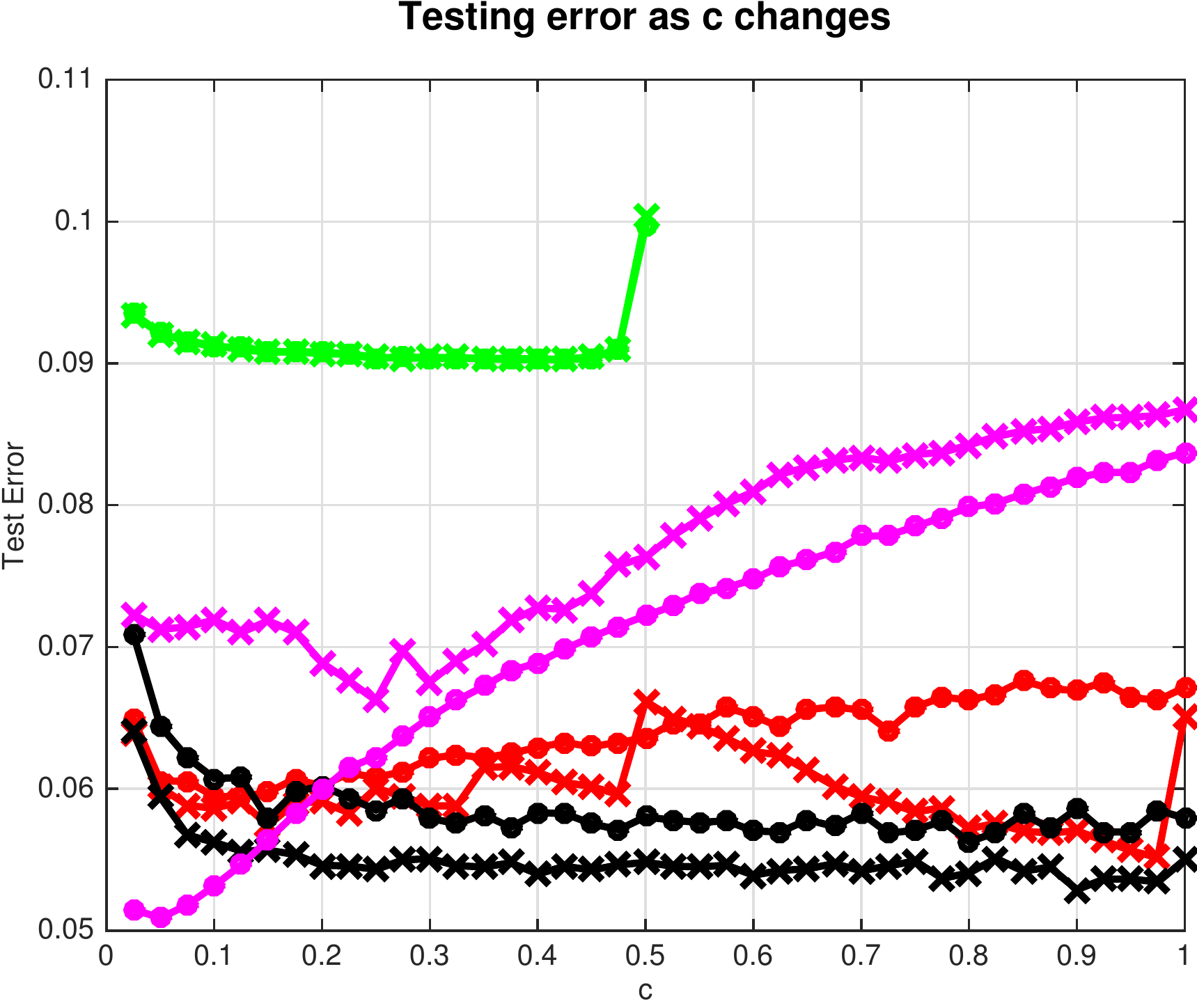}%
\end{center}
\makebox[ \textwidth]{{\it svmguide1, from left to right: T = 500,1000,2000,4000}}
\begin{center}
	\includegraphics[width=0.25 \textwidth]{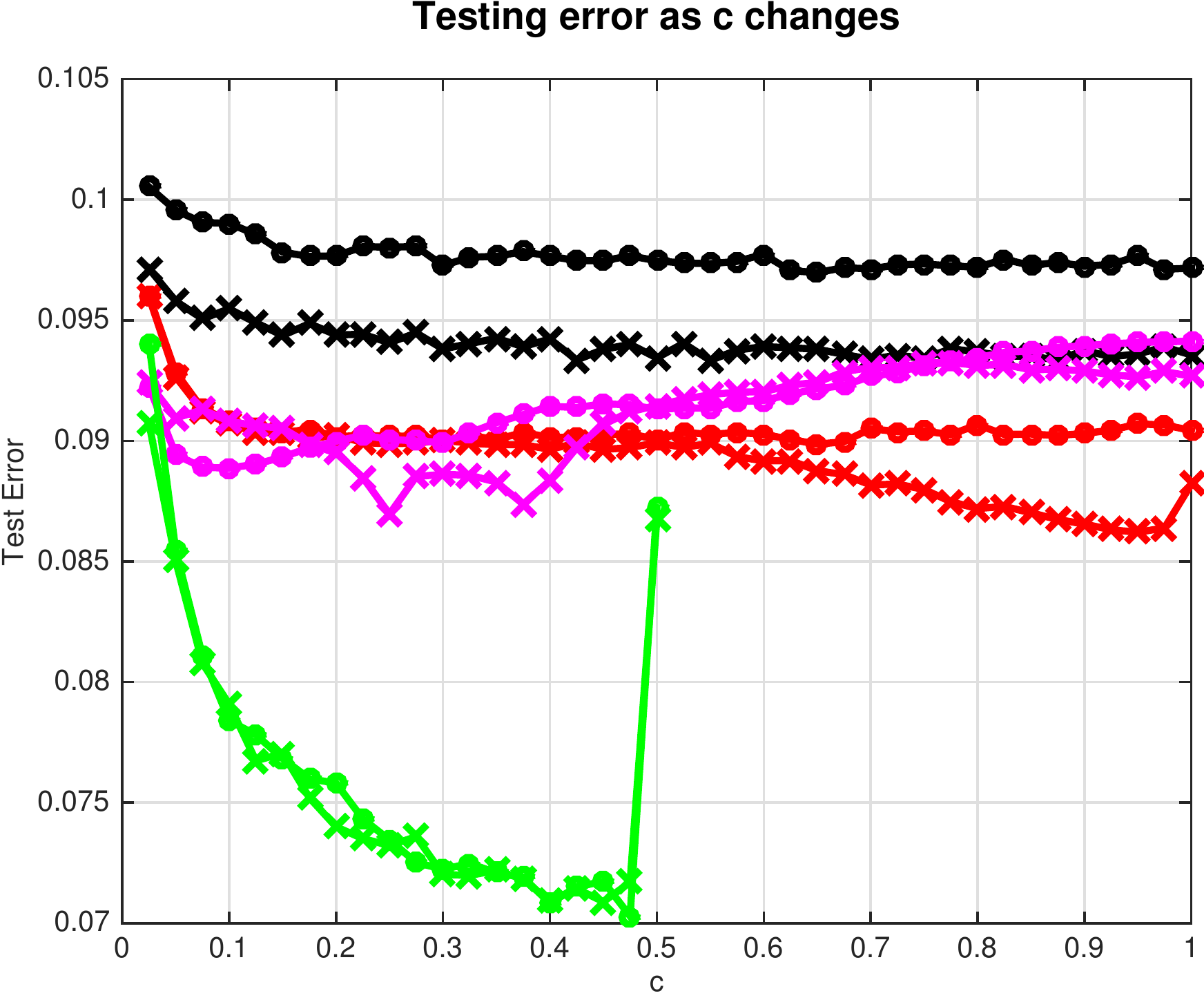}%
	\includegraphics[width=0.25 \textwidth]{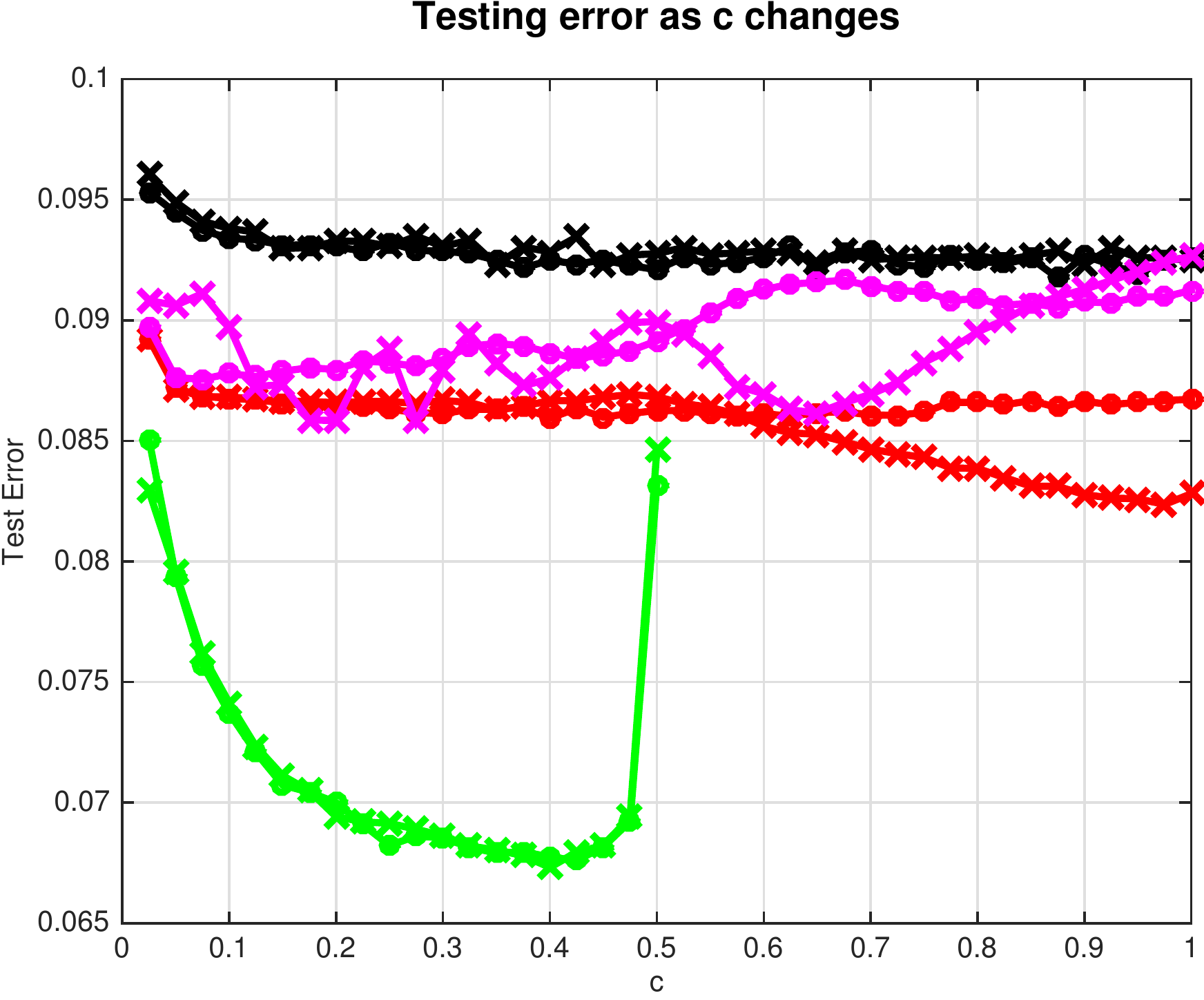}%
	\includegraphics[width=0.25 \textwidth]{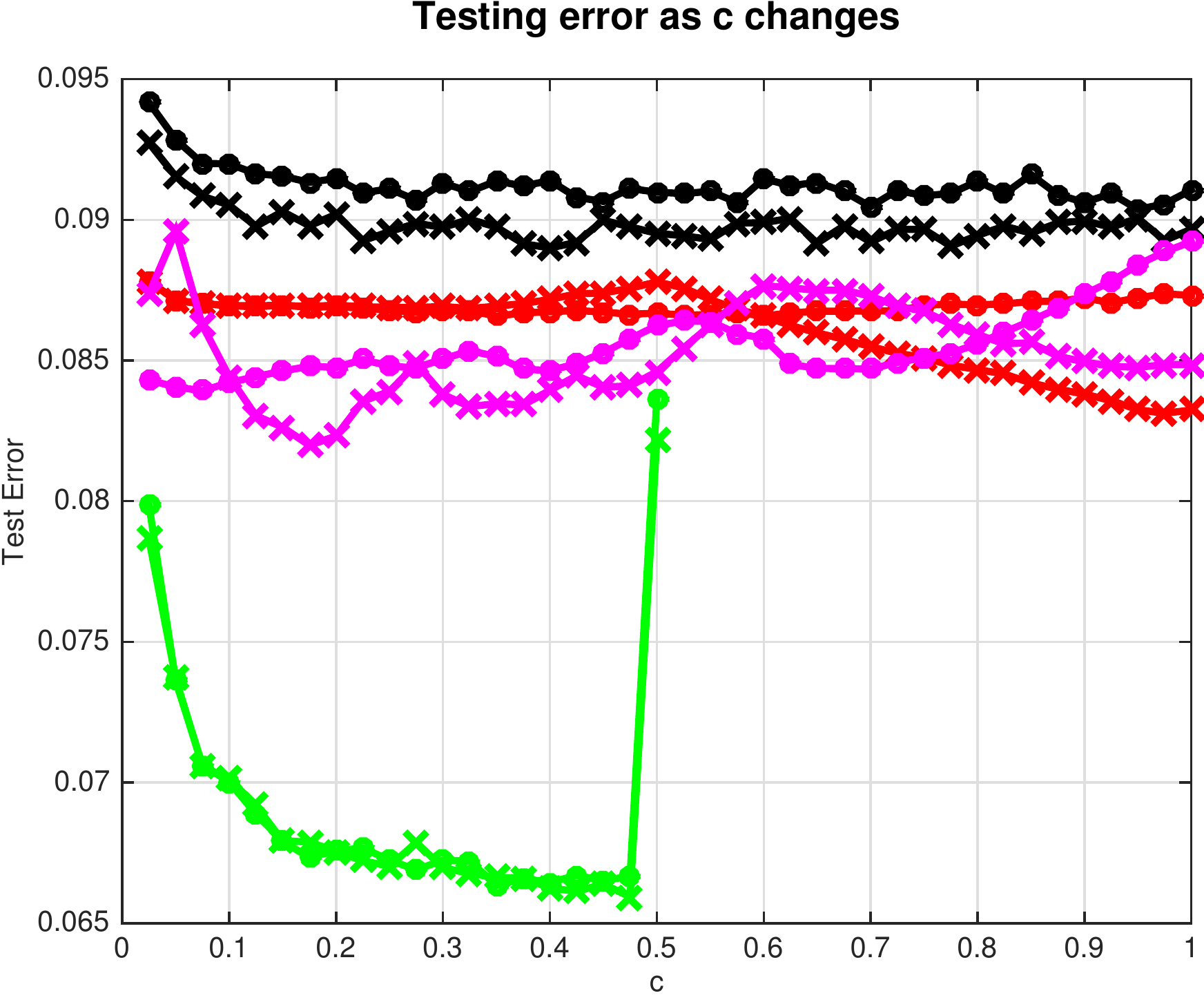}%
	\includegraphics[width=0.25 \textwidth]{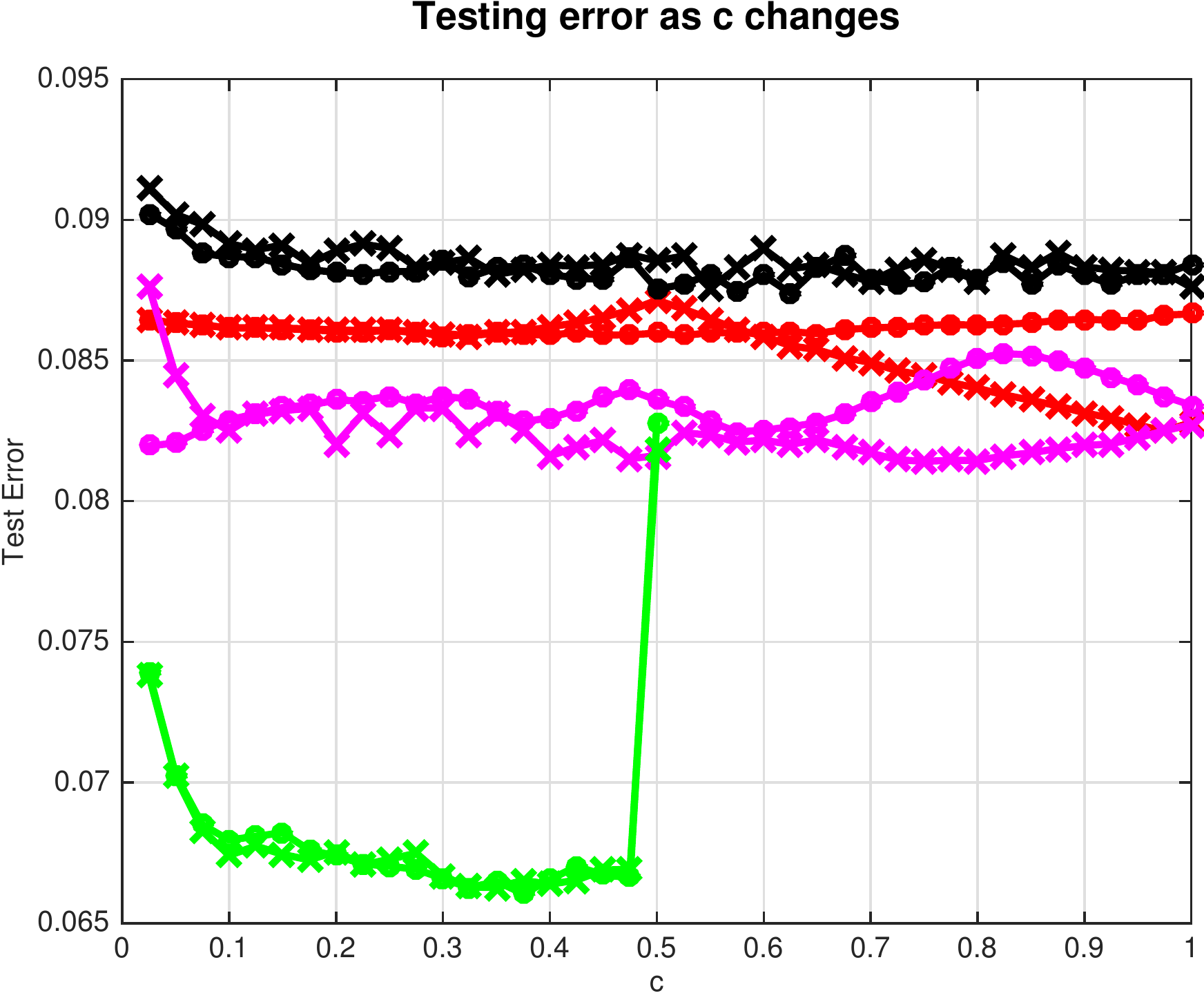}%
\end{center}
\makebox[ \textwidth]{{\it ijcnn1, from left to right: T = 4000,8000,16000,32000}}
\vspace{-0.3 cm}
\caption{Illustration of generalization errors as $c$ varied}
\label{fig:testerror_appendix}
\vspace{-0.3 cm}
\end{figure*}

\end{document}